\documentclass[
a4paper,
]{article}

\usepackage{papers}

\usepackage{enumitem}

\usepackage{float}

\usepackage
[
]
{geometry}

\usepackage{setspace}
\newcommand{\FBV}{\mathop{\bigvee}
\limits
}

\usepackage{makecell}

\usepackage{lineno}

\title{Historical/temporal necessities/possibilities, \\ and a logical theory of them in branching time
}

\author{
Fengkui Ju${}^{1,2}$ and Woxuan Zhou${}^{3,4}$\footnote{Corresponding author} \vspace{5pt} \\
{\small $^1$School of Philosophy, Beijing Normal University, 100875 Beijing, China} \vspace{3pt} \\
{\small $^2$\href{mailto:fengkui.ju@bnu.edu.cn}{fengkui.ju@bnu.edu.cn}} \vspace{3pt} \\
{\small $^3$Institute for Logic, Language and Computation, University of Amsterdam,} \\
{\small 1098 XG Amsterdam, The Netherlands} \vspace{3pt} \\
{\small $^4$\href{mailto:w.zhou@uva.nl}{w.zhou@uva.nl}}
}
\date{}

\begin{document}

\maketitle

\setlength{\parskip}{0.33em}


\begin{abstract}

\noindent In this paper, we do three kinds of work.
First, we recognize four notions of necessity and two notions of possibility related to time flow, namely strong/weak historical/temporal necessities, as well as historical/temporal possibilities, which are motivated more from a linguistic perspective than from a philosophical one.
Strong/weak historical necessities and historical possibility typically concern the possible futures of the present world, and strong/weak temporal necessities and temporal possibility concern possible timelines of alternatives of the present world.
Second, we provide our approach to the six notions and present a logical theory of them in branching time.
Our approach to the six notions is as follows. The agent has a system of ontic rules that determine expected timelines. She treats some ontic rules as undefeatable, determining accepted timelines.
The domains of strong/weak historical necessities, respectively, consist of accepted and expected timelines passing through the present moment, and historical possibility is the dual of strong historical necessity.
The domains of strong/weak temporal necessities, respectively, consist of accepted and expected timelines, and temporal possibility is the dual of strong temporal necessity.
The logical theory has six operators: a last-moment operator, a next-moment operator, and four operators for the four notions of necessity.
Formulas' evaluation contexts consist of a tree-like model representing a time flow, a context representing the agent's system of ontic rules, a timeline, and an instant.
Third, we offer an axiomatic system for the logical theory and show its soundness and completeness.

\medskip

\noindent \textbf{Keywords:} strong/weak historical/temporal necessities, historical/temporal possibilities, ontic systems, undefeatable ontic rules, accepted/expected timelines

\end{abstract}

\section{Introduction}
\label{section: Introduction}

\newcommand{\putaway}[1]{}

\subsubsection*{Our work}

\emph{Necessities} and \emph{possibilities} are studied in various contexts, including linguistic, logical, philosophical, ethical, and legal. From a semantic perspective, they are respectively universal and existential quantifications over domains of alternatives\footnote{A better notion than alternative is possibility. However, possibility is used as a modality in this paper. To avoid confusion, we use alternative.}. There are many kinds of necessities and possibilities: deontic, epistemic, metaphysical, logical, and so forth, which are over different types of alternatives.
Necessities can be expressed by ``necessary'', ``must'', ``have to'', ``should'', ``ought to'', and so on, and possibilities can be expressed by ``possible'', ``can'', ``could'', ``might'', and so on.
For general discussions about necessities and possibilities, we refer to \cite{fintel_modality_2006}, \cite{portner_modality_2009}, and \cite{kment_varieties_2021}.
There are \emph{strong} and \emph{weak} necessities, and the former have bigger domains than the latter \cite{mcnamara_must_1996, copley_what_2006, fintel_how_2008, portner_extreme_2016}. Strong necessities are often expressed by ``necessary'', ``must'', or ``have to'', and weak necessities are often expressed by ``should'' or ``ought to''.

\emph{Temporal alternatives} are possible states that we think our world, in the \emph{ontic} sense, was in or could be in in the past, is in or could be in in the present, or will possibly be in or would possibly be in in the future.
In this paper, we first distinguish four notions of necessity and two notions of possibility over temporal alternatives: strong/weak historical/temporal necessities, as well as historical/temporal possibilities. These notions are motivated more from a linguistic perspective than from a philosophical perspective. Second, we present our approach to the six notions, which is based on ontic systems, and propose a logical theory of them in branching time. Third, we provide an axiomatic system for the theory and show its soundness and completeness.

\subsubsection*{Historical/temporal necessities/possibilities}

Historical necessities/possibility are over temporal alternatives, typically concerning what the present world will possibly be like in the future, given what has happened. What follows are some typical examples:

\begin{exe}

\ex\label{exe:There will necessarily be a sea battle tomorrow}

\exs{The two countries have been friendly for decades, but now our king doesn't believe that his daughter died accidentally in your country. There will \textbf{necessarily} be a sea battle tomorrow.}

\ex\label{exe:I should be late for work}

One morning, Adam drove to work as usual. Halfway, his car had a flat tire. After parking the car, Adam sent a message to his boss:

\vspace{2pt}

\exs{I \textbf{should} be late for work.}

\ex\label{exe:He might be discharged from the hospital soon}

\exs{His blood pressure has been high these days, but it came down today. He \textbf{might} be discharged from the hospital soon.}

\end{exe}

Temporal necessities and possibility are over temporal alternatives, concerning what the world could have been like in the past, what it is like in the present, or what it could be like in the future. What follows are some typical examples:

\begin{exe}

\ex\label{he would still necessarily have lost the election}

\exs{Even if he hadn’t made the mistake, he would still \textbf{necessarily} have lost the election.}

\ex\label{ex:I ought to be dead right now}

Suppose Jones is in a building when an earthquake hits. The building collapses. Luckily, nothing falls upon Jones and he emerges from the rubble as the only survivor. Talking to the media, Jones says the following:

\vspace{2pt}

\exs{I \textbf{ought to} be dead right now.}
\ecs{From \cite{yalcin_modalities_2016}}

\ex\label{exe:We could have done that years ago}

\exs{We \textbf{could} have done that years ago and saved thousands of American lives.}
\ecs{From Corpus of Contemporary American English, COCA}

\end{exe}

Strong historical/temporal necessities can be expressed by ``necessarily'' or ``must'', while weak historical/temporal necessities can be expressed by ``should'' or ``ought to''.

\subsubsection*{Our ways to distinguish historical/temporal necessities/possibilities}

We will focus on the following kinds of sentences, where $p$ does not concern the future:
\begin{itemize}

\item 

\begin{itemize}

\item 

$p$

\item

$p$ will be true

\end{itemize}

\item

\begin{itemize}

\item 

It must be the case that $p$.

\item

It must be the case that $p$ will be true.

\end{itemize}

\item 

\begin{itemize}

\item 

It should be the case that $p$.

\item

It should be the case that $p$ will be true.

\end{itemize}

\item 

\begin{itemize}

\item 

It could be the case that $p$.

\item

It could be the case that $p$ will be true.

\end{itemize}

\end{itemize}

\noindent We will distinguish the four notions of necessity and the two notions of possibility by discussing the dual relation, monotonicity, the implication relation, and the coexistence of these kinds of sentences.

\subsubsection*{Our approach to historical/temporal necessities/possibilities}

Our approach to the six notions is as follows.
There is an agent that accepts a system of ontic rules concerning how the world evolves in time. The system determines which timelines are \emph{expected}. The agent treats some ontic rules as undefeatable, determining which timelines are \emph{accepted}.
The domains of strong/weak historical necessities, respectively, consist of accepted and expected timelines passing through the present moment, and historical possibility is the dual of strong historical necessity.
The domains of strong/weak temporal necessities, respectively, consist of accepted and expected timelines, no matter whether they pass through the present moment, and temporal possibility is the dual of strong temporal necessity.

\subsubsection*{Our logical theory}

The logical theory of historical/temporal necessities/possibilities has six operators: a \emph{last moment} operator, a \emph{next moment} operator, and four operators for the four notions of necessity.
Formulas' evaluation contexts consist of a tree-like model representing a time flow, a context representing an ontic system, a timeline, and an instant.

\subsubsection*{The structure of this paper}

In Section \ref{section: Before we begin}, we lay the groundwork for discussing historical/temporal necessities/possibilities.
In Section \ref{section: Historical/temporal necessities}, we introduce historical/temporal necessities.
In Section \ref{section: Historical/temporal possibilities}, we introduce historical/temporal possibilities.
In Section \ref{section: Strong/weak historical/temporal necessities}, we discuss strong/weak historical/temporal necessities. In Section \ref{section: Putting everything together}, we sum up the discussed features of the six notions.

In Section \ref{section: Our approach to strong/weak historical/temporal necessities based on ontic systems}, we present our approach to the six notions.
Following the approach, we propose a logical theory for them in Section \ref{section: A logical theory for strong/weak historical/temporal necessities in branching time}.
In Section \ref{section: Comparison to some related work}, we compare the logical theory to some related work.

We present an axiomatic system for the logical theory in Section \ref{section: An axiomatic system for SWHTN}, whose soundness is proved in Section \ref{section: Soundness} and whose completeness is proved in Section \ref{section: Completeness}.
We make some concluding remarks in Section \ref{section: Looking backward and forward}.

\part{}
\textit{%
In this part, we distinguish the four notions of necessity and the two notions of possibility.
}

\newcommand{\Fdefs}[1]{\textbf{#1}}

\section{Before discussing historical/temporal necessities/possi-bilities}
\label{section: Before we begin}

In this section, we briefly discuss strong/weak deontic/epistemic necessities, temporal alternatives, our suppositions about timeflow, will-sentences, and their declaration.

\subsection{Strong/weak deontic/epistemic necessities}
\label{section:??}

Deontic necessities expressed by ``must'' and by ``should'' have different properties. For detailed discussion, we refer to \cite{mcnamara_must_1996}, \cite{fintel_how_2008} and \cite{portner_extreme_2016}. Here, we briefly mention some differences between them.

First, ``should $\phi$'' is weaker than ``must $\phi$''.

\begin{exe}

\ex\label{exe:??}

\exs{After using the bathroom, everybody \textbf{should} wash their hands; employees \textbf{must}.}
\ecs{From \cite{portner_extreme_2016}}

\end{exe}

Second, ``should $\phi$'' can coexist with ``can $\neg \phi$'', but ``must $\phi$'' cannot.

\begin{exe}

\ex\label{exe:??}

\exs{Although you can skip the meeting, you \textbf{should} attend.}
\ecs{Adapted from \cite{sep-logic-deontic}}

\ex\label{exe:??}

\exs{\#Although you can skip the meeting, you \textbf{must} attend.}

\end{exe}

Third, ``should $\phi$'' can coexist with ``$\neg \phi$'', but ``must $\phi$'' cannot.

\begin{exe}

\ex

\exs{Sam \textbf{should} go to confession, but he’s not going to.}
\ecs{From \cite{Ninan05}}

\ex

\exs{\# Sam \textbf{must} go to confession, but he’s not going to.} 
\ecs{From \cite{Ninan05}}

\end{exe}

Fourth, ``should $\phi$'' is gradable, but ``must $\phi$'' is not.

\begin{exe}

\ex\label{exe:??}

\exs{You \textbf{should} call Zoe more than Barbara.}
\ecs{From \cite{portner_extreme_2016}}

\ex\label{exe:??}

\exs{\#You \textbf{must} call Zoe more than Barbara.}
\ecs{From \cite{portner_extreme_2016}}

\end{exe}

In addition, it is commonly noted that ``should $\phi$'' is not monotonic. However, ``must $\phi$'' seems monotonic.

\begin{exe}

\ex\label{exe:??}

\exs{Jones \textbf{should} help his neighbor tomorrow, but if it rains, he can stay at home.}

\ex\label{exe:??}

\exs{\# Jones \textbf{must} help his neighbor tomorrow, but if it rains, he can stay at home.}

\end{exe}

\medskip

``Can'' is the \emph{dual} of ``must'' \cite{mcnamara_must_1996}, which explains why ``must $\phi$'' cannot coexist with ``can $\neg \phi$''.
Does deontic ``should'' have a dual? It is unclear yet.

\medskip

Some differences between deontic ``must'' and ``should'' also apply to epistemic ``must'' and ``should''. For example, ``should $\phi$'' can coexist with ``can $\neg \phi$'', but ``must $\phi$'' cannot.

\begin{exe}

\ex\label{exe:??}

\exs{The man standing there \textbf{should} be Jones, although he can be Jack.}

\ex\label{exe:??}

\exs{\# The man standing there \textbf{must} be Jones, although he can be Jack.}

\end{exe}

\noindent However, neither ``should $\phi$'' nor ``must $\phi$'' can coexist with $\neg \phi$, which is different from epistemic ``should'' and ``must''.

\begin{exe}

\ex\label{exe:??}

\exs{\# Do you see the special hat? The man standing there \textbf{should} be Jones, but he is Jack.}

\ex\label{exe:??}

\exs{\# Do you see the special hat? The man standing there \textbf{must} be Jones, but he is Jack.}

\end{exe}

\newcommand{\will}{\mathsf{will}\hspace{1pt}}

\newcommand{\dec}{\mathsf{D}\hspace{1pt}}

\subsection{Temporal alternatives and timeflow}

\Fdefs{Temporal alternatives} are possible states that we think our world, in the \emph{ontic} sense, was in or could be in in the past, is in or could be in in the present, or will possibly be in or would possibly be in in the future.

Temporal alternatives are different from \emph{epistemic} alternatives.
Consider an example. Suppose Adam bought a lottery ticket yesterday; the winning number was just revealed, and Adam lost. Assume Bob knows that Adam lost. Then, the alternative in which Adam wins is temporal, but not epistemic, for Bob. Assume Bob does not know that Adam lost. Then, the alternative in which Adam wins is both temporal and epistemic for Bob.

In this work, we assume the common thought behind branching-time temporal logics \cite{prior_past_1967}.
The world is in a state, that is, the \emph{actual} one. It has a unique past, that is, the past is determined. It can evolve in various ways and has different possible futures. A \emph{timeline} is a complete flow of time.

\subsection{Will-sentences and their declaration}

Sentences such as ``there will be a sea battle tomorrow'' are called \Fdefs{will-sentences}. Later, they will be used to distinguish strong/weak historical/temporal necessities. Here, we briefly discuss them. We will revisit them. There are numerous issues related to will-sentences \cite{ohrstrom_future_2020}. In this paper, we do not aim to offer any deep analysis of them.

The history is determined. Consequently, the sentences not involving the future have settled truth values. In contrast, the future is open, and will-sentences do not have settled truth values.
However, will-sentences can be \emph{declared}, and their \Fdefs{declarations} seem to have truth values.

In the sequel, we use $p$ to indicate sentences not involving the future, $\will p$ to will sentences ``it will be the case that $p$'', and $\dec \will p$ to declarations of $\will p$.

Declarations often have wider scopes than Boolean connectives. For example, in the following sentence, ``not'' is inside but not outside the scope of the declaration.

\begin{exe}

\ex\label{exe:??}

\exs{It is not that it \textbf{will} be sunny tomorrow.}

\end{exe}

Will-sentences can occur in scopes of modals, explicitly or implicitly. What follows are some examples: 

\begin{exe}

\ex

\exs{There will \textbf{necessarily} be a sea battle tomorrow.}

\ex

\exs{There will \textbf{possibly} be a sea battle tomorrow.}

\end{exe}

The negation of $\will p$ is equivalent to $\will \neg p$. For example, intuitively, the following two sentences have the same meaning:

\begin{exe}

\ex\label{exe:??}

\exs{It \textbf{will} be cloudy tomorrow.}

\ex\label{exe:??}

\exs{It is not that it \textbf{will} be sunny tomorrow.}

\end{exe}

\newcommand{\HN}{\mathsf{HN}\hspace{1pt}}
\newcommand{\TN}{\mathsf{TN}\hspace{1pt}}

\section{Historical/temporal necessities}
\label{section: Historical/temporal necessities}

We first provide two motivating examples to illustrate historical/temporal necessities.

\begin{exe}

\ex\label{Football match}

Four soccer teams, A, B, C, and D, are competing in a friendly tournament. Teams A and B are very strong, while C and D are much weaker. The format is simple: A plays C, and B plays D; the winners advance to the final. Yesterday, A defeated C. Today, B was supposed to play D, but B withdrew, allowing D to advance directly to the final tomorrow.

\begin{xlist}

\ex Alice: \exs{Team A will \textbf{necessarily} win the championship tomorrow.}

\ex Bob: \exs{I don't think so—B’s withdrawal was unexpected.}

\end{xlist}

\end{exe}

\noindent Both Alice's and Bob's words seem reasonable, but they contradict each other at first glance. We think that Alice uses ``necessarily'' as a historical modal, whereas Bob understands it as a temporal modal.

\begin{exe}
 
\ex

While conducting a scientific expedition in the remote forests of South America, Adam was bitten by a venomous snake and fell into a coma. Fortunately, a local villager found him and administered an antidote. Soon after, Adam regained consciousness.

\begin{xlist}

\ex Alice: \exs{Adam \textbf{should} be fine soon.}

\ex Bob: \exs{Adam \textbf{should} be dead in a few hours, and the villager saved his life.}

\end{xlist}

\end{exe}

\noindent Both Alice's and Bob’s statements make good sense, but they are seemingly incompatible. We think that Alice uses ``should'' as a historical modal, while Bob uses it as a temporal modal.

\Fdefs{Historical necessities} are over temporal alternatives, concerning what the world was like in the past, is like in the present, or will possibly be like in the future, \emph{given what has happened}. Typically, they concern the future.
\Fdefs{Temporal necessities} are over temporal alternatives, concerning what the world was like or could be like in the past, is like or could be like in the present, or will possibly be like or would possibly be like in the future. 
Their difference lies in that the former concerns only possible timelines passing through the present world, while the latter concerns possible timelines from some point in the past, determined by \emph{contexts}.
In the sequel, we use $\HN \phi$ to indicate \emph{$\phi$ is a historical necessity}, and $\TN \phi$ to \emph{$\phi$ is a temporal necessity}.

What follows are some examples of historical necessities:

\begin{exe}

\ex\label{exe:Those flowers must be dead in a few days}

\exs{When Adam left home yesterday, he forgot to move the flowers indoors. Given the current conditions, all of those flowers will \textbf{necessarily} die within a few days.}

\ex

\exs{Though overall city spending will not \textbf{necessarily} increase because of the pay scales, the new system will put pressure on our budgets.}
\ecs{From COCA}

\ex\label{exe:He must lose}

\exs{He \textbf{must} lose. I think he knows he is going to lose because of all the executive orders he is shuffling out.}
\ecs{From COCA}

\ex\label{exe:We should have arrived before sunset}

\exs{This road is usually bustling, but the traffic is good today. We \textbf{should} have arrived before sunset.}

\exr{exe:I should be late for work}

One morning, Adam drove to work as usual. Halfway, his car had a flat tire. After parking the car, Adam sent a message to his boss:

\vspace{2pt}

\exs{I \textbf{should} be late for work.}

\end{exe}

\noindent Intuitively, in these examples, we only consider possible futures of the present world, given what has happened.

Here are some examples of temporal necessities:

\begin{exe}

\exr{he would still necessarily have lost the election}

\exs{Even if he hadn’t made the mistake, he would still \textbf{necessarily} have lost the election.}

\ex

\exs{This is not to imply that, had the Democrats allowed Bernie Sanders to emerge as the nominee, he would \textbf{necessarily} have won.}
\ecs{From COCA}

\ex\label{exe:I should be dead by tomorrow morning}

Adam's car broke down on a wild road in Siberia on a winter night. Fortunately, someone drove by and offered him a ride. Adam said to the kind person in the car:

\vspace{2pt}

\exs{I \textbf{should} be dead by tomorrow morning, and you saved my life.}

\ex\label{exe:The flowers should bloom in a few days}

\exs{The flowers \textbf{should} bloom in a few days, but the unseasonably cold weather has slowed their growth.}

\end{exe}

\noindent Intuitively, in these examples, besides possible futures of the present world, we also consider possible futures of alternatives to the present world.

\subsubsection*{Does $p$ imply $\HN p$ or $\TN p$?}

There is a principle in the literature of temporal logic, called \emph{the principle of necessity of the past}~\cite{ohrstrom_future_2020}: \emph{if a proposition not concerning the future is true in the present world, then it is necessary}.

As mentioned, historical necessities are universal quantifications over domains of possible timelines passing through the present world. History is deterministic. Thus, what has happened is the case for all possible timelines passing through the present world. Consequently, the principle of necessity of the past holds for historical necessities.

The principle of necessity of the past does not hold for temporal necessities. The reason is that temporal necessities also concern timelines not passing through the present world. What follows are two counterexamples.

\begin{exe}

\ex\label{ex:My favorite number is necessarily 4}

Adam's favorite number is four because he was the fourth baby born in the fourth hospital on May fourth. It is odd to say the following:

\vspace{2pt} 

\exs{\# Adam's favorite number is \textbf{necessarily} four.}

\ex

Adam and Bob bought lottery tickets yesterday. Later, they exchanged their
tickets. The winning number has just been revealed, and Adam wins. The following sentence is strange:

\vspace{2pt} 

\exs{\# Adam \textbf{should} be the winner.}
 
\end{exe}

\noindent Here are two more counterexamples:

\begin{exe}

\ex\label{exe:He lost the election yesterday, necessarily}

\exs{Over the past few months, this candidate has made numerous disrespectful remarks towards vulnerable groups. In view of his conduct and the circumstances of this period, his defeat in the election was \textbf{necessary}.}

\ex\label{exe:The river has flooded its banks, as it should}

\exs{It rained too much recently. The river has flooded its banks, as it \textbf{should}.}

\end{exe}

\noindent In the example \ref{exe:He lost the election yesterday, necessarily}, it is true that he lost the election, but it is still felicitous to say that it was necessary for him to lose the election. This means that ``necessary'' has meaning contributions. Similarly, ``should'' has meaning contributions in the example \ref{exe:The river has flooded its banks, as it should}.

Given that the domain of a historical necessity is not empty, the converse of the principle of necessity holds for the historical necessity: \emph{if a proposition not concerning the future is not true in the present world, then it is not necessary}. The arguments go as follows. Assume that a proposition not involving the future is not true. Since history is deterministic, it is not true for all possible timelines passing through the present world. Then, it is not a historical necessity, given that its domain is not empty.
Consequently, it is not felicitous to utter $\HN p$, where $p$ does not involve the future. This is why historical necessities typically concern the future.

\newcommand{\N}{\mathsf{N}\hspace{1pt}}

\subsubsection*{Is $\HN \phi$ or $\TN \phi$ preserved under the temporal shift to the past?}

Let $t$ and $t'$ be two moments, and $\phi$ and $\phi'$ be two sentences. We say $\phi'$ is a \Fdefs{temporal transformation} of $\phi$ from $t$ to $t'$ if $\phi'$ is the result of changing the temporal reference of $\phi$ by the difference between $t$ and $t'$. Here are some examples:

\begin{exe}

\ex

\begin{xlist}

\ex

Now: \exs{there is a sea battle.}

\ex 

Yesterday: \exs{there will be a sea battle tomorrow.}

\ex

Tomorrow: \exs{there was a sea battle yesterday.}

\end{xlist}

\ex

\begin{xlist}

\ex

Now: \exs{there will be a sea battle tomorrow.}

\ex 

Yesterday: \exs{there will be a sea battle the day after tomorrow.}

\ex

Tomorrow: \exs{there is a sea battle.}

\end{xlist}

\end{exe}

\noindent For each of the two groups, the three sentences in it are temporal transformations of each other with respect to the respective moments.

Temporal necessities are preserved under the temporal shift to the past: \emph{if $\TN \phi$ is true at a moment $t$, then for any earlier moment $t'$, $\TN \phi'$ is true at $t'$, where $\phi'$ is the temporal transformation of $\phi$ from $t$ to $t'$.} The reason is that when moving to the past, the timelines that temporal necessities concern are invariant. Look at the following examples:

\begin{exe}

\ex\label{exe:he will necessarily miss the deadline}

Adam hasn't done anything since taking over this project a few weeks ago. Bob tells Carol that Adam did not work on his project yesterday. Carol says:

\vspace{2pt}

\exs{Regardless of whether he worked on the project yesterday or not, he will \textbf{necessarily} miss the deadline tomorrow.}

\exr{exe:I should be dead by tomorrow morning}

Adam's car broke down on a wild road in Siberia on a winter night. Fortunately, someone drove by and offered him a ride. Adam said to the kind person in the car:

\vspace{2pt}

\exs{I \textbf{should} be dead by tomorrow morning, and you saved my life.}

\end{exe}

\noindent Consider the first example. Assume \emph{it were yesterday now}. It seems fine for Carol to say ``Adam will necessarily miss the deadline the day after tomorrow''. Consider the second example. Assume \emph{now it were the moment just before the person drove by}. It seems fine for Adam to say ``I should be dead by tomorrow morning''.

Historical necessities are not preserved under the temporal shift to the past. The reason is that when moving to the past, the timelines that historical necessities concern tend to be more. Look at the following examples:

\begin{exe}

\exr{exe:Those flowers must be dead in a few days}

\exs{When Adam left home yesterday, he forgot to move the flowers indoors. Given the current conditions, all of those flowers will \textbf{necessarily} die within a few days. }

\exr{exe:We should have arrived before sunset}

\exs{This road is usually bustling, but the traffic is good today. We \textbf{should} have arrived before sunset.}

\end{exe}

\noindent Consider the first example. Assume \emph{now it were a moment before Adam left home}. It seems odd to say ``All of those flowers will necessarily die within a few days''. Consider the second example. Assume \emph{now it were very early in the morning}. Given that this road is usually bustling, it seems odd to say ``We should have arrived before sunset''.

\paragraph{Remarks}

It seems that no work in the literature has explicitly discussed the distinctions between historical and temporal necessities. However, some literature has discussed some special kinds of historical or temporal necessities, which will be discussed in Section \ref{section: Strong/weak historical/temporal necessities}.

Whether ``must/necessary/should/ought to'' expresses a historical necessity or a temporal necessity is determined by context.

Temporal necessities concern possible timelines from some point in the past, which is determined by contexts. How do contexts determine the point in the past? This needs investigation.

Can historical necessities be viewed as special temporal necessities that concern only possible timelines passing through the present world? This also needs studies.

It seems that temporal necessities can be expressed by historical necessities with some other things: \emph{$\phi$ is a temporal necessity now} seems equivalent to \emph{it was the case at the starting moment of the context that $\phi'$ was a historical necessity}, where $\phi'$ is the temporal transformation of $\phi$ from the present moment to the starting moment of the context.

\newcommand{\HP}{\mathsf{HP}\hspace{1pt}}
\newcommand{\TP}{\mathsf{TP}\hspace{1pt}}

\section{Historical/temporal possibilities}
\label{section: Historical/temporal possibilities}

\Fdefs{Historical possibility} is over temporal alternatives, concerning what the world was like in the past, is like in the present, or will possibly be like in the future, \emph{given what has happened}. Typically, they concern the future.
\Fdefs{Temporal possibility} is over temporal alternatives, concerning what the world was like or could be like in the past, is like or could be like in the present, or will possibly be like or would possibly be like in the future. 
Their difference lies in that the former only concerns possible timelines passing through the present world, while the latter concerns possible timelines from some point in the past, which is determined by context.
In the sequel, we use $\HP \phi$ and $\TP \phi$ to respectively indicate \emph{$\phi$ is a historical possibility} and \emph{$\phi$ is a temporal necessity}.

What follows are some examples of historical possibility:

\begin{exe}

\exr{exe:He might be discharged from the hospital soon}

\exs{His blood pressure has been high these days, but it came down today. He \textbf{might} be discharged from the hospital soon.}

\ex\label{exe:We might be late for the meeting}

\exs{It is not rush hour now, but the traffic conditions are still not good. We \textbf{might} be late for the meeting.}

\ex\label{exe:??}

\exs{In the past few days, the weather has been warm, but it is very cold tonight. The stray cat outside \textbf{might} be dead tomorrow.}

\ex\label{exe:The river might overflow its banks tomorrow}

\exs{It rarely rains in this place, but today's rain is very heavy. The river \textbf{might} overflow its banks tomorrow.}

\ex\label{exe:They could be calling with the job offer soon}

\exs{They \textbf{could} be calling with the job offer soon, but I do not expect it after not hearing back for weeks.}

\ex\label{exe:??}

\exs{My tea \textbf{could} arrive tomorrow, but I do not expect it after the shipping delays.}

\ex\label{exe:??}

\exs{In 1932 it was \textbf{possible} for Great Britain to avoid war with Germany; but in 1937 it was \textbf{impossible}.}
\ecs{From \cite{thomason_combinations_1984}}

\end{exe}

\noindent Intuitively, in these examples, except for the last one, ``might/could'' concerns possible futures of the present world. In the last example, ``possible'' concerns possible futures of the world in 1932.

Some examples of temporal possibility follow:

\begin{exe}

\exr{exe:We could have done that years ago}

\exs{We \textbf{could} have done that years ago and saved thousands of American lives.}
\ecs{From COCA}

\ex\label{exe:He might have won the game}

\exs{He \textbf{might} have won the game, but he didn't in the end.}
\ecs{Adapted from \cite{portner_modality_2009}}

\ex\label{exe:We could have avoided being here}

\exs{We \textbf{could} have avoided being here, but here we are.}
\ecs{From COCA}

\ex\label{exe:I could be relaxing at home right now}

\exs{I \textbf{could} be relaxing at home right now.}

\ex\label{exe:I could be dead in a week}

Adam contracted a severe respiratory virus and was admitted to a hospital. Fortunately, a newly developed miracle drug just hit the market. After taking the medication for one day, Adam's condition stabilized. Adam said to his son:

\vspace{2pt}

\exs{I \textbf{could} be dead in a week.}

\ex\label{exe:I could be dead in a few days}

While climbing a snowy mountain, Adam encountered an avalanche and got trapped in a small valley. He had very little food. Fortunately, he discovered a stash of canned goods, which was enough to sustain him until the snow melted. He wrote in his journal:

\vspace{2pt}

\exs{I \textbf{could} be dead in a few days.}

\end{exe}

\noindent Intuitively, in the first two examples, ``could/might $\phi$'' means that $\phi$ \emph{was/is/will be} the case for some possible timeline of alternatives of the world.

\subsubsection*{Does $\HP p$ or $\TP p$ imply $p$?}

We call the following \emph{the principle of possibility of the past}: \emph{if a proposition not concerning the future is possible, then it is true}.

This principle holds for historical possibility, that is an existential quantification over domains of possible timelines passing through the present world. The arguments go as follows. History is deterministic. Then, if a proposition not involving the future is false, then it is false for all possible timelines passing through the present world. Thus, if a proposition is true for some possible timeline passing through the present world, then it is true.

The principle of possibility of the past does not hold for temporal possibility. The reason is that temporal possibility might concern possible timelines not passing through the present world. All the examples \ref{exe:We could have done that years ago}, \ref{exe:He might have won the game}, \ref{exe:We could have avoided being here}, and \ref{exe:I could be relaxing at home right now} are counterexamples.

It is easy to see that the converse of the principle of possibility of the past holds for historical possibility, given that its domain is not empty. Consequently, historical possibility is redundant for propositions not involving the future.

\subsubsection*{Is $\HP \phi$ or $\TP \phi$ is preserved under the temporal shift to the future?}

Temporal possibility is preserved under the temporal shift to the future: \emph{if $\TP \phi$ is true at a moment $t$, then for any later moment $t'$, $\TP \phi'$ is true at $t'$, where $\phi'$ is the temporal transformation of $\phi'$ from $t$ to $t'$.} The reason is that when moving to the future, the timelines that temporal possibility concerns are invariant. Look at the following examples:

\begin{exe}

\exr{exe:We could have done that years ago}

\exs{We \textbf{could} have done that years ago and saved thousands of American lives.}
\ecs{From COCA}

\exr{exe:I could be relaxing at home right now}
\exs{I \textbf{could} be relaxing at home right now.}

\end{exe}

\noindent Consider the first example. Assume \emph{now it were many years later}. It seems fine to say ``We could have done that many years ago and saved thousands of American lives''. Consider the second example. Assume \emph{now it were one day later}. It seems fine to say ``I could be relaxing at home yesterday''.

Historical possibility is not preserved under the temporal shift to the future. The reason is that when moving to the future, the timelines that historical possibility concerns tend to be less. Look at the following examples:

\begin{exe}

\exr{exe:He might be discharged from the hospital soon}

\exs{His blood pressure has been high these days, but it came down today. He \textbf{might} be discharged from the hospital soon.}

\exr{exe:We might be late for the meeting}
\exs{It is not rush hour now, but the traffic conditions are still not good. We \textbf{might} be late for the meeting.}

\end{exe}

\noindent Consider the first example. Assume that later his blood pressure gets high. Then, we cannot say ``He might be discharged from the hospital soon''. Consider the second example. Assume that soon the traffic conditions improve. Then we cannot say ``We might be late for the meeting.''

\subsubsection*{Can $\dec \will p$ coexist with $\HP \will \neg p$ or $\TP \will \neg p$?}

Intuitively, the declaration of $\will p$ seems quite strong and is not compatible with treating $\will \neg p$ as a historical possibility. For example, it seems weird to say ``The river will not overflow its banks tomorrow, but given the present situation, it might.''

The declaration of $\will p$ is compatible with treating $\will \neg p$ as a temporal possibility. For example, it is fine to say ``The river will not overflow its banks tomorrow, but it could.''

\section{Strong/weak historical/temporal necessities}
\label{section: Strong/weak historical/temporal necessities}

In this section, we distinguish strong and weak historical necessities, and distinguish strong and weak temporal necessities.

\newcommand{\XSHN}{\mathsf{SHN}\hspace{1pt}}

\subsection{Strong historical necessity}

Recall that historical necessities are over temporal alternatives, typically concerning what the world will possibly like in the future, \emph{given what has happened}.
\Fdefs{Strong historical necessity} is the historical necessity that can be expressed by ``necessary/must'', that has a similar meaning to \emph{inevitability}.
In the sequel, we use $\XSHN \phi$ to indicate \emph{$\phi$ is a strong historical necessity.}

What follows are some examples of strong historical necessity:

\begin{exe}

\exr{exe:There will necessarily be a sea battle tomorrow}

\exs{The two countries have been friendly for decades, but now our king doesn't believe that his daughter died accidentally in your country. There will \textbf{necessarily} be a sea battle tomorrow.}

\ex

\exs{This means resources have to be reallocated into other sectors to ensure a robust recovery, not simply a resumption of the old binge. But this will not \textbf{necessarily} be facilitated by ultra-low interest rates.}
\ecs{From COCA}

\ex\label{exe:He must be dead soon}

\exs{The lost driver in the desert is usually very careful, but he didn't bring any water this time. He \textbf{must} be dying soon.}

\exr{exe:He must lose}

\exs{He \textbf{must} lose. I think he knows he is going to lose because of all the executive orders he is shuffling out.}
\ecs{From COCA}

\end{exe}


\subsubsection*{Can $\XSHN \will p$ coexist with $\HP \will \neg p$?}

Intuitively, $\XSHN \will p$ seems very strong and to exclude the historical possibility of $\will \neg p$. For example, it seems strange to say, ``Those flowers must be dead in a few days, although they might survive.''

\subsubsection*{Can $\dec \will p$ coexist with $\XSHN \will \neg p$?}

Intuitively, $\dec \will p$ is stronger than $\HP \will p$. As discussed above, $\HP \will p$ cannot coexist with $\XSHN \will \neg p$. Consequently, $\dec \will p$ cannot coexist with $\XSHN \will \neg p$.
For example, it is weird to say, ``There will necessarily be a sea battle tomorrow, but there will not be.''

\subsubsection*{Present work on strong historical necessity}

Strong historical necessity is the notion of \emph{historical necessity} in the sense of Thomason \cite{thomason_combinations_1984}. It figures in Aristotle's discussion of the Sea Battle Puzzle in his \emph{On Interpretation} and frequently arises in debates about free will.
It is one of the most important notions of necessity, and there has been a lot of work about it in the literature, such as Prior \cite{prior_past_1967} and {\AA}qvist \cite{aqvist_logic_1999}.

\newcommand{\XWHN}{\mathsf{WHN}\hspace{1pt}}

\subsection{Weak historical necessity}

\Fdefs{Weak historical necessity} is the historical necessity that can be expressed by ``should/ought to''.
In the sequel, we use $\XWHN \phi$ to indicate \emph{$\phi$ is a weak historical necessity}.

Here are some examples of weak historical necessity:

\begin{exe}

\ex\label{ex:He should lose the election}

\exs{Our candidate has a good public image, but today, someone exposed a scandal about him. He \textbf{should} lose the election.}

\ex\label{exe:??}

\exs{He didn't notice the fallen rocks on the road and crashed into them. His injuries are severe. He \textbf{ought to} be dead soon.}

\exr{exe:We should have arrived before sunset}

\exs{This road is usually bustling, but the traffic is good today. We \textbf{should} have arrived before sunset.}

\ex\label{exe:??}

\exs{This area usually gets a lot of rain in the summer, but this year has been unusually dry. The entire forest \textbf{should} be dead in a few weeks.}

\end{exe}


\subsubsection*{Can $\XWHN \will p$ coexist with $\HP \will \neg p$?}

The answer is yes. For instance, it is fine to say, ``He should lose the election, although he might win.''

\subsubsection*{Can $\dec \will p$ coexist with $\XWHN \will \neg p$?}

In regular situations, $\dec \will p$ does not coexist with $\XWHN \will \neg p$. For example, it seems strange to say ``Given the present situation, he should lose the election, although he will not''. Intuitively, $\dec \will p$ is stronger than $\XWHN \will p$. For example, ``it will rain soon'' seems stronger than ``it should rain soon''. Suppose both $\dec \will p$ and $\XWHN \will \neg p$ are true in a situation. Then both $\XWHN \will p$ and $\XWHN \will \neg p$ are true in the situation, which would be special.

\newcommand{\XSTN}{\mathsf{STN}\hspace{1pt}}

\subsection{Strong temporal necessity}

Recall that temporal necessities are over temporal alternatives, concerning what the world was like or could be like in the past, is like or could be like in the present, or will possibly be like or would possibly be like in the future. 
\Fdefs{Strong temporal necessity} is the temporal necessity that can be expressed by ``necessary/must''.
In the sequel, we use $\XSTN \phi$ to indicate \emph{$\phi$ is a strong temporal necessity.}

We consider some examples of strong temporal necessity:

\begin{exe}

\exr{exe:He lost the election yesterday, necessarily}

\exs{Over the past few months, this candidate has made numerous disrespectful remarks towards vulnerable groups. In view of his conduct and the circumstances of this period, his defeat in the election was \textbf{necessary}.} 

\exr{he would still necessarily have lost the election}

\exs{Even if he hadn’t made the mistake, he would still \textbf{necessarily} have lost the election.}

\ex

\exs{I don't think that if I had been adopted by a different family that things would have \textbf{necessarily} been easier.}
\ecs{From COCA}

\ex

\exs{That is not to say that the outcome would \textbf{necessarily} have been the same had the challenge been phrased or put in a different form as a service agreement in GATS or even under NAFTA.}
\ecs{From COCA}

\exr{exe:he will necessarily miss the deadline}

Adam hasn't done anything since taking over this project a few weeks ago. Bob tells Carol that Adam did not work on his project yesterday. Carol says:

\vspace{2pt}

\exs{Regardless of whether he worked on the project yesterday or not, he will \textbf{necessarily} miss the deadline tomorrow.}

\end{exe}

%
%
%
%
%

\subsubsection*{Can $\XSTN p$ coexist with $\TP \neg p$?}

Intuitively, $\XSTN p$ seems very strong and excludes the temporal possibility of $\neg p$. For example, it is odd to say ``He necessarily lost the election yesterday, although he could win.''

\subsubsection*{Can $\XSTN \will p$ coexist with $\TP \will \neg p$?}

Again, intuitively, $\XSTN \will p$ seems very strong and excludes the temporal possibility of $\will \neg p$. For example, it seems strange to say ``No matter how hard we’ve trained, we will necessarily have a challenging game tomorrow, although we could have an easy game tomorrow.''

\subsubsection*{Can $p$ coexist with $\XSTN \neg p$?}

Intuitively, $p$ does not coexist with $\XSTN \neg p$. For example, it is very weird to say, ``He necessarily lost the election yesterday, although he won.'' In fact, the answer should be no, since $p$ implies $\TP p$, which does not coexist with $\XSTN \neg p$.

\subsubsection*{Can $\dec \will p$ coexist with $\XSTN \will \neg p$?}

Intuitively, the answer is no. For example, it is strange to say ``Regardless of whether he worked on the project yesterday or not, he will necessarily miss the deadline, although he will not miss it.'' Again, the answer should be no, since $\dec \will p$ is stronger than $\TP \will p$, which does not coexist with $\XSTN \will \neg p$.

\newcommand{\XWTN}{\mathsf{WTN}\hspace{1pt}}

\subsection{Weak temporal necessity}

\Fdefs{Weak temporal necessity} is the temporal necessity that can be expressed by ``should/ought to''.
In the sequel, we use $\XWTN \phi$ to indicate \emph{$\phi$ is a weak temporal necessity.}

Here are some examples of weak temporal necessity:

\begin{exe}

\ex\label{exe:The flowers should have perfumed the room}

\exs{The flowers \textbf{should} have perfumed the room, but for some reason they did not.}
\ecs{From COCA}

\ex\label{exe:he should have died not once but twice}

\exs{My miracle is my son: he \textbf{should} have died not once but twice.}
\ecs{From COCA}

\exr{ex:I ought to be dead right now}
Suppose Jones is in a building when an earthquake hits. The building collapses. Luckily, nothing falls upon Jones and he emerges from the rubble as the only survivor. Talking to the media, Jones says the following:

\vspace{2pt}

\exs{I \textbf{ought to} be dead right now.}
\ecs{From \cite{yalcin_modalities_2016}}

\ex\label{exe:It should be cold by now, but it isn't}

\exs{I put the beer into the refrigerator one hour ago. It \textbf{should} be cold by now, but it isn't.}
\ecs{Adapted from \cite{copley_what_2006}}

\exr{exe:I should be dead by tomorrow morning}
Adam's car broke down on a wild road in Siberia on a winter night. Fortunately, someone drove by and offered him a ride. Adam said to the kind person in the car: 

\vspace{2pt}

\exs{I \textbf{should} be dead by tomorrow morning, and you saved my life.}

\exr{exe:The flowers should bloom in a few days}

\exs{The flowers \textbf{should} bloom in a few days, but the unseasonably cold weather has slowed their growth.}

\end{exe}

%
%
%
%

\subsubsection*{Can $\XWTN p$ coexist with $\TP \neg p$?}

The answer is yes. As discussed below, $\XWTN p$ can coexist with $\neg p$. Note $\neg p$ implies $\TP \neg p$. Consequently, $\XWTN p$ can coexist with $\TP \neg p$.

\subsubsection*{Can $\XWTN \will p$ coexist with $\TP \will \neg p$?}

The answer is yes. As discussed below, $\XWTN \will p$ can coexist with $\dec \neg p$. Note $\dec \neg p$ implies $\TP \will \neg p$. Consequently, $\XWTN \will p$ can coexist with $\TP \will \neg p$.

\subsubsection*{Can $p$ coexist with $\XSTN \neg p$?}

The answer is yes. See examples \ref{exe:The flowers should have perfumed the room}, \ref{exe:he should have died not once but twice}, \ref{ex:I ought to be dead right now}, and \ref{exe:It should be cold by now, but it isn't}.

\subsubsection*{Can $\dec \will p$ coexist with $\XSTN \will \neg p$?}

The answer is yes. For example, in example \ref{exe:I should be dead by tomorrow morning}, it is fine to say ``I should be dead by tomorrow morning, but I will not'', and in example \ref{exe:The flowers should bloom in a few days}, it is fine to say ``The flowers should bloom in a few days, but they will not''.

\subsubsection*{Present work on weak temporal necessity}

There has been some work on weak temporal necessity in the literature. Here we briefly mention some of them.

Copley \cite{copley_what_2006} thinks that ``should'' can express a \emph{metaphysical} necessity. What follows is an example of metaphysical necessity from her:

\begin{exe}

\ex\label{exe:It should have rained}

Yesterday evening, you saw that the clouds had been building up and that a thunderstorm had been approaching. This morning, before opening the curtain, you say the following:

\vspace{2pt}

\exs{It \textbf{should} have rained.}

\end{exe}

\noindent Copley thinks that metaphysical ``should'' is concerned with the evidence of facts and the reasoning from earlier causing facts to later caused facts. We think that metaphysical necessity expressed by ``should'' is weak temporal necessity.

Thomson \cite{thomson_normativity_2008} thinks that metaphysical necessity expressed by ``should'' can be handled by using \emph{time} and \emph{probability}: \emph{It should be the case that $\phi$} if and only if \emph{it was probably the case that $\phi$}.

Yalcin \cite{yalcin_modalities_2016} considers this idea natural but incorrect. What follows is a counterexample from him.

\begin{exe}

\ex 

An urn has one black and four white marbles. A marble is selected at random. We observe that it is black. Then we can say:
\exs{It was \textbf{probable} that the marble selected would be white.} 
However, it would be odd to say: \exs{The marble selected \textbf{should} be white.}

\end{exe}

Yalcin \cite{yalcin_modalities_2016} thinks that metaphysical necessity expressed by ``should'' is closely related to \emph{normality}\footnote{Yalcin \cite{yalcin_modalities_2016} used the notion \emph{pseudo-epistemic} necessity, instead of metaphysical necessity.}: it quantifies over the domain of normal worlds.
Based on this idea, he presented a formal analysis of metaphysical necessity expressed by ``should'', which is based on Veltman \cite{veltman_defaults_1996} and Yalcin \cite{yalcin_epistemic_2007}.

\subsection{Conditional strong/weak historical/temporal necessities}

Conditional strong/weak historical/temporal necessities are very common. Here, we just mention one difference between conditional weak historical/temporal necessities and conditional strong historical/temporal necessities: the former is nonmonotonic, while the latter seems monotonic.

The following two examples are for nonmonotonicity of conditional weak historical necessity:

\begin{exe}

\ex\label{ex:}

\exs{Our candidate has a good public image, but today, someone exposed a scandal about him. He \textbf{should} lose the election. However, if someone can expose a scandal about his opponent, he might win.}

\ex\label{exe:??}

\exs{He didn't notice the fallen rocks on the road and crashed into them. His injuries are severe. He \textbf{ought to} be dead soon. However, if a helicopter can rescue him soon, he might survive.}

\end{exe}

What follows are two examples of nonmonotonicity of conditional weak temporal necessity:

\begin{exe}

\ex\label{exe:}

\exs{If his mother had taken a tram, she \textbf{should} be home by now; however, if she had taken a tram but the strikers had gathered in a different street, she could still be on the way.}

\ex\label{exe:}

\exs{If Adam had not looked at his phone, there \textbf{should} be no accident now; however, if Adam had not looked at his phone but the car in front had reversed, there could still be an accident now.}

\end{exe}

It seems a bit hard to show that strong historical/temporal necessities are monotonic. But intuitively, strong historical/temporal necessities are too strong to be \emph{defeasible}.

The following two examples illustrate the monotonicity of strong historical necessities:

\begin{exe}

\ex\label{exe:}

\exs{\# Given that the only flight has been canceled, they \textbf{must} be late for the meeting, but if they take a train, they might make it on time.}

\ex\label{exe:}

\exs{\# Given what happened, there will \textbf{necessarily} be a sea battle tomorrow, but if one country compromises, the battle might be avoided.}

\end{exe}

The following two examples illustrate the monotonicity of strong temporal necessities:

\begin{exe}

\ex

\exs{\# Over the past few months, this candidate has made numerous disrespectful remarks towards vulnerable groups. He lost the election yesterday, as was \textbf{necessary}, but if he had apologized before the election, the result might have been different.}

\ex

\exs{\# There was heavy rainfall yesterday, yet the government took no action. So, \textbf{necessarily}, the street has been flooded. However, if the residents had placed sandbags at the entrance this morning, the street might be fine now.}

\end{exe}

\paragraph{Remarks}

In the above, we distinguish strong historical/temporal necessities and weak historical/temporal necessities. One may wonder whether there are strong and weak historical/temporal possibilities. This is an interesting question, but we have to leave it for future investigation.

\section{Summing up the discussed features of historical/tempo-ral necessities/possibilities}
\label{section: Putting everything together}

\paragraph{Duals}

Strong historical necessity (but not weak historical necessity) and historical possibility are duals of each other, and strong temporal necessity (but not weak temporal necessity) and temporal possibility are duals of each other.

\paragraph{Monotonicity}

Strong historical necessity and strong temporal necessity are monotonic, but weak historical necessity and weak temporal necessity are nonmonotonic.

\paragraph{Implications}

\begin{enumerate}[label=(\arabic*),leftmargin=3.33em]

\item Strong historical necessity is stronger than weak historical necessity, and strong temporal necessity is stronger than weak temporal necessity.

\item The principle of necessity of the past holds for strong/weak historical necessities but does not hold for strong/weak temporal necessities.

\item

The principle of possibility of the past applies to historical possibility but does not apply to temporal possibility.

\item 

Temporal necessities are preserved under the temporal shift to the past, but historical necessities are not.

\item 

Temporal possibility is preserved under the temporal shift to the future, but historical possibility is not.

\end{enumerate}

\paragraph{Coexistence}

See the tables \ref{tab:placeholder0}, \ref{tab:placeholder1} and \ref{tab:placeholder2}.

\begin{table}[H]

\centering

\begin{tabular}{|c|c|c|}
\hline
& $\HP \will p$ & $\TP \will p$ \\
\hline
$\dec \will \neg p$ & not co-existent & co-existent \\
\hline
\end{tabular}

\caption{Some features of historical/temporal possibilities}
\label{tab:placeholder0}

\end{table}

\begin{table}[H]

\centering

\begin{tabular}{|c|c|c|}
\hline
 & $\XSHN \will p$ & $\XWHN \will p$ \\
\hline
$\HP \will \neg p$ & not co-existent & co-existent \\
\hline
$\dec \will \neg p$ & not co-existent & trivially co-existent \\
\hline
\end{tabular}

\medskip

\caption{Some features of strong/weak historical necessities}
\label{tab:placeholder1}

\end{table}

\begin{table}[H]

\centering

\begin{tabular}{|c|c|c|}
\hline
& $\XSTN p$ & $\XWTN p$ \\
\hline
$\TP \neg p$ & not co-existent & co-existent \\
\hline
$\neg p$ & not co-existent & co-existent \\
\hline
\end{tabular}

\medskip

\begin{tabular}{|c|c|c|}
\hline
 & $\XSTN \will p$ & $\XWTN \will p$ \\
\hline
$\TP \will \neg p$ & not co-existent & co-existent \\
\hline
$\dec \will \neg p$ & not co-existent & co-existent \\
\hline
\end{tabular}

\caption{Some features of strong/weak temporal necessities}

\label{tab:placeholder2}

\end{table}

\part{}

\textit{%
In this part, we state our approach to the four notions of necessity and the two notions of possibility, and present a logical theory of them.
}

\section{Our approach to historical/temporal necessities/possi-bilities based on ontic systems}
\label{section: Our approach to strong/weak historical/temporal necessities based on ontic systems}

Generally speaking, our approach to the six notions is as follows.
There is an agent who accepts a system of ontic rules governing how the world evolves over time from some point in the past, in which some ontic rules are undefeatable. The system as a whole determines which timelines are \emph{expected}. Undefeatable ontic rules determine which timelines are \emph{accepted}.
The domains of strong and weak historical necessities, respectively, consist of accepted and expected timelines passing through the present moment, and historical possibility is the dual of strong historical necessity.
The domains of strong and weak temporal necessities, respectively, consist of accepted and expected timelines, and temporal possibility is the dual of strong temporal necessity.

\subsection{Ontic systems}

There is an agent. The agent accepts some \emph{ontic rules} concerning how the world evolves from some point in the past.
Ontic rules can be of many types: natural laws (\emph{Light is faster than sound}), common natural phenomena (\emph{It is cold in Beijing during the winter}), daily experiences (\emph{Lack of engine oil causes damage to engines}), etc.

Conflicts may arise among ontic rules. Consider an example. Suppose that it is winter and the El Ni\~{n}o condition has occurred. Then, the following ontic rules are conflicting: \emph{It is cold in Beijing in winter}, and \emph{the El Ni\~{n}o condition causes a warm winter in Beijing}.

There is an \emph{order} among ontic rules for resolving conflicts. The order results from many factors. Here, we mention two of them. First, some types of rules tend to override other kinds of rules. For example, natural laws tend to override daily experiences. Second, special rules tend to override general rules. For example, the rule \emph{the El Ni\~{n}o condition causes a warm winter in Beijing} tends to override the rule \emph{it is cold in Beijing in winter}.

The agent might learn new ontic rules and is ready for some new rules to defeat some old ones. However, she might take some rules as \emph{undefeatable}. For example, she might consider the following rule undefeatable: \emph{The sun rises tomorrow}.

The set of ontic rules, the order among ontic rules, and the set of undefeatable ontic rules form an \emph{ontic system}.

The ontic system as a whole determines which timelines are \emph{expected}.
What has happened may not be what was expected. However, the agent can often live with unexpected things without changing her ontic system.

The set of undefeatable ontic rules determines which timelines are \emph{accepted}. Expected timelines are accepted, but not the other way around. The agent cannot live with unaccepted things without changing her ontic system.

\paragraph{Remarks}

The notion ``ontic rules'' may be misleading in the following sense: ontic rules are agent-independent. In this work, they are agent-dependent.

It is not always clear what ontic systems we accept. Additionally, different individuals may have distinct ontic systems.

\subsection{About strong historical necessity and historical possibility}

Strong historical necessity is a universal quantification over the domain of accepted timelines passing through the present moment, and historical possibility is an existential quantification over the same domain.

We look at two examples mentioned above:

\begin{exe}

\exr{exe:He must be dead soon}

\exs{The lost driver in the desert is usually very careful, but he didn't bring any water this time. He \textbf{must} be dying soon.}

\exr{exe:They could be calling with the job offer soon}

\exs{They \textbf{could} be calling with the job offer soon, but I do not believe it after not hearing back for weeks.}

\end{exe}

\noindent The understanding of the first example can be as follows. The speaker accepts an ontic rule: \emph{A man without much water in the desert cannot live long.} The speaker treats the rule as undefeatable. Considering what has happened, that is, the lost driver did not take any water, by the rule, the lost driver will die soon.
The understanding of the second example can be as follows. The possibility that the speaker will receive an offer soon is not excluded by any undefeatable ontic rule, although it is excluded by some defeasible rules.

\subsection{About weak historical necessity}

Weak historical necessity is a universal quantification over the domain of expected timelines passing through the present moment.

Consider two examples mentioned above:

\begin{exe}

\exr{ex:He should lose the election}

\exs{Our candidate has a good public image, but today, someone exposed a scandal about him. He \textbf{should} lose the election.}

\exr{exe:We should have arrived before sunset}
\exs{This road is usually bustling, but the traffic is good today. We \textbf{should} have arrived before sunset.}

\end{exe}

\noindent The understanding of the first example can be as follows. The speaker accepts an ontic rule: \emph{A candidate with a scandal will lose the election.} The speaker does not regard the rule as undefeatable. Considering what has happened, namely that the candidate has been involved in a scandal, by the rule, the candidate will lose the election. The speaker still considers it possible that the candidate will not lose.
The understanding of the second example can be as follows. The speaker accepts an ontic rule: \emph{Good traffic lets us arrive before sunset}. The speaker does not treat the rule as undefeatable. Considering what has happened, that is, the traffic is good, by the rule, we will arrive before sunset. The speaker still considers it possible that they will not arrive before sunset.

\subsection{About strong temporal necessity and temporal possibility}

Strong temporal necessity is a universal quantification over the domain of accepted timelines passing through some point in the past. Temporal possibility is an existential quantification over the same domain.

Here are two examples.

\begin{exe}

\ex\label{exe:Now the streets are flooded, necessarily}

\exs{There was heavy rainfall yesterday, yet the government took no action. So, \textbf{necessarily}, the streets are flooded now.}

\exr{exe:He might have won the game}
\exs{He \textbf{might} have won the game, but he didn't in the end.}
\ecs{Adapted from \cite{portner_modality_2009}}

\end{exe}

\noindent The understanding of the first example can be as follows. The speaker accepts an ontic rule: \emph{The streets will be flooded with the government's omission}. The speaker treats the rule as undefeatable. By the rule, it is the case that the streets are flooded. 
The understanding of the second example can be as follows. The possibility that he won the game is not excluded by any undefeatable ontic rule.

\subsection{About weak temporal necessity}

Weak temporal necessity is a universal quantification over the domain of expected timelines.

Here are two examples.

\begin{exe}

\exr{ex:I ought to be dead right now}

Suppose Jones is in a building when an earthquake hits. The building collapses. Luckily, nothing falls upon Jones and he emerges from the rubble as the only survivor. Talking to the media, Jones says the following:

\vspace{2pt}

\exs{I \textbf{ought to} be dead right now.}
\ecs{From \cite{yalcin_modalities_2016}}

\exr{exe:I should be dead by tomorrow morning}

Adam's car broke down on a wild road in Siberia on a winter night. Fortunately, someone drove by and offered him a ride. Adam said to the kind person in the car: 

\vspace{2pt}

\exs{I \textbf{should} be dead by tomorrow morning, and you saved my life.}

\end{exe}

\noindent The understanding of the first example can be as follows. Jones accepts an ontic rule: \emph{people do not survive in collapsing buildings}. By this rule, he is dead. Jones does not treat the rule as undefeatable.
The understanding of the third example can be as follows. Adam accepts an ontic rule: \emph{people will be dead on winter nights in Siberia without a heating source}. By this rule, he will be dead tomorrow morning. Jones does not treat the rule as undefeatable.

\subsection{About conditional strong/weak historical/temporal necessities}

There is a special set of timelines, called a \Fdefs{hypothesis set}. Propositions can \emph{update} the hypothesis set in a way. Ontic systems and hypothesis sets together determine accepted timelines and expected timelines.

For ontic systems and hypothesis sets together determining accepted timelines, what follows is always the case: when updating a hypothesis set with a proposition, we will not obtain more accepted timelines.

For ontic systems and hypothesis sets together determining accepted timelines, the following can happen: when updating a hypothesis set with a proposition, we obtain more expected timelines.

Conditional strong temporal (historical) necessity with the antecedent $\phi$ is a universal quantification over the set of accepted timelines (passing through the present world), determined by the ontic system and the resulting hypothesis set by updating the original hypothesis set with $\phi$.

Conditional weak temporal (historical) necessity with the antecedent $\phi$ is a universal quantification over the set of expected timelines (passing through the present world), determined by the ontic system and the resulting hypothesis set by updating the original hypothesis set with $\phi$.

This approach to conditional necessity is closely connected to the update semantics for conditionals proposed by Veltman \cite{veltman_making_2005}: when evaluating a conditional, we first update something with its antecedent and then evaluate its consequent with respect to the update result.

\subsection{About will-sentences and their declaration}

Will-sentences $\will p$ have truth values only when they are evaluated with respect to specific timelines.

As mentioned in Section \ref{section: Historical/temporal possibilities}, $\dec \will p$ cannot coexist with $\HP \will \neg p$. This means that $\dec \will p$ implies $\XSHN \will p$. The argument goes as follows.
Since $\dec \will p$ cannot coexist with $\HP \will \neg p$, $\dec \will p$ implies $\neg \HP \will \neg p$. $\HP$ is the dual of $\XSHN$. Thus, $\neg \HP \will \neg p$ is equivalent to $\XSHN \neg \will \neg p$. Note $\neg \will \neg p$ is equivalent to $\will p$. Then, $\XSHN \neg \will \neg p$ is equivalent to $\XSHN \will p$. Thus, $\dec \will p$ implies $\XSHN \will p$.

Does $\XSHN \will p$ imply $\dec \will p$? More specifically, if an agent accepts $\XSHN \will p$, can she declare $\will p$? We think so. Suppose someone says, ``It must be the case tomorrow that $p$.'' Intuitively, she can say ``It will be the case tomorrow that $p$.''

To sum up, we think that $\dec \will p$ has the same truth value as $\XSHN \will p$.

\subsection{Explanations of discussed features of historical/temporal necessities/possibilities}

\paragraph{Duals}

\emph{Strong historical necessity (but not weak historical necessity) and historical possibility are duals of each other, and strong temporal necessity (but not weak temporal necessity) and temporal possibility are duals of each other.}

We view strong historical necessity and historical possibility as the universal and existential quantifiers, respectively, over the domain of accepted timelines passing through the present world. We view weak historical necessity as the universal quantifier over the domain of expected timelines passing through the present world, that is a subset of the domain of accepted timelines passing through the present world. Consequently, strong historical necessity (but not weak historical necessity) is the dual of historical possibility.

The situation for strong and weak temporal necessities and temporal possibility is similar; the only difference is that all accepted (expected) timelines are considered, not only those passing through the present world.

\paragraph{Monotonicity}

\emph{Strong historical necessity and strong temporal necessity are monotonic, but weak historical necessity and weak temporal necessity are nonmonotonic.}

This is a complicated issue. Simply speaking, this is related to the following two facts of how ontic systems and hypothesis sets determine accepted and expected timelines together: (1) when an ontic system and a hypothesis set determine a set of accepted timelines, the order among ontic rules does not play a role, and the ``bigger'' the hypothesis set is, the smaller the set of accepted timelines is; (2) however, when an ontic system and a hypothesis set determine a set of expected timelines, the order among ontic rules does play a role, and it might not be the case that the ``bigger'' the hypothesis set is, the smaller the set of accepted timelines is.

\paragraph{Implications}

\begin{enumerate}[label=(\arabic*),leftmargin=3.33em]

\item

\emph{Strong historical necessity is stronger than weak historical necessity, and strong temporal necessity is stronger than weak temporal necessity.}

The reason is that expected timelines are accepted timelines, but not the other way around.

\item 

\emph{The principle of necessity of the past holds for strong/weak historical necessities but does not hold for strong/weak temporal necessities.}

Please see the discussion in Section \ref{section: Historical/temporal necessities}.

\item 

\emph{The principle of possibility of the past holds for historical possibility but does not hold for temporal possibility.}

Please see the discussion in Section \ref{section: Historical/temporal possibilities}.

\item 

\emph{Strong and weak temporal necessities are preserved under the temporal shift to the past, but neither strong nor weak historical necessity is.}

Strong and weak temporal necessities concern accepted and expected timelines, respectively, which are the same at different moments. Therefore, temporal necessities are preserved under the temporal shift to the past.

Strong and weak historical necessities concern accepted and expected timelines passing through the present moment, respectively, which tend to become more when moving to the past. Therefore, neither strong nor weak historical necessity is preserved under the temporal shift to the past.

\item 

\emph{Temporal possibility is preserved under the temporal shift to the future, but historical possibility is not.}

Temporal possibility concerns accepted timelines, which are the same at different moments. Therefore, temporal possibility is preserved under the temporal shift to the future.

Historical possibility concerns accepted timelines passing through the present moment, which tend to become less when moving to the future. Therefore, historical possibility is not preserved under the temporal shift to the future.

\end{enumerate}

\paragraph{Coexistence}

See the tables \ref{tab:placeholder0p}, \ref{tab:placeholder1p} and \ref{tab:placeholder2p}.

\begin{table}[htb]

\centering

\begin{tabular}{|c|c|c|}
\hline
& $\HP \will p$ & $\TP \will p$ \\
\hline
$\dec \will \neg p$ & not co-existent $(1)$ & co-existent $(2)$ \\
\hline
\end{tabular}

\begin{itemize}

\item

Note: (a) $\dec \will \neg p$ means that $\will \neg p$ is true for all accepted timelines passing through the present world (there is at least one), (b) $\HP \will p$ indicates that $\will p$ is true for some accepted timeline passing through the present world, and (c) $\TP \will p$ indicates that $\will p$ is true for some accepted timeline, which might not pass through the present world. This explains $(1)$ and $(2)$.

\end{itemize}

\caption{Explanations of some features of historical/temporal possibilities}

\label{tab:placeholder0p}

\end{table}

\begin{table}[htb]

\centering

\begin{tabular}{|c|c|c|}
\hline
 & $\XSHN \will p$ & $\XWHN \will p$ \\
\hline
$\HP \will \neg p$ & not co-existent $(1)$ & co-existent $(2)$ \\
\hline
$\dec \will \neg p$ & not co-existent (3) & trivially co-existent (4) \\
\hline
\end{tabular}

\begin{itemize}

\item

$(1)$ can be explained by (a) $\XSHN$ and $\HP$ are duals of each other and (b) $\will p$ and $\will \neg p$ are negations of each other.

\item 

Note: (a) $\HP \will \neg p$ indicates that $\will \neg p$ is true for some accepted timeline passing through the present world; (b) $\XWHN \will p$ indicates that $\will p$ is true for all expected timelines passing through the present world. Accepted timelines passing through the present world might not be expected. Thus, $\HP \will \neg p$ can coexist with $\XWHN \will p$. This explains $(2)$.

\item

$(3)$ can be explained by the fact that $\dec \will \neg p$ is equivalent to $\XSHN \will \neg p$ and $\XSHN \will \neg p$ cannot coexist with $\XSHN \will p$.

\item 

Note: (a) $\dec \will \neg p$ indicates that $\will \neg p$ is true for all accepted timelines passing through the present world; (b) $\XWHN \will p$ indicates that $\will p$ is true for all expected timelines passing through the present world.
Also note: all expected timelines are accepted.

If there is an expected timeline passing through the present world, then $\dec \will \neg p$ and $\XWHN \will p$ cannot both be true. Suppose there is no expected timeline passing through the present world. In this case, $\XWHN \will p$ trivially holds, and $\dec \will \neg p$ can be true. This explains $(4)$.

\end{itemize}

\caption{Explanations of some features of strong/weak historical necessities}

\label{tab:placeholder1p}

\end{table}

\begin{table}[htb]

\centering

\begin{tabular}{|c|c|c|}
\hline
& $\XSTN p$ & $\XWTN p$ \\
\hline
$\TP \neg p$ & not co-existent $(1)$ & co-existent $(2)$ \\
\hline
$\neg p$ & not co-existent $(3)$ & co-existent $(4)$ \\
\hline
\end{tabular}

\medskip

\begin{tabular}{|c|c|c|}
\hline
 & $\XSTN \will p$ & $\XWTN \will p$ \\
\hline
$\TP \will \neg p$ & not co-existent $(5)$ & co-existent $(6)$ \\
\hline
$\dec \will \neg p$ & not co-existent $(7)$ & co-existent $(8)$ \\
\hline
\end{tabular}

\begin{itemize}

\item

$(1)$ and $(5)$ can be explained by that $\XSTN$ and $\TP$ are duals.

\item 

$(2)$ and $(6)$ can be explained by the fact that the domain of $\TP$, that is, the set of accepted timelines, might not be a subset of the domain of $\XWTN$, that is, the set of expected timelines.

\item 

Note: the domain of $\XSTN$ consists of all accepted timelines, which includes the set of all accepted timelines passing through the present world, which is not empty. Thus, $\XSTN p$ implies $p$. Thus, $\XSTN p$ cannot coexist with $\neg p$. This explains $(3)$.

\item 

$(4)$ and $(8)$ can be explained by the fact that there might be no expected timeline passing through the present world.

\item 

Note: $\XSTN \will p$ implies $\XSHN \will p$, which cannot coexist with $\dec \will \neg p$. Thus, $\XSTN \will p$ cannot coexist with $\dec \will \neg p$. This explains $(7)$.

\end{itemize}

\caption{Explanations of some features of strong/weak temporal necessities}

\label{tab:placeholder2p}

\end{table}

\section{A logical theory of historical/temporal necessities/possi-bilities in branching time}
\label{section: A logical theory for strong/weak historical/temporal necessities in branching time}

\newcommand{\FMF}{\mathsf{MF}}

\newcommand{\FIF}{\mathsf{IF}}

\newcommand{\SHN}{[\hspace{-2pt}[ \mathtt{H} ]\hspace{-2pt}]}
\newcommand{\SHP}{\langle\hspace{-3pt}\langle \mathtt{H} \rangle\hspace{-3pt}\rangle}

\newcommand{\WHN}{[\mathtt{H}]}
\newcommand{\WHP}{\langle\mathtt{H}\rangle}

\newcommand{\Froot}{\mathtt{beg}}

\subsection{Language}

\begin{definition}[The language $\Phi_{\SWHTN}$] \label{def:The language SWONT}

Let $\AP$ be a countable set of atomic propositions and $p$ range over it. The language $\Phi_{\SWHTN}$ of the Logic for Historical/Temporal Necessities/Possibilities in Branching Time ($\SWHTN$) is defined as follows:
\[\phi ::= \bot \mid p \mid \neg \phi \mid (\phi \land \phi) \mid \XXX \phi \mid \YYY \phi \mid \SHN \phi \mid \WHN \phi \mid \STN \phi \mid \WTN \phi\]

\end{definition}

The intuitive reading of the featured formulas of this language is as follows:
\begin{itemize}

\item $\XXX \phi$: \emph{$\phi$ will be true at the next moment.}

We have discussed the differences between will-sentences and declarations of them. Here, note that $\XXX \phi$ does not indicate the declaration of ``$\phi$ will be true at the next moment''. The meaning of \emph{moment} will be clear when the semantics is given.

\item $\YYY \phi$: \emph{if the last moment exists, $\phi$ was true at it.}

\item $\SHN \phi$: \emph{$\phi$ must be true at the present moment.}

$\SHN$ indicates strong historical necessity.

\item $\WHN \phi$: \emph{$\phi$ should be true at the present moment.}

$\WHN$ indicates weak historical necessity.

\item $\STN \phi$: \emph{$\phi$ must be true at the present instant.}

$\STN$ indicates strong temporal necessity. The meaning of \emph{instant} will be clear when the semantics is given.

\item $\WTN \phi$: \emph{$\phi$ should be true at the present instant.}

$\WTN$ indicates strong temporal necessity.

\end{itemize}

The propositional connectives $\top, \lor, \rightarrow$, and $\leftrightarrow$ are defined as usual. Here are some other derivative expressions:
\begin{itemize}

\item 

Define $\Froot$ as $\YYY \bot$, meaning: \emph{this is the beginning of the context}.

\item 

Define the dual $\YYY' \phi$ of $\YYY \phi$ as $\neg \YYY \neg \phi$, meaning: \emph{the last moment exists and $\phi$ was true at it.}

\item

Define the dual $\SHP \phi$ of $\SHN \phi$ as $\neg \SHN \neg \phi$, meaning: \emph{$\phi$ can be true at the present moment}. $\SHP$ indicates historical possibility.

\item

Define the dual $\WHP \phi$ of $\WHN \phi$ as $\neg \WHN \neg \phi$. It is unclear for us whether $\WHP \phi$ has a corresponding expression in natural language, and we introduce it for technical reasons.

\item

The dual $\STP \phi$ of $\STN \phi$ is defined as $\neg \STN \neg \phi$, meaning: \emph{$\phi$ could be true at the present instant}. $\STP$ indicates temporal possibility.

\item

Define the dual $\WTP \phi$ of $\WTN \phi$ as $\neg \WTN \neg \phi$. $\WTP \phi$ does not seem to have an intuitive reading, and we introduce it for technical reasons.

\end{itemize}

\subsection{Models}

\begin{definition}[Models for $\Phi_\SWHTN$] \label{def:Models for SWONT}

A tuple $\MM = (W, r, <, V)$ is a \defstyle{model} for $\Phi_\SWHTN$ if the following conditions are met:
\begin{itemize}

\item $W$ is a nonempty set of states;

\item $r$ is in $W$, called the \Fdefs{root};

\item $<$ is a serial relation on $W$ such that for every $w \in W$, there is a unique finite sequence $x_0, \dots, x_n$ of states such that $x_0 = r$, $x_n = w$ and $x_0 < \dots < x_n$;

\item $V: \mathsf{AP} \to \mathcal{P}(W)$ is a valuation.

\end{itemize}

\end{definition}

Models are based on the so-called \emph{serial discrete rooted trees}. Intuitively, models indicate how the world can evolve over time. There is a starting point but no ending point. The past is determined, but the future is open.

\begin{example}[Models] \label{example:Pointed models}

Figure \ref{fig:a pointed model} indicates a model.

\begin{figure}[htb]

\begin{center}

\begin{tikzpicture}[node distance=25mm,scale=0.8,every node/.style={transform shape}]
\tikzstyle{every state}=[draw=black,text=black,minimum size=11mm]

\node[state] (s-0-1) [] {$w^0_1$};
\node [above=6mm] at (s-0-1) {$p,q$};

\node[] (a) [above of=s-0-1] {};
\node[state] (s-1-1) [left of=a] {$w^1_1$};
\node [above=6mm] at (s-1-1) {$p$};

\node[state] (s-1-2) [right of=a] {$w^1_2$};
\node [above=6mm] at (s-1-2) {$q$};

\node[] (b) [above of=a] {};
\node[state] (s-2-2) [left of=b] {$w^2_2 \atop \{p\}$};
\node [above=6mm] at (s-2-2) {$p$};

\node[state] (s-2-1) [left of=s-2-2] {$w^2_1 \atop \{\}$};
\node [above=6mm] at (s-2-1) {};

\node[state] (s-2-3) [right of=b] {$w^2_3 \atop \{q\}$};
\node [above=6mm] at (s-2-3) {$q$};

\node[state] (s-2-4) [right of=s-2-3] {$w^2_4 \atop \{\}$};
\node [above=6mm] at (s-2-4) {};

\node [above=15mm] at (s-2-1) {$\vdots$};
\node [above=15mm] at (s-2-2) {$\vdots$};
\node [above=15mm] at (s-2-3) {$\vdots$};
\node [above=15mm] at (s-2-4) {$\vdots$};

\path
(s-0-1) edge [->,left] node {} (s-1-1)
(s-0-1) edge [->,left] node {} (s-1-2)

(s-1-1) edge [->,left] node {} (s-2-1)
(s-1-1) edge [->,left] node {} (s-2-2)

(s-1-2) edge [->,left] node {} (s-2-3)
(s-1-2) edge [->,left] node {} (s-2-4)
;

\end{tikzpicture}
\end{center}

\caption{A model} \label{fig:a pointed model}
\end{figure}

\end{example}

\newcommand{\msete}[2]{\mathbf{E}_{#2}(#1)}

\newcommand{\mseta}[2]{\mathbf{A}_{#2}(#1)}

\subsection{Contexts}

Let $\MM = (W, r, <, V)$ be a model and $x_0$ be a state of it. The following are some notions and notations that will be used later:
\begin{itemize}

\item

An infinite sequence $\pi = x_0, x_1, \dots$ of states is called a \Fdefs{timeline} from $x_0$ if $x_0 < x_1 < \dots$. Note that timelines might not start from the root $r$.
We use $\mathtt{TL} (\MM, x_0)$ to indicate the set of all timelines of $\MM$ from $x_0$.

\item

For any timeline $\pi$ from $x_0$ and state $w$, if $w$ is an element of $\pi$, we say $\pi$ \Fdefs{passes through} $w$. In the sequel, we use $w \in \pi$ to indicate $\pi$ passes through $w$.

\item 

For every state $w$, $w$ is called a \Fdefs{moment} with respect to $x_0$ if there is a timeline $\pi$ from $x_0$ passing through $w$.

\item

Let $\pi$ be a timeline from $x_0$ and $i$ be a natural number. We use $\pi[i]$ to indicate the $i+1$-th element of $\pi$. We call $i$ the \Fdefs{instant} of $\pi[i]$.
We use $\pi[i,\infty]$ to indicate the suffix of $\pi$ from $\pi[i]$. In the sequel, natural numbers are also called instants.

\end{itemize}

\begin{definition}[Contexts]
\label{def:Contexts}

Let $\MM$ be a model. A tuple $\CC = (z, \DDD, \succ, \UU)$ is called a \defstyle{context} for $\MM$ if the following conditions are met:
\begin{itemize}

\item 

$z$ is a state of $\MM$;

\item $\DDD$ is a finite (possibly empty) set of (possibly empty) sets of timelines of $\MM$ from $z$;

\emph{The elements of $\DDD$ are called \defstyle{ontic rules}. Intuitively, rules regulate how the world evolves. Note, we allow the \emph{absurd} rule $\emptyset$.}

\item $\succ$ is an irreflexive and transitive relation on $\DDD$;

\emph{Intuitively, $\DD_1 \succ \DD_2$ means that $\DD_1$ has higher \emph{priority} than $\DD_2$.}

\item $\UU$ is a (possibly empty) subset of the set of maximal elements of $\DDD$ with respect to $\succ$.

\emph{The elements of $\UU$ are called \defstyle{undefeatable ontic rules}.}

\end{itemize}

\end{definition}

We use $\epsilon$ to indicate the special context $(\emptyset, \emptyset, \emptyset)$.

\begin{definition}[Hierarchy of ontic rules in contexts]
\label{def:Hierarchy of ontic rules in contexts}

Let $\CC = (z, \DDD, \succ, \UU)$ be a context for a model $\MM$. Define $\HI{\CC}$, \defstyle{the hierarchy of ontic rules in $\CC$}, as the sequence $(\DDD_0, \dots, \DDD_n)$, which is constructed in the following way:
\begin{itemize}
\item Let $\DDD_0 = \{\DD \in \DDD \mid \text{$\DD$ is a maximal element of $\DDD$}\}$;
\item If $\DDD_0 \cup \dots \cup \DDD_k \neq \DDD$, let $\DDD_{k+1} = \{\DD \in \DDD \mid \text{$\DD$ is a maximal element of $\DDD - (\DDD_0 \cup \dots \cup \DDD_k)$}\}$, or else stop.
\end{itemize}

\end{definition}

Similar ways of defining hierarchies can also be found in the literature on social choice theory, such as \cite{jiang_hierarchical_2018}.

Here are some observations about $\HI{\CC} = (\DDD_0, \dots, \DDD_n)$. First, if $\DDD_0 = \emptyset$, then $n = 0$. Second, $\DDD_0, \dots, \DDD_n$ are pairwise disjoint, and their union is $\DDD$.

\begin{example}[Hierarchy of ontic rules in contexts]
\label{example:Hierarchy of ontic rules in contexts}
~
\begin{itemize}
\item $\HI{\epsilon} = \emptyset$. Note here, $\emptyset$ is not the empty sequence but the sequence with the empty set as its only element.
\item Let $\CC = (z, \DDD, \succ, \UU)$ be a context for a model $\MM$, where $\DDD = \{\DD_1, \DD_2, \DD_3\}$, $\DD_1 \succ \DD_2$, and $\DD_1 \succ \DD_3$. Then, $\HI{\CC} = (\{\DD_1\},\{\DD_2,\DD_3\})$.
\end{itemize}

\end{example}

\begin{definition}[Accepted and expected timelines by contexts]
\label{def:Accepted timelines and expected timelines by contexts}

Let $\MM = (W,r,<,V)$ be a model, $\CC = (z, \DDD, \succ, \UU)$ be a context for $\MM$, $\HI{\CC} = (\DDD_0, \dots, \DDD_n)$ be the hierarchy of ontic rules in $\CC$, and $w$ be a moment with respect to $z$.

Define the set $\seta{\CC}$ of \defstyle{accepted timelines by $\CC$} as $\EN{\UU}$. Note specially, $\EN{\emptyset} = \mathtt{TL} (\MM, z)$.

Define the set $\sete{\CC}$ of \defstyle{expected timelines by $\CC$} as follows:
\begin{itemize}
\item Suppose $\EN{\DDD_0} = \emptyset$. Then, $\sete{\CC} := \emptyset$.
\item Suppose $\EN{\DDD_0} \neq \emptyset$. Then, $\sete{\CC} := \EN{\DDD_0} \cap \dots \cap \EN{\DDD_k}$, where $(\DDD_0, \dots, \DDD_k)$ is the longest initial segment of $(\DDD_0, \dots, \DDD_n)$ such that $\EN{\DDD_0} \cap \dots \cap \EN{\DDD_k} \neq \emptyset$.
\end{itemize}

Define the set $\mseta{\CC}{w}$ of \defstyle{accepted timelines by $\CC$ passing through $w$} as $\{\pi' \in \seta{\CC} \mid w \in \pi\}$.

Define the set $\msete{\CC}{w}$ of \Fdefs{expected timelines by $\CC$ passing through $w$} as $\{\pi' \in \sete{\CC} \mid w \in \pi\}$. 

\end{definition}

Accepted timelines by $\CC$ are those meeting all undefeatable ontic rules. Intuitively, expected timelines by $\CC$ are determined as follows. From the top level of the hierarchy, consider as many levels as possible until the ontic rules in the considered levels do not collectively \emph{enable} any timelines. Expected timelines are those that meet all the ontic rules at the considered levels.

It is easy to see $\sete{\CC} \subseteq \seta{\CC}$ and $\msete{\CC}{w} \subseteq \mseta{\CC}{w}$. Although it is possible that $\seta{\CC} = \emptyset$, this kind of context will not be considered in the semantics.

\begin{example}[Expected timelines by contexts]
\label{example:Accepted timelines and expected timelines by contexts}

Let $\CC = (z, \DDD, \succ, \UU)$ be a context for a model $\MM$.
\begin{itemize}
\item Assume $\HI{\CC} = \emptyset$. Then, $\sete{\CC} = \mathtt{TL} (\MM, z)$.

\item Assume $\HI{\CC} = (\{\emptyset\},\{\{\pi_1,\pi_2\}\})$. Then, $\sete{\CC} = \EN{\{\emptyset\}} = \emptyset$.
\item Assume $\HI{\CC} = (\{\{\pi_1,\pi_2,\pi_3\},\{\pi_2,\pi_3,\pi_4\}\}, \{\{\pi_1,\pi_4\}\})$.
Then, $\sete{\CC} = \bigcap \{\{\pi_1,\pi_2,\pi_3\},$ $\{\pi_2,\pi_3,\pi_4\}\} = \{\pi_2, \pi_3\}$.
\end{itemize}

\end{example}

\begin{definition}[Contextualized models and contextualized pointed models] \label{def:Contextualized pointed models}

For every model $\MM$, context $\CC$, timeline $\pi$ in $\seta{\CC}$, and natural number $i$, $(\MM, \CC)$ is called a \defstyle{contextualized model} and $(\MM, \CC, \pi, i)$ is called a \defstyle{contextualized pointed model}.

\end{definition}

\noindent The reason for requiring $\pi$ to be in $\seta{\CC}$ is that the timelines outside $\seta{\CC}$ are unacceptable for the agent, and there is no point for her to consider them.
Note that for any contextualized pointed model $(\MM, \CC, \pi, i)$, neither $\seta{\CC}$ nor $\mseta{\CC}{\pi[i]}$ is empty.

\begin{example}[Contextualized pointed models]
\label{example:Contextualized pointed models}

Let $(\MM, \CC, \pi_3, 1)$ be a contextualized pointed model, where $\MM$ is indicated by figure \ref{fig:Another contextualized pointed model}, and $\CC = (w^0_1, \DDD, \succ, \UU)$ is a context such that $\DDD = \{\DD_1, \DD_2, \DD_3\}$, $\DD_1 \succ \DD_2 \succ \DD_3$, and $\UU = \{\DD_1\}$, where $\DD_1 = \{\pi_1,\pi_2,\pi_3\}$, $\DD_2 = \{\pi_1,\pi_2\}$, and $\DD_3 = \{\pi_4\}$. It can be verified that $\HI{\CC} = (\{\DD_1\},\{\DD_2\},\{\DD_3\})$, $\seta{\CC} = \{\pi_1,\pi_2,\pi_3\}$ and $\sete{\CC} = \{\pi_1,\pi_2\}$, which means that $\pi_4$ is unaccepted, and $\pi_3$ and $\pi_4$ are unexpected.

\begin{figure}[htb]

\begin{center}

\begin{tikzpicture}[node distance=25mm,scale=0.8,every node/.style={transform shape}]
\tikzstyle{every state}=[draw=black,text=black,minimum size=11mm]

\node[state] (s-1-1) [] {$w^1_1$};
\node[state] (s-1-2) [right of=s-1-1] {$w^1_2$};
\node[state] (s-1-3) [right of=s-1-2] {$w^1_3$};
\node[state] (s-1-4) [right of=s-1-3] {$w^1_4$};
\node[state] (s-0-1) [below=25mm, right=7mm] at (s-1-2) {$w^0_1$};

\node[state] (s-0-0) [below of=s-0-1] {$w^0_0$};

\node [above=6mm] at (s-1-1) {$p,q$};
\node [above=6mm] at (s-1-2) {$p,q$};
\node [above=6mm] at (s-1-3) {$p$};
\node [above=6mm] at (s-1-4) {$r$};

\path
(s-0-1) edge [->,left] node {} (s-1-1)
(s-0-1) edge [->,left] node {} (s-1-2)
(s-0-1) edge [->,left,color=red] node {} (s-1-3)
(s-0-1) edge [->,left] node {} (s-1-4)
(s-0-0) edge [->,left] node {} (s-0-1);

\node [above=15mm] at (s-1-1) {$\pi_1 \atop \vdots$};
\node [above=15mm] at (s-1-2) {$\pi_2 \atop \vdots$};
\node [above=15mm] at (s-1-3) {$\pi_3 \atop \red{\vdots}$};
\node [above=15mm] at (s-1-4) {$\pi_4 \atop \vdots$};

\end{tikzpicture}

\end{center}

\caption{A model for Example \ref{example:Contextualized pointed models}}

\label{fig:Another contextualized pointed model}

\end{figure}

\end{example}

\subsection{Semantics}

\begin{definition}[Semantics for $\Phi_\SWHTN$]
\label{def:Semantics for SWON}

$\MM, \CC, \pi, i \Vdash \phi$, formulas $\phi$ in $\Phi_\SWHTN$ being true at contextualized pointed models $(\MM, \CC, \pi, i)$, is defined as follows:

\begin{tabular}{lll}
$\MM, \CC, \pi, i \not \Vdash \bot$ & & \\
$\MM, \CC, \pi, i \Vdash p$ & $\Leftrightarrow$ & \parbox[t]{22em}{$\pi[i] \in V(p)$} \\
$\MM, \CC, \pi, i \Vdash \neg \phi$ & $\Leftrightarrow$ & \parbox[t]{22em}{$\MM, \CC, \pi, i \not \Vdash \phi$} \\
$\MM, \CC, \pi, i \Vdash \phi \land \psi$ & $\Leftrightarrow$ & \parbox[t]{22em}{$\MM, \CC, \pi, i \Vdash \phi$ and $\MM, \CC, \pi, i \Vdash \psi$} \\
$\MM, \CC, \pi, i \Vdash \XXX \phi$ & $\Leftrightarrow$ & \parbox[t]{22em}{$\MM, \CC, \pi, i+1 \Vdash \phi$} \\
$\MM, \CC, \pi, i \Vdash \YYY \phi$ & $\Leftrightarrow$ & \parbox[t]{22em}{if $i > 0$, then $\MM, \CC, \pi, i-1 \Vdash \phi$} \\
$\MM, \CC, \pi, i \Vdash \SHN \phi$ & $\Leftrightarrow$ & \parbox[t]{22em}{$\MM, \CC, \pi', i \Vdash \phi$ for every $\pi' \in \mseta{\CC}{\pi[i]}$} \\
$\MM, \CC, \pi, i \Vdash \WHN \phi$ & $\Leftrightarrow$ & \parbox[t]{22em}{$\MM, \CC, \pi', i \Vdash \phi$ for every $\pi' \in \msete{\CC}{\pi[i]}$} \\
$\MM, \CC, \pi, i \Vdash \STN \phi$ & $\Leftrightarrow$ & \parbox[t]{22em}{$\MM, \CC, \pi', i \Vdash \phi$ for every $\pi' \in \seta{\CC}$} \\
$\MM, \CC, \pi, i \Vdash \WTN \phi$ & $\Leftrightarrow$ & \parbox[t]{22em}{$\MM, \CC, \pi', i \Vdash \phi$ for every $\pi' \in \sete{\CC}$} \\
\end{tabular}

\end{definition}

It can be verified that:

\medskip

\begin{tabular}{lll}
$\MM, \CC, \pi, i \Vdash \Froot$ & $\Leftrightarrow$ & \parbox[t]{22em}{$i = 0$} \\
$\MM, \CC, \pi, i \Vdash \YYY' \phi$ & $\Leftrightarrow$ & \parbox[t]{22em}{$i > 0$ and $\MM, \CC, \pi, i-1 \Vdash \phi$} \\
$\MM, \CC, \pi, i \Vdash \SHP \phi$ & $\Leftrightarrow$ & \parbox[t]{22em}{$\MM, \CC, \pi', i \Vdash \phi$ for some $\pi' \in \mseta{\CC}{\pi[i]}$} \\
$\MM, \CC, \pi, i \Vdash \WHP \phi$ & $\Leftrightarrow$ & \parbox[t]{22em}{$\MM, \CC, \pi', i \Vdash \phi$ for some $\pi' \in \msete{\CC}{\pi[i]}$} \\
$\MM, \CC, \pi, i \Vdash \STP \phi$ & $\Leftrightarrow$ & \parbox[t]{22em}{$\MM, \CC, \pi', i \Vdash \phi$ for some $\pi' \in \seta{\CC}$} \\
$\MM, \CC, \pi, i \Vdash \WTP \phi$ & $\Leftrightarrow$ & \parbox[t]{22em}{$\MM, \CC, \pi', i \Vdash \phi$ for some $\pi' \in \sete{\CC}$} \\
\end{tabular}

\medskip

Note $\Froot$ indicates the starting point of the context but not the starting point of the model.

\medskip

We say that a formula $\phi$ is \defstyle{valid} ($\models \phi$) if $\MM,\CC,\pi,i \Vdash \phi$ for all contextualized pointed models $(\MM,\CC,\pi,i)$; $\phi$ is \defstyle{satisfiable} if $\MM,\CC,\pi,i \Vdash \phi$ for some contextualized pointed model $(\MM,\CC,\pi,i)$.

\subsection{Analysis of some examples}

In this subsection, we analyze some examples of strong/weak historical/temporal necessities using our formalism.

\begin{example}
\label{example:tiger}

The people of a tribe captured some animals and put them in isolated cages in a locked room. They would deal with these animals in the following way: every evening, two animals are randomly released by a special device; the next morning, they remove the dead and return the living to their cages.

\begin{exe}

\ex\label{exe:The tiger must be alive tomorrow morning}

Yesterday morning, they captured three animals: a tiger, a lion, and a dog.
Yesterday evening, the tiger and the lion were freed.
This morning, they found that the tiger had killed the lion.
The chief says:

\vspace{2pt}

\exs{The tiger \textbf{must} be alive tomorrow morning.}

\vspace{2pt}

\emph{``Must'' in this sentence expresses \emph{strong historical necessity}. The chief accepts two ontic rules: \emph{tigers kill dogs}, and \emph{lions kill dogs}. He treats them as undefeatable.
According to the two rules, considering all possible timelines passing through the present world, it is unacceptable that the tiger will be killed. Thus, he can utter the sentence.
}

\vspace{2pt}

\ex\label{exe:The lion should be alive now}

Yesterday morning, they captured three animals: a lion, a leopard, and a goat.
Yesterday evening, the lion and the leopard were freed.
This morning, they found that the leopard had killed the lion.
The chief says:

\vspace{2pt}

\exs{The lion \textbf{should} be alive now.}

\vspace{2pt}

\emph{``Should'' in this sentence expresses \emph{weak temporal necessity}. How is this sentence true? The chief accepts three ontic rules: \emph{lions kill leopards}, \emph{lions kill goats}, and \emph{leopards kill goats}. He does not treat the first as undefeatable, but treats the second and third as undefeatable.
According to these ontic rules, considering all possible timelines, it is unexpected that the lion is killed. Thus, he can utter the sentence.
}

\end{exe}

\end{example}

\begin{example}
\label{example:??}

We show how \ref{exe:The tiger must be alive tomorrow morning} is analyzed in the formalism.

We use $f_x$ to indicate \emph{$x$ is free} and use $a_x$ to indicate \emph{$x$ is alive}, where $x$ can be $t$ (the tiger), $l$ (the lion) or $d$ (the dog).

The figure \ref{fig:A model tiger} denotes a model $\MM$. Let $\CC = (w^0_1, \DDD, \succ, \UU)$ be a context, where:
\begin{itemize}

\item $\DDD$ has two ontic rules:

$\DD_1 = \{\pi_1, \pi_3,\pi_4,\pi_5,\pi_6,\pi_7,\pi_9,\pi_{10}\}$ and $\DD_2 = \{\pi_1,\pi_2,\pi_3,\pi_5,\pi_6,\pi_9,\pi_{10},\pi_{11}\}$;

\item $\succ = \emptyset$;

\item $\UU = \{\DDD_1, \DDD_2\}$.

\end{itemize}

The ontic rule $\DD_1$ says that \emph{if the tiger and the dog are free, the tiger is alive}.
The ontic rule $\DD_2$ says that \emph{if the lion and the dog are free, the lion is alive}.

The following can be verified:

$\HI{\CC} = (\{\DD_1, \DD_2\})$; $\seta{\CC} = \{\pi_1, \pi_3,\pi_5,\pi_6,\pi_7,\pi_9,\pi_{10}\}$; $\mseta{\CC}{\pi_1[1]} = \{\pi_1\}$.

The formula $\SHN \XXX a_t$ means \emph{the tiger must be alive right tomorrow}. It can be verified that $\SHN \XXX a_t$ is true at $(\MM, \CC, \pi_1, 1)$.

\begin{figure}[htb]

\begin{center}

\begin{tikzpicture}[node distance=33mm,scale=0.75,every node/.style={transform shape}]
\tikzstyle{every state}=[draw=black,text=black,minimum size=11mm]

\node[state,fill=blue!10] (s-1-1) [] {$w^1_1$};
\node[] (l-1-1) [above=5mm] at (s-1-1) {$f_t,f_l \atop {a_t,a_d}$};

\node[state] (s-1-1-1) [above=35mm,left=3mm] at (s-1-1) {$w^2_1$};
\node[] (l-1-1-1) [above=5mm] at (s-1-1-1) {$f_t,f_d \atop {a_t}$};

\node[state] (s-1-1-2) [above=35mm,right=3mm] at (s-1-1) {$w^2_2$};
\node[] (l-1-1-2) [above=5mm] at (s-1-1-2) {$f_t,f_d \atop {a_d}$};


\node[state] (s-1-2) [right of=s-1-1] {$w^1_2$};
\node[] (l-1-2) [above=5mm] at (s-1-2) {$f_t,f_l \atop {a_l,a_d}$};

\node[state] (s-1-2-1) [above=35mm,left=3mm] at (s-1-2) {$w^2_3$};
\node[] (l-1-2-1) [above=5mm] at (s-1-2-1) {$f_l,f_d \atop {a_l}$};

\node[state] (s-1-2-2) [above=35mm,right=3mm] at (s-1-2) {$w^2_4$};
\node[] (l-1-2-2) [above=5mm] at (s-1-2-2) {$f_l,f_d \atop {a_d}$};


\node[state] (s-1-3) [right of=s-1-2] {$w^1_3$};
\node[] (l-1-3) [above=5mm] at (s-1-3) {$f_l,f_d \atop {a_t,a_l}$};

\node[state] (s-1-3-1) [above=35mm,left=3mm] at (s-1-3) {$w^2_5$};
\node[] (l-1-3-1) [above=5mm] at (s-1-3-1) {$f_t,f_l \atop {a_t}$};

\node[state] (s-1-3-2) [above=35mm,right=3mm] at (s-1-3) {$w^2_6$};
\node[] (l-1-3-2) [above=5mm] at (s-1-3-2) {$f_t,f_l \atop {a_l}$};


\node[state] (s-1-4) [right of=s-1-3] {$w^1_4$};
\node[] (l-1-4) [above=5mm] at (s-1-4) {$f_l,f_d \atop {a_t,a_d}$};

\node[state] (s-1-4-1) [above=35mm,left=3mm] at (s-1-4) {$w^2_7$};
\node[] (l-1-4-1) [above=5mm] at (s-1-4-1) {$f_t,f_d \atop {a_t}$};

\node[state] (s-1-4-2) [above=35mm,right=3mm] at (s-1-4) {$w^2_8$};
\node[] (l-1-4-2) [above=5mm] at (s-1-4-2) {$f_t,f_d \atop {a_d}$};


\node[state] (s-1-5) [right of=s-1-4] {$w^1_5$};
\node[] (l-1-5) [above=5mm] at (s-1-5) {$f_t,f_d \atop {a_t,a_l}$};

\node[state] (s-1-5-1) [above=35mm,left=3mm] at (s-1-5) {$w^2_9$};
\node[] (l-1-5-1) [above=5mm] at (s-1-5-1) {$f_t,f_l \atop {a_t}$};

\node[state] (s-1-5-2) [above=35mm,right=3mm] at (s-1-5) {$w^2_{10}$};
\node[] (l-1-5-2) [above=5mm] at (s-1-5-2) {$f_t,f_l \atop {a_l}$};


\node[state] (s-1-6) [right of=s-1-5] {$w^1_6$};
\node[] (l-1-6) [above=5mm] at (s-1-6) {$f_t,f_d \atop {a_l,a_d}$};

\node[state] (s-1-6-1) [above=35mm,left=3mm] at (s-1-6) {$w^2_{11}$};
\node[] (l-1-6-1) [above=5mm] at (s-1-6-1) {$f_l,f_d \atop {a_l}$};

\node[state] (s-1-6-2) [above=35mm,right=3mm] at (s-1-6) {$w^2_{12}$};
\node[] (l-1-6-2) [above=5mm] at (s-1-6-2) {$f_l,f_d \atop {a_d}$};

\node[state] (s-0-1) [below=35mm, right=12mm] at (s-1-3) {$w^0_1$};
\node[] (l-0-1) [above=5mm] at (s-0-1) {$a_t, a_l, a_d$};

\node [above=14mm] at (s-1-1-1) {$\pi_1 \atop \vdots$};
\node [above=14mm] at (s-1-1-2) {$\pi_2 \atop \vdots$};

\node [above=14mm] at (s-1-2-1) {$\pi_3 \atop \vdots$};
\node [above=14mm] at (s-1-2-2) {$\pi_4 \atop \vdots$};

\node [above=14mm] at (s-1-3-1) {$\pi_5 \atop \vdots$};
\node [above=14mm] at (s-1-3-2) {$\pi_6 \atop \vdots$};

\node [above=14mm] at (s-1-4-1) {$\pi_7 \atop \vdots$};
\node [above=14mm] at (s-1-4-2) {$\pi_8 \atop \vdots$};

\node [above=14mm] at (s-1-5-1) {$\pi_9 \atop \vdots$};
\node [above=14mm] at (s-1-5-2) {$\pi_{10} \atop \vdots$};

\node [above=14mm] at (s-1-6-1) {$\pi_{11} \atop \vdots$};
\node [above=14mm] at (s-1-6-2) {$\pi_{12} \atop \vdots$};

\path
(s-0-1) edge [->,left] node {} (s-1-1)
(s-0-1) edge [->,left] node {} (s-1-2)
(s-0-1) edge [->,left] node {} (s-1-3)
(s-0-1) edge [->,left] node {} (s-1-4)
(s-0-1) edge [->,left] node {} (s-1-5)
(s-0-1) edge [->,left] node {} (s-1-6)
(s-1-1) edge [->,left] node {} (s-1-1-1)
(s-1-1) edge [->,left] node {} (s-1-1-2)
(s-1-2) edge [->,left] node {} (s-1-2-1)
(s-1-2) edge [->,left] node {} (s-1-2-2)
(s-1-3) edge [->,left] node {} (s-1-3-1)
(s-1-3) edge [->,left] node {} (s-1-3-2)
(s-1-4) edge [->,left] node {} (s-1-4-1)
(s-1-4) edge [->,left] node {} (s-1-4-2)
(s-1-5) edge [->,left] node {} (s-1-5-1)
(s-1-5) edge [->,left] node {} (s-1-5-2)
(s-1-6) edge [->,left] node {} (s-1-6-1)
(s-1-6) edge [->,left] node {} (s-1-6-2)
;

\end{tikzpicture}

\end{center}

\caption{A model for \ref{exe:The tiger must be alive tomorrow morning}}
\label{fig:A model tiger}

\end{figure}

\end{example}

\begin{example}
\label{example:??}

We show how \ref{exe:The lion should be alive now} is analyzed in the formalism.

We use $f_x$ to indicate \emph{$x$ is free} and use $a_x$ to indicate \emph{$x$ is alive}, where $x$ can be $l$ (the lion), $p$ (the leopard) or $g$ (the goat).

The figure \ref{fig:A model lion} denotes a model $\MM$. Let $\CC = (w^0_1, \DDD, \succ, \UU)$ be a context, where:
\begin{itemize}
\item $\DDD$ has three ontic rules: 

$\DD_1 = \{\pi_1, \pi_2, \pi_5,\pi_7,\pi_8,\pi_9,\pi_{11},\pi_{12}\}$;
$\DD_2 = \{\pi_1,\pi_3,\pi_4,\pi_5,\pi_6,\pi_7,\pi_9,\pi_{10}\}$;

$\DD_3 = \{\pi_1, \pi_2, \pi_3,\pi_5,\pi_6,\pi_9,\pi_{10},\pi_{11}\}$.

\item $\succ = \emptyset$;
\item $\UU = \{\DDD_2, \DDD_3\}$.
\end{itemize}

The ontic rule $\DD_1$ says that \emph{if the lion and the leopard are free, the lion is alive}.
The ontic rule $\DD_2$ says that \emph{if the lion and the goat are free, the lion is alive}.
The ontic rule $\DD_3$ says that \emph{if the leopard and the goat are free, the leopard is alive}.

The following can be checked: $\HI{\CC} = (\{\DD_1, \DD_2, \DD_3\})$; $\seta{\CC} = \{\pi_1,\pi_3,\pi_5,\pi_6,\pi_9,\pi_{10}\}$; $\sete{\CC} = \{\pi_1,\pi_5,\pi_9\}$.

The formula $\WHN a_l$ means \emph{the lion should be alive now}. It can be verified that $\WHN a_l$ is true at $(\MM, \CC, \pi_3, 1)$.

\begin{figure}[htb]

\begin{center}

\begin{tikzpicture}[node distance=33mm,scale=0.75,every node/.style={transform shape}]
\tikzstyle{every state}=[draw=black,text=black,minimum size=11mm]

\node[state] (s-1-1) [] {$w^1_1$};
\node[] (l-1-1) [above=5mm] at (s-1-1) {$f_l,f_p \atop {a_l,a_g}$};

\node[state] (s-1-1-1) [above=35mm,left=3mm] at (s-1-1) {$w^2_1$};
\node[] (l-1-1-1) [above=5mm] at (s-1-1-1) {$f_l,f_g \atop {a_l}$};

\node[state] (s-1-1-2) [above=35mm,right=3mm] at (s-1-1) {$w^2_2$};
\node[] (l-1-1-2) [above=5mm] at (s-1-1-2) {$f_l,f_g \atop {a_g}$};


\node[state,fill=blue!10] (s-1-2) [right of=s-1-1] {$w^1_2$};
\node[] (l-1-2) [above=5mm] at (s-1-2) {$f_l,f_p \atop {a_p,a_g}$};

\node[state] (s-1-2-1) [above=35mm,left=3mm] at (s-1-2) {$w^2_3$};
\node[] (l-1-2-1) [above=5mm] at (s-1-2-1) {$f_p,f_g \atop {a_p}$};

\node[state] (s-1-2-2) [above=35mm,right=3mm] at (s-1-2) {$w^2_4$};
\node[] (l-1-2-2) [above=5mm] at (s-1-2-2) {$f_p,f_g \atop {a_g}$};


\node[state] (s-1-3) [right of=s-1-2] {$w^1_3$};
\node[] (l-1-3) [above=5mm] at (s-1-3) {$f_p,f_g \atop {a_l,a_p}$};

\node[state] (s-1-3-1) [above=35mm,left=3mm] at (s-1-3) {$w^2_5$};
\node[] (l-1-3-1) [above=5mm] at (s-1-3-1) {$f_l,f_p \atop {a_l}$};

\node[state] (s-1-3-2) [above=35mm,right=3mm] at (s-1-3) {$w^2_6$};
\node[] (l-1-3-2) [above=5mm] at (s-1-3-2) {$f_l,f_p \atop {a_p}$};


\node[state] (s-1-4) [right of=s-1-3] {$w^1_4$};
\node[] (l-1-4) [above=5mm] at (s-1-4) {$f_p,f_g \atop {a_l,a_g}$};

\node[state] (s-1-4-1) [above=35mm,left=3mm] at (s-1-4) {$w^2_7$};
\node[] (l-1-4-1) [above=5mm] at (s-1-4-1) {$f_l,f_g \atop {a_l}$};

\node[state] (s-1-4-2) [above=35mm,right=3mm] at (s-1-4) {$w^2_8$};
\node[] (l-1-4-2) [above=5mm] at (s-1-4-2) {$f_l,f_g \atop {a_g}$};


\node[state] (s-1-5) [right of=s-1-4] {$w^1_5$};
\node[] (l-1-5) [above=5mm] at (s-1-5) {$f_l,f_g \atop {a_l,a_p}$};

\node[state] (s-1-5-1) [above=35mm,left=3mm] at (s-1-5) {$w^2_9$};
\node[] (l-1-5-1) [above=5mm] at (s-1-5-1) {$f_l,f_p \atop {a_l}$};

\node[state] (s-1-5-2) [above=35mm,right=3mm] at (s-1-5) {$w^2_{10}$};
\node[] (l-1-5-2) [above=5mm] at (s-1-5-2) {$f_l,f_p \atop {a_p}$};


\node[state] (s-1-6) [right of=s-1-5] {$w^1_6$};
\node[] (l-1-6) [above=5mm] at (s-1-6) {$f_l,f_g \atop {a_p,a_g}$};

\node[state] (s-1-6-1) [above=35mm,left=3mm] at (s-1-6) {$w^2_{11}$};
\node[] (l-1-6-1) [above=5mm] at (s-1-6-1) {$f_p,f_g \atop {a_p}$};

\node[state] (s-1-6-2) [above=35mm,right=3mm] at (s-1-6) {$w^2_{12}$};
\node[] (l-1-6-2) [above=5mm] at (s-1-6-2) {$f_p,f_g \atop {a_g}$};

\node[state] (s-0-1) [below=35mm, right=12mm] at (s-1-3) {$w^0_1$};
\node[] (l-0-1) [above=5mm] at (s-0-1) {$a_l, a_p, a_g$};

\node [above=14mm] at (s-1-1-1) {$\pi_1 \atop \vdots$};
\node [above=14mm] at (s-1-1-2) {$\pi_2 \atop \vdots$};

\node [above=14mm] at (s-1-2-1) {$\pi_3 \atop \vdots$};
\node [above=14mm] at (s-1-2-2) {$\pi_4 \atop \vdots$};

\node [above=14mm] at (s-1-3-1) {$\pi_5 \atop \vdots$};
\node [above=14mm] at (s-1-3-2) {$\pi_6 \atop \vdots$};

\node [above=14mm] at (s-1-4-1) {$\pi_7 \atop \vdots$};
\node [above=14mm] at (s-1-4-2) {$\pi_8 \atop \vdots$};

\node [above=14mm] at (s-1-5-1) {$\pi_9 \atop \vdots$};
\node [above=14mm] at (s-1-5-2) {$\pi_{10} \atop \vdots$};

\node [above=14mm] at (s-1-6-1) {$\pi_{11} \atop \vdots$};
\node [above=14mm] at (s-1-6-2) {$\pi_{12} \atop \vdots$};

\path
(s-0-1) edge [->,left] node {} (s-1-1)
(s-0-1) edge [->,left] node {} (s-1-2)
(s-0-1) edge [->,left] node {} (s-1-3)
(s-0-1) edge [->,left] node {} (s-1-4)
(s-0-1) edge [->,left] node {} (s-1-5)
(s-0-1) edge [->,left] node {} (s-1-6)
(s-1-1) edge [->,left] node {} (s-1-1-1)
(s-1-1) edge [->,left] node {} (s-1-1-2)
(s-1-2) edge [->,left] node {} (s-1-2-1)
(s-1-2) edge [->,left] node {} (s-1-2-2)
(s-1-3) edge [->,left] node {} (s-1-3-1)
(s-1-3) edge [->,left] node {} (s-1-3-2)
(s-1-4) edge [->,left] node {} (s-1-4-1)
(s-1-4) edge [->,left] node {} (s-1-4-2)
(s-1-5) edge [->,left] node {} (s-1-5-1)
(s-1-5) edge [->,left] node {} (s-1-5-2)
(s-1-6) edge [->,left] node {} (s-1-6-1)
(s-1-6) edge [->,left] node {} (s-1-6-2)
;

\end{tikzpicture}

\end{center}

\caption{A model for \ref{exe:The lion should be alive now}}

\label{fig:A model lion}

\end{figure}

\end{example}

\subsection{Discussed features of historical/temporal necessities/possibilities are reflected in the formalism}

\paragraph{Duals}

\emph{Strong historical necessity (but not weak historical necessity) and historical possibility are duals of each other, and strong temporal necessity (but not weak temporal necessity) and temporal possibility are duals of each other.}

It is easy to check that all these points are reflected in the formalism.

\paragraph{Monotonicity}

\emph{Strong historical necessity and strong temporal necessity are monotonic, but weak historical necessity and weak temporal necessity are nonmonotonic.}

We do not deal with conditional strong/weak historical/temporal necessities in this paper. To address this, we need to define hypothesis sets and how these, in conjunction with ontic systems, determine the set of accepted timelines and the set of expected timelines. 
We claim that with this done, the following can be achived: (1) when an ontic system and a hypothesis set determine the set of accepted timelines, the ``bigger'' the hypothesis set is, the smaller the set of accepted timelines is; (2) when an ontic system and a hypothesis set determine the set of expected timelines, it may happen that the hypothesis set gets ``bigger'', but the set of accepted timelines gets smaller. Consequently, strong historical necessity and strong temporal necessity would be monotonic, but weak historical necessity and weak temporal necessity would not be.

\paragraph{Implications}

\begin{enumerate}[label=(\arabic*)]

\item

\emph{Strong historical necessity is stronger than weak historical necessity, and strong temporal necessity is stronger than weak temporal necessity.}

\item 

\emph{The principle of necessity of the past holds for strong/weak historical necessities but does not hold for strong/weak temporal necessities.}

\item 

\emph{The principle of possibility of the past holds for historical possibility but does not hold for temporal possibility.}

\item 

\emph{Strong and weak temporal necessities are preserved under the temporal shift to the past, but neither strong nor weak historical necessity is.}

\item 

\emph{Temporal possibility is preserved under the temporal shift to the future, but historical possibility is not.}

\end{enumerate}

These points are reflected in the following fact:

\begin{fact}
~

\begin{enumerate}[label=(\arabic*)]

\item

For all $\phi$, $\models \SHN \phi \rightarrow \WHN \phi$ and $\models \STN \phi \rightarrow \WTN \phi$, but it is not that for all $\phi$, $\models \WHN \phi \rightarrow \SHN \phi$, and it is not that for all $\phi$, $\models \WTN \phi \rightarrow \STN \phi$.

\item 

For all $\chi$ containing no $\XXX$, $\models \chi \rightarrow \SHN \chi$ and $\models \chi \rightarrow \WHN \chi$, but it is not that for all $\chi$ containing no $\XXX$, $\models \chi \rightarrow \STN \chi$, and it is not that for all $\chi$ containing no $\XXX$, $\models \chi \rightarrow \WTN \chi$.

\item 

For all $\chi$ containing no $\XXX$, $\models \SHP \chi \rightarrow \chi$, but it is not that for all $\chi$ containing no $\XXX$, $\models \STP \chi \rightarrow \chi$.

\item

\begin{enumerate}

\item

for all $\phi$, $\models \Box \phi \rightarrow \YYY \Box \XXX \phi$, where $\Box \in \{\STN, \WTN\}$.

\item

it is not that for all $\phi$, $\models \Box \phi \rightarrow \YYY \Box \XXX \phi$, where $\Box \in \{\SHN, \WHN\}$.

\end{enumerate}

\item

\begin{enumerate}

\item

for all $\phi$, $\models \Diamond \phi \rightarrow \XXX \Diamond \YYY \phi$, where $\Diamond \in \{\STP, \WTP\}$.

\item

it is not that for all $\phi$, $\models \Diamond \phi \rightarrow \XXX \Diamond \YYY \phi$, where $\Diamond \in \{\SHP, \WHP\}$.

\end{enumerate}

\end{enumerate}

\end{fact}

\begin{proof}~

Items (1), (2), (3), (4b), and (5b) are easy to check, and we skip their proofs.
\begin{enumerate}

\item[(4a)]

We consider only the case $\Box = \STN$; the case $\Box = \WTN$ is similar.

Fix a formula $\phi$ and a contextualized pointed model $(\MM, \CC, \pi, i)$.
Assume $\MM, \CC, \pi, i \not \Vdash \YYY \STN \XXX \phi$. Then $i > 0$ and $\MM, \CC, \pi, i-1 \not \Vdash \STN \XXX \phi$.
Then there is $\pi' \in \seta{\CC}$ such that $\MM, \CC, \pi', i-1 \not \Vdash \XXX \phi$. Then $\MM, \CC, \pi', i \not \Vdash \phi$. Then $\MM, \CC, \pi, i \not \Vdash \STN \phi$. Then $\MM, \CC, \pi, i \Vdash \STN \phi \rightarrow \YYY \STN \XXX \phi$.

\item[(5b)]

We consider only the case $\Diamond = \STP$; the case $\Diamond = \WTP$ is similar.

Fix a formula $\phi$ and a contextualized pointed model $(\MM, \CC, \pi, i)$.
Assume $\MM, \CC, \pi, i \not \Vdash \XXX \STP \YYY \phi$. Then $\MM, \CC, \pi, i+1 \not \Vdash \STP \YYY \phi$. Then there is $\pi' \in \seta{\CC}$ such that $\MM, \CC, \pi', i+1 \not \Vdash \YYY \phi$. Then $\MM, \CC, \pi', i \not \Vdash \phi$. Then $\MM, \CC, \pi, i \not \Vdash \STP \phi$. Then $\MM, \CC, \pi, i \Vdash \STP \phi \rightarrow \XXX \STP \YYY \phi$.

\end{enumerate}

\end{proof}

\paragraph{Coexistence}
~

\begin{table}[H]

\centering

\begin{tabular}{|c|c|c|}
\hline
& $\HP \will p$ & $\TP \will p$ \\
\hline
$\dec \will \neg p$ & not co-existent $(1)$ & co-existent $(2)$ \\
\hline
\end{tabular}

\caption{Some features of historical/temporal possibilities}
\label{tab:placeholder0pp}

\end{table}

Note that we treat $\dec \will \neg p$ as equivalent to $\SHN \will \neg p$. The features of historical/temporal possibilities given in table \ref{tab:placeholder0pp} are reflected in the following fact, which is easy to check:

\begin{fact}~

\begin{enumerate}[label=(\arabic*)]

\item

$\SHN \XXX \neg p \land \SHP \XXX p$ is not satisfiable.

\item 

$\SHN \XXX \neg p \land \STP \XXX p$ is satisfiable.

\end{enumerate}

\end{fact}

\begin{table}[H]

\centering

\begin{tabular}{|c|c|c|}
\hline
 & $\XSHN \will p$ & $\XWHN \will p$ \\
\hline
$\HP \will \neg p$ & not co-existent $(1)$ & co-existent $(2)$ \\
\hline
$\dec \will \neg p$ & not co-existent (3) & trivially co-existent (4) \\
\hline
\end{tabular}

\caption{Some features of strong/weak historical necessities}

\label{tab:placeholder1pp}

\end{table}

The features of historical/temporal necessities given in table \ref{tab:placeholder1pp} are reflected in the following fact, which is easy to check:

\begin{fact}~

\begin{enumerate}[label=(\arabic*)]

\item

$\SHP \XXX \neg p \land \SHN \XXX p$ is not satisfiable.

\item 

$\SHP \XXX \neg p \land \WHN \XXX p$ is satisfiable.

\item 

$\SHN \XXX \neg p \land \SHN \XXX p$ is not satisfiable.

\item 

$\SHN \XXX \neg p \land \WHN \XXX p$ is satisfiable, but $(\SHN \XXX \neg p \land \WHN \XXX p) \rightarrow \WHN \bot$ is valid.

\end{enumerate}

\end{fact}

\begin{table}[H]

\centering

\begin{tabular}{|c|c|c|}
\hline
& $\XSTN p$ & $\XWTN p$ \\
\hline
$\TP \neg p$ & not co-existent $(1)$ & co-existent $(2)$ \\
\hline
$\neg p$ & not co-existent $(3)$ & co-existent $(4)$ \\
\hline
\end{tabular}

\medskip

\begin{tabular}{|c|c|c|}
\hline
 & $\XSTN \will p$ & $\XWTN \will p$ \\
\hline
$\TP \will \neg p$ & not co-existent $(5)$ & co-existent $(6)$ \\
\hline
$\dec \will \neg p$ & not co-existent $(7)$ & co-existent $(8)$ \\
\hline
\end{tabular}

\caption{Some features of strong/weak temporal necessities}
\label{tab:placeholder2pp}

\end{table}

The features of strong/weak temporal necessities given in table \ref{tab:placeholder2pp} are reflected in the following fact, which is easy to check:

\begin{fact}~

\begin{enumerate}[label=(\arabic*)]

\item

$\STP \neg p \land \STN p$ is not satisfiable.

\item

$\STP \neg p \land \WTN p$ is satisfiable.

\item

$\neg p \land \STN p$ is not satisfiable.

\item

$\neg p \land \WTN p$ is satisfiable.

\item

$\STP \XXX \neg p \land \STN \XXX p$ is not satisfiable.

\item

$\STP \XXX \neg p \land \WTN \XXX p$ is satisfiable.

\item

$\SHN \XXX \neg p \land \STN \XXX p$ is not satisfiable.

\item

$\SHN \XXX \neg p \land \WTN \XXX p$ is satisfiable.

\end{enumerate}

\end{fact}

\newcommand{\AT}{\mathtt{AT}}
\newcommand{\ET}{\mathtt{ET}}

\subsection{Technically, the semantics can be simplified without changing the set of valid formulas}

In the semantics given above, formulas are evaluated at contextualized pointed models $(\MM,\CC,\pi,i)$, where $\MM = (W,r,<,V)$ and $\CC = (z, \DDD, \succ, \UU)$.
The function of $\CC$ is to determine a set $\seta{\CC}$ of accepted timelines and a set $\sete{\CC}$ of expected timelines from $z$. In this subsection, we show that the semantics can be simplified as follows: (1) we can assume $r = z$; (2) we can directly give the two sets $\seta{\CC}$ and $\sete{\CC}$.
The reason this can be done is as follows. First, when evaluating a formula at $(\MM,\CC,\pi,i)$, only those states \emph{reachable} from $z$ are used. Second, the operator $\STN/\WTN$ is a universal quantifier over the set $\seta{\CC}/\sete{\CC}$ of accepted/expected timelines; the operator $\SHN/\WHN$ is a universal quantifier over the set $\mseta{\CC}{\pi[i]}/\msete{\CC}{\pi[i]}$ of accepted/expected timelines; however, how the two sets are determined is not reflected in the truth conditions of $\STN \phi$ and $\WTN \phi$.

\begin{definition}[Simplified contexts]
\label{def:Simplified contexts}

Let $\MM = (W,r,<,V)$ be a model. A tuple $\CC = (\AT,\ET)$ is a \Fdefs{simplified context} for $\MM$ if $\AT$ is a (possibly empty) set of timelines of $\MM$ from $r$ and $\ET$ is a (possibly empty) subset of $\AT$.

\end{definition}

For every model $\MM$, simplified context $\CC = (\AT,\ET)$, timeline $\pi$ in $\AT$, and natural number $i$, we call $(\MM, \CC)$ a \defstyle{simplified contextualized model} and $(\MM, \CC, \pi, i)$ a \defstyle{simplified contextualized pointed model}.

Truth conditions of formulas in $\Phi_\SWHTN$ at simplified contextualized pointed models are defined in the \emph{imagined} way.

Define \Fdefs{generated submodels} as usual.
The following fact indicates how the semantics can be simplified.

\begin{fact}
\label{fact:the semantics can be simplified}
~

\begin{enumerate}[label=(\arabic*),leftmargin=3.33em]

\item 

Let $(\MM, \CC, \pi, i)$, where $\MM = (W,r,<,V)$ and $\CC = (z, \DDD, \succ, \UU)$, be a contextualized pointed model. Let $(\MM',\CC',\pi,i)$, where $\MM' = (W',z,<',V')$ and $\CC' = (\AT, \ET)$, be a simplified contextualized pointed model meeting the following conditions:

\begin{enumerate}

\item 

$\MM'$ is the generated submodel of $\MM$ from $z$;

\item 

$\AT = \seta{\CC}$ and $\ET = \sete{\CC}$.

\end{enumerate}

Then, for every $\phi \in \Phi_\SWHTN$, $\MM, \CC, \pi, i \Vdash \phi$ if and only if $\MM', \CC', \pi, i \Vdash \phi$.

\item 

Let $(\MM,\CC',\pi,i)$, where $\MM= (W,r,<,V)$ and $\CC' = (\AT, \ET)$, be a simplified contextualized pointed model. Let $(\MM, \CC, \pi, i)$, where $\CC = (r, \DDD, \succ, \UU)$, be a contextualized pointed model such that $\DDD = \{\AT,\ET\}$, $\succ = \emptyset$, and $\UU = \{\AT\}$.

Then, for every $\phi \in \Phi_\SWHTN$, $\MM, \CC, \pi, i \Vdash \phi$ if and only if $\MM, \CC', \pi, i \Vdash \phi$.

\end{enumerate}

\end{fact}

\noindent This result can be easily shown, and we skip its proof.

Technically, the semantics can be simplified. However, this does not mean that the original semantics is not meaningful. As discussed before, weak historical/temporal necessities are nonmonotonic in nature, which is reflected in the original semantics. In the simplified semantics, how expected timelines are determined is implicit. Consequently, weak historical/temporal necessities are monotonic in it.
Based on the original semantics, we plan to address conditional weak historical/temporal necessities, where the antecedent changes the expected timelines.

In the sequel, when showing technical results, we will use the simplified semantics.

\paragraph{Remarks}

From Fact \ref{fact:the semantics can be simplified}, we can see that if we do not assume a starting point for models, the set of valid formulas will still be the same.

\section{Comparison to some related work from technical perspectives}
\label{section: Comparison to some related work}

In the literature, there have been many formal theories of historical necessity in the sense of Thomason \cite{thomason_combinations_1984}: a proposition is a historical necessity if it is \emph{inevitable}, regardless of the future actions of agents.
We refer to \cite{goranko_temporal_2024} for a comprehensive discussion of them.
In this section, we briefly compare the strong historical necessity $\SHN \phi$ to the historical necessity $\Box \phi$ in these theories.

Most theories of historical necessity are based on tree models, including Prior's Ockhamist temporal logic~\cite{prior_past_1967} and the family of Full Computation Tree Logic $\mathsf{CTL^*}$~\cite{emerson_sometimes_1986}.
In these theories, the evaluation context of $\Box \phi$ is a tuple $(\MM,w,\pi)$, where $\MM$ is a tree model, $w$ is a state, and $\pi$ is a timeline passing through $w$. $\Box \phi$ is true at an evaluation context $(\MM,w,\pi)$ if $\phi$ is true at $(\MM,w,\pi')$ for all timelines $\pi'$ passing through $w$.

In addition, some theories of historical necessity are based on so-called \emph{bundled tree models}, including Burgess \cite{burgess_unreal_1978,burgess_decidability_1980}. %
In these theories, the evaluation context of $\Box \phi$ is a tuple $(\MM,\mathtt{B},w,\pi)$, where $\MM$ is a tree model, $\mathtt{B}$, called a bundle, is a set of timelines such that every state of $\MM$ occurs in some timeline in $\mathtt{B}$, $w$ is a state, and $\pi$ is a timeline in $\mathtt{B}$ passing through $w$. $\Box \phi$ is true at an evaluation context $(\MM,\mathtt{B},w,\pi)$ if $\phi$ is true at $(\MM,w,\pi')$ for all timelines $\pi'$ in $\mathtt{B}$ passing through $w$.
The evaluation context $(\MM,w,\pi)$ mentioned above can be viewed as a special case of $(\MM,\mathtt{B},w,\pi)$: the former has an implicit bundle consisting of all timelines of $\MM$.
Technically, $(\MM,\mathtt{B},w,\pi)$ makes some difference from $(\MM,w,\pi)$, about which we refer to \cite{goranko_temporal_2024}.

The strong historical necessity $\SHN$ is close to historical necessity based on bundled tree models: bundles can be viewed as a special case of sets of accepted timelines.

\part{}

\textit{%
In this part, we provide an axiomatic system for the logical theory and show its soundness and completeness.
}

\newcommand{\TbXYHb}{
{
\mathtt{T}
(
\mathtt{X}
\cdot
\mathtt{Y}
\cdot
\mathtt{H}
)
}
}

\newcommand{\XYH}{
{
\mathtt{X}
\cdot
\mathtt{Y}
\cdot
\mathtt{H}
}
}

\newcommand{\HXHY}{
{
\mathtt{HX}
\cdot
\mathtt{HY}
}
}

\newcommand{\YHX}{
{\mathtt{Y}
\cdot
\mathtt{HX}
}
}

\newcommand{\YmHX}{
{
\mathtt{Y}\text{-}\mathtt{HX}
}
}

\newcommand{\HX}{\mathtt{HX}}

\newcommand{\TbYnbHXbb}{
{
\mathtt{T} ( \mathtt{Y^n} ( \mathtt{HX} ) )
}
}

\newcommand{\YnbHXb}{
{
\mathtt{Y^n} ( \mathtt{HX} )
}
}

\newcommand{\TbHXb}{
{
\mathtt{T} ( \mathtt{HX} )
}
}

\section{Axiomatization}
\label{section: An axiomatic system for SWHTN}

Before presenting the axiomatic system for $\SWHTN$, we need to define a few things.

\begin{definition}[Moment formulas of $\Phi_\SWHTN$]

\[
\FMF ::= \bot \mid p \mid \neg \FMF \mid (\FMF \land \FMF) \mid \YYY \FMF \mid \SHN \phi \mid \WHN \phi \mid \STN \phi \mid \WTN \phi
\]
\noindent where $\phi$ is in $\Phi_\SWHTN$.

\end{definition}

The following result, which is easy to show, indicates that the truth value of a moment formula is only dependent on moments:

\begin{fact}
\label{fact:moment formulas}

Let $\FMF$ be a moment formula, $\MM$ be a model, $\CC = (\AT,\ET)$ be a context, $\pi,\pi'$ be two timleines in $\AT$, and $i$ be an instant such that $\pi[i] = \pi'[i]$. Then, $\MM,\CC,\pi, i \Vdash \FMF$ if and only if $\MM,\CC,\pi', i \Vdash \FMF$.

\end{fact}

\begin{definition}[Instant formulas of $\Phi_\SWHTN$]

\[
\FIF ::= \bot \mid \neg \FIF \mid (\FIF \land \FIF) \mid \YYY \FIF \mid \STN \phi \mid \WTN \phi
\]
\noindent where $\phi$ is in $\Phi_\SWHTN$.

\end{definition}

The following result, which is easy to prove, indicates that the truth value of an instant formula is only dependent on instants:

\begin{fact}
\label{fact:instant formulas}

Let $\FIF$ be an instant formula, $\MM$ be a model, $\CC = (\AT,\ET)$ be a context, $\pi,\pi'$ be two timleines in $\AT$, and $i$ be an instant. Then, $\MM,\CC,\pi, i \Vdash \FIF$ if and only if $\MM,\CC,\pi', i \Vdash \FIF$.

\end{fact}

\newcommand{\yd}{\mathsf{yd}}

\begin{definition}[$\YYY$-depth]

For every $\phi$ in $\Phi_\SWHTN$, define $\yd (\phi)$, the \Fdefs{$\YYY$-depth} of $\phi$, as follows:
\begin{itemize}

\item $\yd (\bot) = \yd (p) = 0$

\item $\yd (\neg \psi) = \yd (\psi)$

\item $\yd (\psi \land \chi) = \max \{\yd (\psi), \yd (\chi)\}$

\item $\yd (\XXX \psi) = \yd (\psi)$

\item $\yd (\YYY \psi) = \yd (\psi) + 1$

\item $\yd (\SHN \psi) = \yd (\WHN \psi) = \yd (\STN \psi) = \yd (\WTN \psi) = \yd (\psi)$

\end{itemize}

\end{definition}

\begin{definition}[The language $\Phi_{\HX}$]

\[\phi ::= \bot \mid p \mid \neg \phi \mid (\phi \land \phi) \mid \SHN \XXX \phi \mid \WHN \XXX \phi \]

\end{definition}

\medskip

\begin{definition}[An axiomatic system for $\SWHTN$]
\label{definition:An axiomatic system for SWHTN}

Define an axiomatic system for $\SWHTN$ as follows:

\noindent Axioms:
\begin{enumerate}[label=(\arabic*),leftmargin=3.33em]

\item 

Propositional tautologies

\item Axioms for $\XXX$:
\begin{enumerate}

\item
\label{axiom:X K}

$\XXX (\phi \rightarrow \psi) \rightarrow (\XXX \phi \rightarrow \XXX \psi)$

\emph{This is the K axiom of normal modal logic for $\XXX$.}

\item $\neg \XXX \neg \phi \rightarrow \XXX \phi$

\emph{This axiom indicates that the relation connected to $\XXX$ is a partial function.}

\item
\label{axiom: X phi not X not phi}

$\XXX \phi \rightarrow \neg \XXX \neg \phi$

\emph{This axiom indicates that the relation connected to $\XXX$ is serial.}

\end{enumerate}

\item Axioms for $\YYY$:
\begin{enumerate}

\item $\YYY (\phi \rightarrow \psi) \rightarrow (\YYY \phi \rightarrow \YYY \psi)$

\emph{This is the K axiom of normal modal logic for $\YYY$.}

\item
\label{axiom:Y partial function}

$\neg \YYY \neg \phi \rightarrow \YYY \phi$

\emph{This axiom indicates that the relation connected to $\YYY$ is a partial function.}

\end{enumerate}

\item Axioms for $\XXX$ and $\YYY$:
\begin{enumerate}

\item
\label{axiom: phi XYp phi}

$\phi \rightarrow \XXX \neg \YYY \neg \phi$

\emph{This axiom indicates that the converse of the relation connected to $\XXX$ is a subset of the relation connected to $\YYY$.}

\item
\label{axiom: phi YXp phi}

$\phi \rightarrow \YYY \neg \XXX \neg \phi$

\emph{This axiom indicates that the converse of the relation connected to $\YYY$ is a subset of the relation connected to $\XXX$.}

\end{enumerate}

\emph{The two axioms together indicate that the relation connected to $\XXX$ and the relation connected to $\YYY$ are converses of each other.}

\item 

Axioms for $\SHN$:
\begin{enumerate}

\item
\label{axiom:SHN K}

$\SHN (\phi \rightarrow \psi) \rightarrow (\SHN \phi \rightarrow \SHN \psi)$

\emph{This is the K axiom of normal modal logic for $\SHN$.}

\item
\label{axiom:moment formula}

$\SHN (\FMF \rightarrow \phi) \rightarrow (\FMF \rightarrow \SHN \phi)$

\emph{Intuitively, this axiom indicates that $\SHN$ is ineffective on moment formulas in its scope.}

\item
\label{axiom:SHN T}

$\SHN \phi \rightarrow \phi$

\emph{This is the T axiom of normal modal logic for $\SHN$.}

\end{enumerate}

\item 

Axioms for $\WHN$:
\begin{enumerate}

\item\label{axiom:WHN K}

$\WHN (\phi \rightarrow \psi) \rightarrow (\WHN \phi \rightarrow \WHN \psi)$

\emph{This is the K axiom of normal modal logic for $\WHN$.}

\item
\label{axiom:moment formula WHN}

$\WHN (\FMF \rightarrow \phi) \rightarrow (\FMF \rightarrow \WHN \phi)$

\emph{Intuitively, this axiom indicates that $\WHN$ is ineffective on moment formulas in its scope.}

\end{enumerate}

\item 

Axioms for inserting $\SHN$ and $\WHN$: 
\begin{enumerate}

\item
\label{axiom:inserting SHN}

$\SHN \XXX \phi \leftrightarrow \SHN \XXX \SHN \phi$

\item 
\label{axiom:inserting WHN}

$\WHN \XXX \phi \leftrightarrow \WHN \XXX \WHN \phi$

\end{enumerate}

\item

Axioms for interactions between $\SHN$ and $\WHN$:
\begin{enumerate}

\item 
\label{axiom:SHN is stronger than WHN}

$\SHN \phi \rightarrow \WHN \phi$

\emph{This axiom indicates that $\SHN$ is at least as strong as $\WHN$.}

\item 
\label{axiom:defining WHN by SHN}

$\WHN \XXX \FMF \leftrightarrow \SHN \XXX ( \WHP \XXX \top \rightarrow \FMF)$

\emph{This axiom indicates that $\WHN \XXX$ can be defined by $\SHN \XXX$ and $\WHP \XXX \top$, given that only moment formulas are considered.}

\end{enumerate}

\item 

Axioms for $\STN$:
\begin{enumerate}

\item

$\STN (\phi \rightarrow \psi) \rightarrow (\STN \phi \rightarrow \STN \psi)$

\emph{This is the K axiom of normal modal logic for $\STN$.}

\item\label{axiom:instant formula strong T}

$\STN (\FIF \rightarrow \phi) \rightarrow (\FIF \rightarrow \STN \phi)$

\emph{Intuitively, this axiom indicates that $\STN$ is ineffective on instant formulas in its scope.}

\end{enumerate}

\item 

Axioms for $\WTN$:
\begin{enumerate}

\item
\label{axiom:WTN K}

$\WTN (\phi \rightarrow \psi) \rightarrow (\WTN \phi \rightarrow \WTN \psi)$

\emph{This is the K axiom of normal modal logic for $\WTN$.}

\item

$\WTN (\FIF \rightarrow \phi) \rightarrow (\FIF \rightarrow \WTN \phi)$

\emph{Intuitively, this axiom indicates that $\WTN$ is ineffective on instant formulas in its scope.}

\end{enumerate}

\item 

Axioms for swapping places of $\XXX$/$\YYY$ and $\STN$/$\WTN$:
\begin{enumerate}

\item
\label{axiom:X STN}

$\XXX \STN \phi \leftrightarrow \STN \XXX \phi$

\item 

\label{axiom:Y STN}

$\YYY \STN \phi \leftrightarrow \STN \YYY \phi$

\item
\label{axiom:X WTN}

$\XXX \WTN \phi \leftrightarrow \WTN \XXX \phi$

\item
\label{axiom:Y WTN}

$\YYY \WTN \phi \leftrightarrow \WTN \YYY \phi$

\end{enumerate}

\item 

Axioms for conditionally defining $\STN$/$\WTN$ by $\SHN$/$\WHN$, $\XXX$, and $\YYY$:
\begin{enumerate}

\item 
\label{axiom:strong T replacement}

$\YYY'^n \Froot \rightarrow (\STN \phi \leftrightarrow \YYY^n \SHN \XXX^n \phi)$

\item 
\label{axiom:weak T replacement}

$\YYY'^n \Froot \rightarrow (\WTN \phi \leftrightarrow \YYY^n \WHN \XXX^n \phi)$

\end{enumerate}

\emph{Note that this does not mean $\STN \phi$ or $\WTN \phi$ is definable.}

\end{enumerate}

\noindent Inference rules:
\begin{enumerate}[label=(\arabic*),leftmargin=3.33em]

\item 

Modus ponens:

from $\phi$ and $\phi \rightarrow \psi$, we can get $\psi$.

\item

Necessitation of $\Box$, where $\Box \in \{\XXX, \YYY, \STN, \WTN, \SHN, \WHN\}$:

from $\phi$, we can get $\Box \phi$.

\item

From-$\YYY$-depth-to-derivability:

from $(\Froot \lor \YYY' \Froot \lor \dots \lor \YYY'^{n+1} \Froot) \rightarrow \phi$, we can get $\phi$, where $n$ is the $\YYY$-depth of $\phi$.

\item

From-root-to-derivability:

from $\YYY \chi \rightarrow \phi$, we can get $\phi$, where $\phi$ is in $\Phi_\HX$.

\end{enumerate}

\end{definition}

We use $\vdash \phi$ to indicate $\phi$ is \Fdefs{derivable} in this system.

\begin{theorem}[Soundness and completeness of $\SWHTN$]

The axiomatic system for $\SWHTN$ given in Definition \ref{definition:An axiomatic system for SWHTN} is sound and complete with respect to the set of valid formulas of $\Phi_{\SWHTN}$.

\end{theorem}

The proof for this result is in the Appendix.

\section{Concluding remarks}
\label{section: Looking backward and forward}

In this paper, we undertake three tasks.
First, we distinguish four notions of necessity and two notions of possibility, namely strong/weak historical/temporal necessities and historical/temporal possibilities, and discuss their features.
Second, we present our approach to the six notions, which is based on ontic systems, and following this approach, we develop a logical theory for them in branching time.
Third, we provide an axiomatic system for the theory and show its soundness and completeness.

There are two kinds of work worth doing in the future.
Somehow, strong historical and temporal necessities do not receive strong support in English. The situation with Chinese seems different. This deserves investigation.  
Another work is to address conditional strong/weak historical/temporal necessities. As mentioned above, our general approach to them is to introduce \emph{hypothesis sets} into contexts and to define how contexts determine accepted timelines and expected timelines.

\subsection*{Acknowledgments}

We thank Valentin Goranko, Maria Aloni, Luca Incurvati, Paul McNamara, Frank Veltman, and Hugh Reilly for their help with this paper.
Thanks also go to the audience at Nicolaus Copernicus University in Toru\'{n}, Beijing Normal University, Shandong University, Hubei University, Hebei University, and Tsinghua University.

\bibliographystyle{alpha}
\bibliography{Modalities,Special-items}

\appendix

\section{Soundness}
\label{section: Soundness}

Before showing the soundness of $\SWHTN$, we need to define two notions and show two lemmas.

\begin{definition}[Cutted contextualized pointed models]

Let $(\MM,\CC,\pi,i)$ be a contextualized pointed model, where $\MM = (W,r,<,V)$, $\CC = (\AT, \ET)$, and $0 < i$. Assume $n < i$ and $w$ is a state not in $W$.

Define a contextualized pointed model $(\MM^\geq_{i-n}, \CC^\geq_{i-n}, \pi^\geq_{i-n}, n+1)$, called the \Fdefs{cutted contextualized pointed model} from $(\MM,\CC,\pi,i)$ with respect to $n$ and $w$, as follows:
\begin{itemize}

\item 

$\MM^\geq_{i-n} = (W',w,<',V')$, where:
\begin{itemize}

\item 

$W' = X \cup \{w\}$, where $X = \{w' \in W \mid w' = \pi'[j] \text{ for some } \pi' \in \mathtt{TL} (\MM,r) \text{ and } j \in \mathbb{N} \text{ such that } i - n \leq j \}$;

\item $<' = \Big( < \cap (X \times X) \Big) \cup \{(w,w') \mid w' \in Y\}$, where $Y = \{w' \in W \mid w' = \pi'[i-n] \text{ for some } \pi' \in \mathtt{TL} (\MM,r) \}$;

\item $V' (p) = V (p) \cap X$ for every $p$.

\end{itemize}

\emph{Intuitively, $\MM^\geq_{i-n}$ is the result of removing states of $\MM$ with instants less than $i - n$ and adding $w$ as the root.}

\item 

$\CC^\geq_{i-n} = (\AT',\ET')$, where:
\begin{itemize}

\item

$\AT' = \{(w, \pi' [i-n,\infty]) \mid \pi' \in \AT \}$;

\item

$\ET' = \{(w, \pi' [i-n,\infty]) \mid \pi' \in \ET \}$.

\end{itemize}

\item 

$\pi^\geq_{i-n} = (w, \pi[i-n,\infty])$.

\end{itemize}

\end{definition}

\begin{lemma}
\label{lemma:cutting}

Let $\phi$ be in $\Phi_\SWHTN$ and its $\YYY$-depth be $n$.
Let $(\MM,\CC,\pi,i)$ be a contextualized pointed model, where $n < i$.
Let $w$ be a state not in the domain of $\MM$.
Let $(\MM^\geq_{i-n}, \CC^\geq_{i-n}, \pi^\geq_{i-n}, n+1)$ be the \Fdefs{cutted} contextualized pointed model from $(\MM,\CC,\pi,i)$ with respect to $n$ and $w$.

Then, $\MM,\CC,\pi,i \Vdash \phi$ if and only if $\MM^\geq_{i-n}, \CC^\geq_{i-n}, \pi^\geq_{i-n}, n+1 \Vdash \phi$.

\end{lemma}

It can be checked $\pi[i] = \pi^\geq_{i-n}[n+1]$. Then, this lemma is easy to verify, and we skip its proof.

\begin{definition}[Generated contextualized submodels]

Let $(\MM,\CC)$ be a contextualized model, where $\CC = (\AT,\ET)$, $w$ be a state of $\MM$, and $i$ be the instant of $w$.

Define $(\MM^w, \CC^w)$, the \Fdefs{generated contextualized submodel} of $(\MM,\CC)$ from $w$, as follows:
\begin{itemize}

\item

$\MM^w$ is the generated submodel of $\MM$ from $w$;

\item $\CC^w = (\AT^w,\ET^w)$, where:
\begin{itemize}

\item

$\AT^w = \{\pi[i,\infty] \mid \pi \in \AT \text{ and } \pi[i] = w\}$;

\item

$\ET^w = \{\pi[i,\infty] \mid \pi \in \ET \text{ and } \pi[i] = w\}$.

\end{itemize}

\end{itemize}

\end{definition}

\begin{lemma}
\label{lemma:generated submodels}

Let $(\MM,\CC,\pi,i)$ be a contextualized pointed model, and $(\MM^{\pi[i]}, \CC^{\pi[i]})$ be the generated contextualized submodel of $(\MM,\CC)$ from $\pi[i]$.

Then, for every $\phi$ in $\Phi_\HX$, $\MM,\CC,\pi,i \Vdash \phi$ if and only if $\MM^{\pi[i]},\CC^{\pi[i]},\pi[i-\infty],0 \Vdash \phi$.

\end{lemma}

This lemma is easy to verify, and we skip its proof.

\begin{theorem}[Soundness of $\SWHTN$]

The axiomatic system for $\SWHTN$ given in Definition \ref{definition:An axiomatic system for SWHTN} is sound with respect to the set of valid formulas of $\Phi_{\SWHTN}$.

\end{theorem}

\begin{proof}~

It suffices to show that all axioms are valid and all inference rules preserve validity. In what follows, we only show some of these cases; other cases are either similar or easy. We assume an arbitrary contextualized pointed model $(\MM, \CC, \pi, i)$, where $\CC = (\AT, \ET)$.
\begin{itemize}

\item

\textbf{The axiom 5b, that is, $\SHN (\FMF \rightarrow \phi) \rightarrow (\FMF \rightarrow \SHN \phi)$, is valid.}

Assume $\MM, \CC, \pi, i \not \Vdash \FMF \rightarrow \SHN \phi$. Then, $\MM, \CC, \pi, i \Vdash \FMF$ and $\MM, \CC, \pi, i \not \Vdash \SHN \phi$.
Then there is $\pi' \in \AT_{\pi[i]}$ such that $\MM, \CC, \pi', i \not \Vdash \phi$. Note that $\FMF$ is a moment formula. By Fact \ref{fact:moment formulas}, $\MM, \CC, \pi', i \Vdash \FMF$. Then $\MM, \CC, \pi', i \not \Vdash \FMF \rightarrow \phi$. Then $\MM, \CC, \pi, i \not \Vdash \SHN (\FMF \rightarrow \phi)$.

\item

\textbf{The axiom 7a, that is, $\SHN \XXX \phi \leftrightarrow \SHN \XXX \SHN \phi$, is valid.}

Assume $\MM, \CC, \pi, i \not \Vdash \SHN \XXX \phi$. Then there is $\pi' \in \AT_{\pi[i]}$ such that $\MM, \CC, \pi', i+1 \not \Vdash \phi$.
Note $\pi' \in \AT_{\pi'[i+1]}$. Then $\MM, \CC, \pi', i+1 \not \Vdash \SHN \phi$. Then $\MM, \CC, \pi', i \not \Vdash \XXX \SHN \phi$. Then $\MM, \CC, \pi, i \not \Vdash \SHN \XXX \SHN \phi$.

Assume $\MM, \CC, \pi, i \not \Vdash \SHN \XXX \SHN \phi$. Then there is $\pi' \in \AT_{\pi[i]}$ such that $\MM, \CC, \pi', i+1 \not \Vdash \SHN \phi$.
Then there is $\pi'' \in \AT_{\pi'[i+1]}$ such that $\MM, \CC, \pi'', i+1 \not \Vdash \phi$. Then $\MM, \CC, \pi'', i \not \Vdash \XXX \phi$. It is easy to check $\pi'' \in \AT_{\pi[i]}$. Then $\MM, \CC, \pi, i \not \Vdash \SHN \XXX \phi$.

\item

\textbf{The axiom 8b, that is, $\WHN \XXX \FMF \leftrightarrow \SHN \XXX ( \WHP \XXX \top \rightarrow \FMF)$, is valid.}

Assume $\MM, \CC, \pi, i \not \Vdash \WHN \XXX \FMF$. Then there is $\pi' \in \ET_{\pi[i]}$ such that $\MM, \CC, \pi', i+1 \not \Vdash \FMF$. It is easy to see $\MM, \CC, \pi', i+1 \Vdash \WHP \XXX \top$. Then, $\MM, \CC, \pi', i+1 \not \Vdash \WHP \XXX \top \rightarrow \FMF$. Note $\pi' \in \AT_{\pi[i]}$. Then $\MM, \CC, \pi, i \not \Vdash \SHN \XXX ( \WHP \XXX \top \rightarrow \FMF)$.

Assume $\MM, \CC, \pi, i \not \Vdash \SHN \XXX ( \WHP \XXX \top \rightarrow \FMF )$. Then there is $\pi' \in \AT_{\pi[i]}$ such that $\MM, \CC, \pi', i+1 \not \Vdash \WHP \XXX \top \rightarrow \FMF$. Then $\MM, \CC, \pi', i+1 \Vdash \WHP \XXX \top$ and $\MM, \CC, \pi', i+1 \not \Vdash \FMF$.
Then, there is $\pi'' \in \ET_{\pi' [i+1]}$. Note $\FMF$ is a moment formula. Then, $\MM, \CC, \pi'', i+1 \not \Vdash \FMF$. Note $\pi'' \in \ET_{\pi [i]}$. Then $\MM, \CC, \pi, i \not \Vdash \WHN \XXX \FMF$.

\item

\textbf{The axiom 9b, that is, $\STN (\FIF \rightarrow \phi) \rightarrow (\FIF \rightarrow \STN \phi)$, is valid.}

Assume $\MM, \CC, \pi, i \not \Vdash \FIF \rightarrow \STN \phi$. Then $\MM, \CC, \pi, i \Vdash \FIF$ and $\MM, \CC, \pi, i \not \Vdash \STN \phi$.
Then there is $\pi' \in \AT$ such that $\MM, \CC, \pi', i \not \Vdash \phi$. Note $\FIF$ is an instant formula. By Fact \ref{fact:instant formulas}, $\MM, \CC, \pi', i \Vdash \FIF$. Then $\MM, \CC, \pi', i \not \Vdash \FIF \rightarrow \phi$. Then $\MM, \CC, \pi, i \not \Vdash \STN (\FIF \rightarrow \phi)$.

\item

\textbf{The axiom 11a, that is, $\XXX \STN \phi \leftrightarrow \STN \XXX \phi$, is valid.}

Assume $\MM, \CC, \pi, i \not \Vdash \XXX \STN \phi$. Then $\MM, \CC, \pi, i+1 \not \Vdash \STN \phi$.
Then there is $\pi' \in \AT$ such that $\MM, \CC, \pi', i+1 \not \Vdash \phi$. Then $\MM, \CC, \pi', i \not \Vdash \XXX \phi$. Then $\MM, \CC, \pi, i \not \Vdash \STN \XXX \phi$.

Assume $\MM, \CC, \pi, i \not \Vdash \STN \XXX \phi$. Then there is $\pi' \in \AT$ such that $\MM, \CC, \pi', i+1 \not \Vdash \phi$. Then $\MM, \CC, \pi, i+1 \not \Vdash \STN \phi$. Then $\MM, \CC, \pi, i \not \Vdash \XXX \STN \phi$.

\item

\textbf{The axiom 11b, that is, $\YYY \STN \phi \leftrightarrow \STN \YYY \phi$, is valid.}

Assume $\MM, \CC, \pi, i \not \Vdash \YYY \STN \phi$. Then $i > 0$ and $\MM, \CC, \pi, i-1 \not \Vdash \STN \phi$.
Then there is $\pi' \in \AT$ such that $\MM, \CC, \pi', i-1 \not \Vdash \phi$. Then $\MM, \CC, \pi', i \not \Vdash \YYY \phi$. Then $\MM, \CC, \pi, i \not \Vdash \STN \YYY \phi$.

Assume $\MM, \CC, \pi, i \not \Vdash \STN \YYY \phi$. Then there is $\pi' \in \AT$ such that $\MM, \CC, \pi', i \not \Vdash \YYY \phi$. Then, $i > 0$ and $\MM, \CC, \pi', i-1 \not \Vdash \phi$. Then $\MM, \CC, \pi, i-1 \not \Vdash \STN \phi$. Then $\MM, \CC, \pi, i \not \Vdash \YYY \STN \phi$.

\item

\textbf{The axiom 12a, that is, $\YYY'^n \Froot \rightarrow (\STN \phi \leftrightarrow \YYY^n \SHN \XXX^n \phi)$, is valid.}

Assume $\MM, \CC, \pi, i \Vdash \YYY'^n \Froot$. Then $i = n$.

Assume $\MM, \CC, \pi, i \not \Vdash \STN \phi$. Then $\MM, \CC, \pi', i \not \Vdash \phi$ for some $\pi' \in \AT$.
Then $\MM, \CC, \pi', 0 \not \Vdash \XXX^n \phi$. Note $\AT = \AT_{\pi[0]}$. Then $\pi' \in \AT_{\pi[0]}$. Then $\MM, \CC, \pi, 0 \not \Vdash \SHN \XXX^n \phi$. Then $\MM, \CC, \pi, i \not \Vdash \YYY^n \SHN \XXX^n \phi$.

Assume $\MM, \CC, \pi, i \not \Vdash \YYY^n \SHN \XXX^n \phi$. Then $\MM, \CC, \pi, 0 \not \Vdash \SHN \XXX^n \phi$. Then $\MM, \CC, \pi', 0 \not \Vdash \XXX^n \phi$ for some $\pi' \in \AT_{\pi[0]}$. Then $\MM, \CC, \pi', i \not \Vdash \phi$. Note $\pi' \in \AT$. Then $\MM, \CC, \pi, i \not \Vdash \STN \phi$.

\item

\textbf{The rule of from-$\YYY$-depth-to-derivability preserves validity}.

Let $\phi \in \Phi_\SWHTN$ and its $\YYY$-depth be $n$. Assume $\not \models \phi$. Then, there is a contextualized pointed model $(\MM,\CC,\pi,i)$ such that $\MM,\CC,\pi,i \not \Vdash \phi$.
It suffices to show $\not \models (\Froot \lor \YYY' \Froot \lor \dots \lor \YYY'^{n+1} \Froot) \rightarrow \phi$.

Assume $i \leq n + 1$. Then $\MM,\CC,\pi,i \Vdash (\Froot \lor \YYY' \Froot \lor \dots \lor \YYY'^{n+1} \Froot)$. Then $\MM,\CC,\pi,i \not \Vdash (\Froot \lor \YYY' \Froot \lor \dots \lor \YYY'^{n+1} \Froot) \rightarrow \phi$. Then $\not \models (\Froot \lor \YYY' \Froot \lor \dots \lor \YYY'^{n+1} \Froot) \rightarrow \phi$.

Assume $n + 1 < i$.
Let $w$ be a state not in the domain of $\MM$.
Let $(\MM^\geq_{i-n}, \CC^\geq_{i-n}, \pi^\geq_{i-n}, n+1)$ be the cutted contextualized pointed model from $(\MM,\CC,\pi,i)$ with respect to $n$ and $w$.
By Lemma \ref{lemma:cutting}, $\MM^\geq_{i-n}, \CC^\geq_{i-n}, \pi^\geq_{i-n}, n+1 \not \Vdash \phi$. Note $\MM^\geq_{i-n}, \CC^\geq_{i-n}, \pi^\geq_{i-n}, n+1 \Vdash \YYY'^{n+1} \Froot$. 
Then $\MM^\geq_{i-n}, \CC^\geq_{i-n}, \pi^\geq_{i-n}, n+1 \not \Vdash (\Froot \lor \YYY' \Froot \lor \dots \lor \YYY'^{n+1} \Froot) \rightarrow \phi$. Then $\not \models (\Froot \lor \YYY' \Froot \lor \dots \lor \YYY'^{n+1} \Froot) \rightarrow \phi$.

\item

\textbf{The rule of from-root-to-derivability preserves validity}.

Let $\chi \in \Phi_\SWHTN$ and $\phi \in \Phi_{\HX}$. Assume $\not \models \phi$. It suffices to show $\not \models \YYY \chi \rightarrow \phi$.

Let $(\MM,\CC,\pi,i)$ be a contextualized pointed model such that $\MM,\CC,\pi,i \not \Vdash \phi$.
Let $(\MM^{\pi[i]}, \CC^{\pi[i]})$ be the generated contextualized submodel of $(\MM,\CC)$ from $\pi[i]$. By Lemma \ref{lemma:generated submodels}, $\MM^{\pi[i]},\CC^{\pi[i]},\pi[i-\infty],0 \not \Vdash \phi$. Note trivially, $\MM^{\pi[i]},\CC^{\pi[i]},\pi[i-\infty],0 \Vdash \YYY \chi$.
Then $\MM^{\pi[i]},\CC^{\pi[i]},\pi[i-\infty],0 \not \Vdash \YYY \chi \rightarrow \phi$.
Then $\not \models \YYY \chi \rightarrow \phi$.

\end{itemize}

\end{proof}

\section{Completeness}
\label{section: Completeness}

The completeness proof is a bit lengthy, and we provide a roadmap here.

\begin{definition}[V-reducibility]

Let $\Phi_1$ and $\Phi_2$ be two sub-languages of $\Phi_\SWHTN$ such that $\Phi_1 \supseteq \Phi_2$. We say $\Phi_1$ is \Fdefs{v-reducible} to $\Phi_2$ ($\Phi_1 \rightleftharpoons \Phi_2$) if for every $\phi \in \Phi_1$, there is $\phi' \in \Phi_2$ such that (1) if $\models \phi$, then $\models \phi'$, and (2) if $\vdash \phi'$, then $\vdash \phi$.

\end{definition}

We will consider the following sequence of languages, where the latter is a sub-language of the former: (0) $\Phi_\SWHTN$, (1) $\Phi_\TbXYHb$, (2) $\Phi_{\XYH}$, (3) $\Phi^{\mathtt{mo}}_{\XYH}$, (4) $\Phi_\YHX$, (5) $\Phi_\YmHX$, (6) $\Phi_\YnbHXb$, and (7) $\Phi_\HX$.
We will show that (1) the former is v-reducible to the latter, and (2) for every $\phi \in \Phi_\HX$, if $\models \phi$, then $\vdash \phi$.
Then the completeness of $\SWHTN$ follows: for every $\phi \in \Phi_\SWHTN$, if $\models \phi$, then $\vdash \phi$.

\subsection{V-reducibility of $\Phi_\SWHTN$ to $\Phi_\TbXYHb$}

\begin{definition}[Language $\Phi_{\XYH}$]

\[
\phi ::= \bot \mid p \mid \neg \phi \mid (\phi \land \phi) \mid \XXX \phi \mid \YYY \phi \mid \SHN \phi \mid \WHN \phi
\]

\end{definition}

\begin{definition}[Language $\Phi_\TbXYHb$]

\[\psi ::= \phi \mid \STN \phi \mid \WTN \phi \mid \neg \psi \mid (\psi \land \psi)\]
where $\phi$ is in $\Phi_\XYH$.

\end{definition}

Note in $\Phi_\TbXYHb$: the two $\mathtt{T}$ operators do not occur in the scope of any operators, including themselves.

The following lemma offers some basic facts about derivability, which will be used often in the sequel:

\begin{lemma}
\label{lemma:make T out}
~

\begin{enumerate}[label=(\arabic*),leftmargin=3.33em]

\item 

For every $\phi, \psi, \psi'$ and $\phi'$ in $\Phi_\SWHTN$, if $\vdash \psi \leftrightarrow \psi'$, then $\vdash \phi \leftrightarrow \phi'$, where $\phi'$ is the result of replacing $\psi$ in $\phi$ by $\psi'$.

\emph{This is the rule of \emph{replacement of equivalence}}.

\item

\begin{enumerate}

\item\label{lemma item: not X not}

$\vdash \XXX \phi \leftrightarrow \neg \XXX \neg \phi$

\emph{This means the dual of $\XXX$ is itself}.

\item 

$\vdash \YYY' \phi \leftrightarrow \neg \YYY \bot \land \YYY \phi$

\end{enumerate}

\item 

\begin{enumerate}

\item
\label{derivative formulas: box conjunction}

$\vdash \Box (\phi \land \psi) \leftrightarrow (\Box \phi \land \Box \psi)$, where $\Box \in \{\XXX, \YYY, \YYY', \SHN, \WHN, \STN, \WTN\}$.

\item
\label{derivative formulas: diamond disjunction}

$\vdash \Diamond (\phi \lor \psi) \leftrightarrow (\Diamond \phi \lor \Diamond \psi)$, where $\Diamond \in \{\XXX, \YYY, \YYY', \SHP, \WHP, \STP, \WTP\}$.

\end{enumerate}

\item 

\begin{enumerate}

\item
\label{derivative formulas: box T disjunction}

$\vdash \Box (\nabla \phi \lor \psi) \leftrightarrow (\nabla \phi \lor \Box \psi)$, where $\Box \in \{\SHN, \WHN, \STN, \WTN \}$ and $\nabla \in \{ \STN, \STP, \WTN, \WTP \}$.

\item
\label{derivative formulas: diamond T conjunction}

$\vdash \Diamond (\nabla \phi \land \psi) \leftrightarrow (\nabla \phi \land \Diamond \psi)$, where $\Diamond \in \{\SHP, \WHP, \STP, \WTP \}$ and $\nabla \in \{ \STN, \STP, \WTN, \WTP \}$.

\end{enumerate}

\item 

$\vdash \nabla_1 \nabla_2 \phi \leftrightarrow \nabla_2 \nabla_1 \phi$, where $\nabla_1 \in \{\XXX,\YYY,\YYY'\}$ and $\nabla_2 \in \{\STN,\STP,\WTN,\WTP\}$.

\end{enumerate}

\end{lemma}

\begin{proof}~

\begin{enumerate}[label=(\arabic*),leftmargin=3.33em]

\item 

Let $\phi, \psi, \psi'$ and $\phi'$ be in $\Phi_\SWHTN$ such that $\psi$ is a subformula of $\phi$, $\vdash \psi \leftrightarrow \psi'$, and $\phi'$ is the result of replacing $\psi$ in $\phi$ by $\psi'$. We want to show $\vdash \phi \leftrightarrow \phi'$.
We put an induction on the structure of $\phi$. We consider only the case $\phi = \XXX \chi$; other cases are either easy or similar.

Assume $\psi = \XXX \chi$. Clearly, $\vdash \phi \leftrightarrow \phi'$.
Assume $\psi$ is a subformula of $\chi$. By the inductive hypothesis, $\vdash \chi \leftrightarrow \chi'$, where $\chi'$ is the result of replacing $\psi$ in $\chi$ by $\psi'$. It is easy to show $\vdash \XXX \chi \leftrightarrow \XXX \chi'$, that is, $\vdash \phi \leftrightarrow \phi'$.

\item

\begin{enumerate}

\item Easy.

\item 

By the axiom \ref{axiom:Y partial function}, $\vdash \neg \YYY \neg \phi \rightarrow \YYY \phi$.
It is easy to show $\vdash \YYY \bot \rightarrow \YYY \neg \phi$. Then $\vdash \neg \YYY \neg \phi \rightarrow \neg \YYY \bot$.
Then $\vdash \neg \YYY \neg \phi \rightarrow (\neg \YYY \bot \land \YYY \phi)$, that is, $\vdash \YYY' \phi \rightarrow (\neg \YYY \bot \land \YYY \phi)$.

It is easy to show $\vdash \YYY \phi \rightarrow (\YYY \neg \phi \rightarrow \YYY \bot)$. Then $\vdash \YYY \phi \rightarrow (\neg \YYY \bot \rightarrow \neg \YYY \neg \phi)$, that is, $\vdash \YYY \phi \rightarrow (\neg \YYY \bot \rightarrow \YYY' \phi)$. Then $\vdash (\YYY \phi \land \neg \YYY \bot) \rightarrow \YYY' \phi$.

\end{enumerate}

\item

\begin{enumerate}

\item

We only show $\vdash \YYY' (\phi \land \psi) \leftrightarrow (\YYY' \phi \land \YYY' \psi)$; other cases are either easy or similar.

It suffices to show $\vdash \YYY (\neg \phi \lor \neg \psi) \leftrightarrow (\YYY \neg \phi \lor \YYY \neg \psi)$. It is easy to show $\vdash (\YYY \neg \phi \lor \YYY \neg \psi) \rightarrow \YYY (\neg \phi \lor \neg \psi)$.

To show $\vdash \YYY (\neg \phi \lor \neg \psi) \rightarrow (\YYY \neg \phi \lor \YYY \neg \psi)$,
it suffices to show $\vdash \YYY (\neg \phi \lor \neg \psi) \rightarrow (\neg \YYY \neg \phi \rightarrow \YYY \neg \psi)$.
It is easy to show $\vdash \big(\YYY (\neg \phi \lor \neg \psi) \land \YYY \phi\big) \rightarrow \YYY \neg \psi$.
By the axiom \ref{axiom:Y partial function}, $\vdash \neg \YYY \neg \phi \rightarrow \YYY \phi$. Then, $\vdash \big(\YYY (\neg \phi \lor \neg \psi) \land \neg \YYY \neg \phi \big) \rightarrow \YYY \neg \psi$.
Then, $\vdash \YYY (\neg \phi \lor \neg \psi) \rightarrow ( \neg \YYY \neg \phi \rightarrow \YYY \neg \psi)$.

\item Easy.

\end{enumerate}

\item 

\begin{enumerate}

\item 

We only show $\vdash \SHN (\STN \phi \lor \psi) \leftrightarrow (\STN \phi \lor \SHN \psi)$; other cases are either easy or similar.

Note $\vdash \SHN (\STN \phi \lor \psi) \rightarrow \SHN (\neg \STN \phi \rightarrow \psi)$. By the axiom \ref{axiom:moment formula}, $\vdash \SHN (\neg \STN \phi \rightarrow \psi) \rightarrow (\neg \STN \phi \rightarrow \SHN \psi)$.
Note $\vdash (\neg \STN \phi \rightarrow \SHN \psi) \rightarrow (\STN \phi \lor \SHN \psi)$. Then $\vdash \SHN (\STN \phi \lor \psi) \rightarrow (\STN \phi \lor \SHN \psi)$.

Note $\vdash \SHN (\STN \phi \rightarrow \STN \phi)$. By the axiom \ref{axiom:moment formula}, $\vdash \STN \phi \rightarrow \SHN \STN \phi$. Note $\vdash \SHN \STN \phi \rightarrow \SHN (\STN \phi \lor \psi)$. Then, $\vdash \STN \phi \rightarrow \SHN (\STN \phi \lor \psi)$.
It is easy to show $\vdash \SHN \psi \rightarrow \SHN (\STN \phi \lor \psi)$.
Then $\vdash (\STN \phi \lor \SHN \psi) \rightarrow \SHN (\STN \phi \lor \psi)$.

\item

Easy.

\end{enumerate}

\item 

We only show $\vdash \YYY' \STN \phi \leftrightarrow \STN \YYY' \phi$; other cases are either easy or similar.

The following is easy to see:
\begin{itemize}

\item 

$\vdash \YYY' \STN \phi$
$\leftrightarrow$
$(\neg \YYY \bot \land \YYY \STN \phi)$ (by (2b) in this lemma)

\item

$\vdash (\neg \YYY \bot \land \YYY \STN \phi)$
$\leftrightarrow$
$(\neg \YYY \bot \land \STN \YYY \phi)$ (by the axiom \ref{axiom:Y STN})

\item 

$\vdash (\neg \YYY \bot \land \STN \YYY \phi)$
$\leftrightarrow$
$\STN (\neg \YYY \bot \land \YYY \phi)$ (by using $\vdash \neg \YYY \bot \leftrightarrow \STN \neg \YYY \bot$, which can be shown by the axiom \ref{axiom:instant formula strong T})

\item 

$\vdash \STN (\neg \YYY \bot \land \YYY \phi)$
$\leftrightarrow$
$\STN \YYY' \phi$ (by (2b) in this lemma)

\end{itemize}

It follows $\vdash \YYY' \STN \phi \leftrightarrow \STN \YYY' \phi$.

\end{enumerate}

\end{proof}

\begin{definition}[Direct containers]

Let $\nabla' \psi'$ be a proper subformula of $\nabla \psi$, where $\nabla$ and $\nabla'$ are modal operators. 

We say $\nabla \psi$ is a \Fdefs{direct container} of $\nabla' \psi'$ if there is no $\nabla'' \psi'' $ such that (1) $\nabla'' \psi'' $ is a proper subformula of $\nabla \psi$, and (2) $\nabla' \psi'$ is a proper subformula of $\nabla'' \psi''$, where $\nabla''$ is a modal operator.

\end{definition}

For example, $\SHN (\XXX p \lor \STN \XXX q)$ is a direct container of $\STN \XXX q$.

\newcommand{\md}{\mathsf{md}}

\begin{definition}[Modal depth]

For every $\phi$ in $\Phi_\SWHTN$, define $\md (\phi)$, the \Fdefs{modal depth} of $\phi$, as follows:
\begin{itemize}

\item $\md (\bot) = \md (p) = 0$

\item $\md (\neg \psi) = \md (\psi)$

\item $\md (\psi \land \chi) = \max \{\md (\psi), \md (\chi)\}$

\item $\md (\XXX \psi) = \md (\YYY \psi) = \md (\SHN \psi) = \md (\WHN \psi) = \md (\STN \psi) = \md (\WTN \psi) = \md (\psi) + 1$

\end{itemize}

\end{definition}

\begin{lemma}
\label{lemma:SWHTN to T(X.Y.H)}

For every $\phi$ in $\Phi_\SWHTN$, there is $\phi'$ in $\Phi_\TbXYHb$ such that $\vdash \phi \leftrightarrow \phi'$.

\end{lemma}

\begin{proof}~

Let $\phi$ be in $\Phi_\SWHTN$. In what follows, we describe a procedure by which we can get a formula $\phi'$ in $\Phi_\TbXYHb$ such that $\vdash \phi \leftrightarrow \phi'$.
\begin{enumerate}

\item

Make $\neg$ in $\phi$ to be next to $\bot$ or atomic propositions, and let $A$ be the result.

\emph{Note $A$ can be generated by propositional literals, the operators $\XXX, \YYY, \YYY', \SHN, \SHP, \WHN, \WHP, \STN,$ $\STP, \WTN, \WTP$, conjunction, and disjunction.}

\item

Go to step 3 if $A$ contains no subformula $\STN B$ meeting the following condition: \textbf{there is some subformula $\nabla C$ of $A$ such that it is a direct container of $\STN B$}.

\emph{Note that if there is no such $\STN B$, then one of the following holds: (1) $A$ contains no $\STN$; (2) $A$ contains $\STN$, but $\STN$ does not occur in the scope of any operators.}

If $A$ contains a subformula $\STN B$ meeting the condition mentioned above, pick such a $\STN B$ and do the following.
\begin{enumerate}

\item 

Assume $\nabla \in \{\XXX, \YYY, \YYY', \SHN, \WHN, \STN, \WTN \}$.
Transform $C$ to $E_1 \land \dots \land E_n$ in the conjunctive normal form.

\emph{Note $\nabla C$ is provably equivalent to $\nabla (E_1 \land \dots \land E_n)$, which is provably equivalent to $\nabla E_1 \land \dots \land \nabla E_n$ by \ref{derivative formulas: box conjunction} in Lemma \ref{lemma:make T out}.}

Replace $\nabla C$ by $\nabla E_1 \land \dots \land \nabla E_n$.

\emph{Note $\STN B$ is a subformula of some $E_i$. Let $E_i = F_1 \lor \dots \lor F_m$. Then $\STN B$ equals some $F_j$. Without loss of any generality, we can assume $\STN B = F_1$. In addition, we can assume $m \geq 2$ (note $\vdash \chi \leftrightarrow (\chi \lor \bot)$).}
\begin{itemize}

\item 

Assume $\nabla = \XXX$. \emph{By \ref{derivative formulas: diamond disjunction} in Lemma \ref{lemma:make T out}, $\XXX E_i$ is provably equivalent to $\XXX F_1 \lor \dots \lor \XXX F_m$. By the axiom \ref{axiom:X STN}, $\XXX \STN B$ is provably equivalent to $\STN \XXX B$.} Replace $\XXX \STN B$ by $\STN \XXX B$. \emph{Note here, we get $\STN$ out of the scope of $\XXX$.}

\item

Assume $\nabla = \YYY$ or $\nabla = \YYY'$. Do similar things as the case $\nabla = \XXX$.

\item 

Assume $\nabla = \SHN$. \emph{By \ref{derivative formulas: box T disjunction} in Lemma \ref{lemma:make T out}, $\SHN E_i$ is provably equivalent to $F_1 \lor \SHN ( F_2 \lor \dots \lor F_m )$.} Replace $\SHN E_i$ by $F_1 \lor \SHN ( F_2 \lor \dots \lor F_m )$. \emph{Note here, we get $\STN$ out of the scope of $\SHN$.}

\item

Assume $\nabla = \WHN$, $\nabla = \STN$, or $\nabla = \WTN$. Do similar things as the case $\nabla = \SHN$.

\end{itemize}

\item 

Assume $\nabla \in \{\SHP, \WHP, \STP, \WTP \}$.
Transform $C$ to $E_1 \lor \dots \lor E_n$ in the disjunctive normal form.

\emph{Note $\nabla C$ is equivalent to $\nabla (E_1 \lor \dots \lor E_n)$, which is equivalent to $\nabla E_1 \lor \dots \lor \nabla E_n$ by \ref{derivative formulas: diamond disjunction} in Lemma \ref{lemma:make T out}.}

Replace $\nabla C$ by $\nabla E_1 \lor \dots \lor \nabla E_n$.

\emph{Note $\STN B$ is a subformula of some $E_i$. Let $E_i = F_1 \land \dots \land F_m$. Then $\STN B$ equals some $F_j$. Without loss of any generality, we can assume $\STN B = F_1$. In addition, we can assume $m \geq 2$ (note $\vdash \chi \leftrightarrow (\chi \land \top)$).}
\begin{itemize}

\item 

Assume $\nabla = \SHP$. \emph{By \ref{derivative formulas: diamond T conjunction}, $\SHP E_i$ is provably equivalent to $F_1 \land \SHP ( F_2 \land \dots \land F_m )$.} Replace $\SHP E_i$ by $F_1 \land \SHP ( F_2 \land \dots \land F_m )$. \emph{Note here, we get $\STN$ out of the scope of $\SHP$.}

\item

Assume $\nabla = \WHP$, $\nabla = \STP$, or $\nabla = \WTP$. Do similar things as the case $\nabla = \SHP$.

\end{itemize}

\end{enumerate}

Repeat.

\emph{Note this step will terminate, as we cannot get $\STN$ out of the scope of some operator forever.}

\item

Do similar things to $\STP B$ as we do to $\STN B$ in step 2.

\item

Do similar things to $\WTN B$ as we do to $\STN B$ in step 2.

\item

Do similar things to $\WTP B$ as we do to $\STN B$ in step 2.

\end{enumerate}

\end{proof}

From this lemma and the soundness of $\SWHTN$, the following result follows:

\begin{lemma}[V-reducibility of $\Phi_\SWHTN$ to $\Phi_\TbXYHb$]
\label{lemma:v-reducibility of SWHTN to T(X.Y.H)}

$\Phi_\SWHTN$ is v-reducible to $\Phi_\TbXYHb$.

\end{lemma}

\newcommand{\FPI}{\mathbb{J}}

\subsection{V-reducibility of $\Phi_{\TbXYHb}$ to $\Phi_\XYH$}

Note the language $\Phi_\XYH$ is defined in the last subsection.

Let $\phi$ be in $\Phi_\TbXYHb$ and $n$ be a natural number. Define $\phi^n$ as the result of replacing every subformula $\STN \psi$ of $\phi$ by $\YYY^n \SHN \XXX^n \psi$ and replacing every subformula $\WTN \psi$ of $\phi$ by $\YYY^n \WHN \XXX^n \psi$. Note $\phi^n$ is well-defined, as for every subformula $\STN \psi$ or $\WTN \psi$ of $\phi$, $\psi$ is in $\Phi_\XYH$. 

\begin{lemma}
\label{lemma:n root equivalence}

For every $\phi \in \Phi_\TbXYHb$ and $n$,
$\vdash \YYY'^n \Froot \rightarrow (\phi \leftrightarrow \phi^n)$.

\end{lemma}

\begin{proof}~

Let $\phi \in \Phi_\TbXYHb$ and $n$ be a natural number. We put an induction on the structure of $\phi$.

Assume $\phi$ is in $\Phi_\XYH$. Note in this case $\phi^n = \phi$. Then, the result holds.

Assume $\phi = \STN \psi$, where $\psi$ is in $\Phi_\XYH$. Note in this case $\phi^n = \YYY^n \SHN \XXX^n \psi$. By the axiom \ref{axiom:strong T replacement}, $\vdash \YYY'^n \Froot \rightarrow (\STN \psi \leftrightarrow \phi^n)$.

Assume $\phi = \WTN \psi$, where $\psi$ is in $\Phi_\XYH$. Note in this case $\phi^n = \YYY^n \WHN \XXX^n \psi$. By the axiom \ref{axiom:weak T replacement}, $\vdash \YYY'^n \Froot \rightarrow (\WTN \psi \leftrightarrow \phi^n)$.

Assume $\phi = \neg \psi$. Note in this case $\phi^n = \neg \psi^n$. By the inductive hypothesis, $\vdash \YYY'^n \Froot \rightarrow (\psi \leftrightarrow \psi^n)$. Then $\vdash \YYY'^n \Froot \rightarrow (\neg \psi \leftrightarrow \neg \psi^n)$, that is, $\vdash \YYY'^n \Froot \rightarrow (\phi \leftrightarrow \phi^n)$.

Assume $\phi = \psi \land \chi$. Note in this case $\phi^n = \psi^n \land \chi^n$. By the inductive hypothesis, $\vdash \YYY'^n \Froot \rightarrow (\psi \leftrightarrow \psi^n)$ and $\vdash \YYY'^n \Froot \rightarrow (\chi \leftrightarrow \chi^n)$. Then $\vdash \YYY'^n \Froot \rightarrow ((\psi \land \chi) \leftrightarrow (\psi^n \land \chi^n))$, that is, $\vdash \YYY'^n \Froot \rightarrow (\phi \leftrightarrow \phi^n)$.

\end{proof}

\begin{lemma}
\label{lemma:from T(X.Y.H.) to X.Y.H. derivability}

Let $\phi$ be in $\Phi_\TbXYHb$ and $n$ be its $\YYY$-depth. Then, $\vdash \phi \rightarrow \Big( (\Froot \rightarrow \phi^0) \land \dots \land (\YYY'^{n+1} \Froot \rightarrow \phi^{n+1}) \Big) $.

\end{lemma}

\begin{proof}
~

Let $k$ be such that $0 \leq k \leq n + 1$. By Lemma \ref{lemma:n root equivalence}, $\vdash \YYY'^k \Froot \rightarrow (\phi \rightarrow \phi^k)$. Then $\vdash \phi \rightarrow (\YYY'^k \Froot \rightarrow \phi^k)$. Note $k$ is arbitrary. Then $\vdash \phi \rightarrow \Big( (\Froot \rightarrow \phi^0) \land \dots \land (\YYY'^{n+1} \Froot \rightarrow \phi^{n+1}) \Big) $.

\end{proof}

\begin{lemma}[V-reducibility of $\Phi_{\TbXYHb}$ to $\Phi_\XYH$]
\label{lemma:v-reducibility of T(X.Y.H.) to X.Y.H.}

$\Phi_{\TbXYHb}$ is v-reducible to $\Phi_\XYH$.

\end{lemma}

\begin{proof}
~

Let $\phi$ be in $\Phi_\TbXYHb$ and $n$ be its $\YYY$-depth. Note $(\Froot \rightarrow \phi^0) \land \dots \land (\YYY'^{n+1} \Froot \rightarrow \phi^{n+1})$ is in $\Phi_\XYH$.
We claim:
\begin{enumerate}[label=(\arabic*),leftmargin=3.33em]

\item 

If $\models \phi$, then 
$\models (\Froot \rightarrow \phi^0) 
\land \dots \land
(\YYY'^{n+1} \Froot \rightarrow \phi^{n+1})$;

\item 

If 
$\vdash (\Froot \rightarrow \phi^0) 
\land \dots \land
(\YYY'^{n+1} \Froot \rightarrow \phi^{n+1})$,
then $\vdash \phi$.

\end{enumerate}

The statement (1) follows from Lemma \ref{lemma:from T(X.Y.H.) to X.Y.H. derivability} and the soundness of $\SWHTN$.

Consider the statement (2).
Assume $\vdash (\Froot \rightarrow \phi^0) \land \dots \land (\YYY'^{n+1} \Froot \rightarrow \phi^{n+1})$.
Then $\vdash \Froot \rightarrow \phi^0$. By Lemma \ref{lemma:n root equivalence}, $\vdash \Froot \rightarrow (\phi \leftrightarrow \phi^0)$. Then $\vdash \Froot \rightarrow \phi$.
Similarly, we can show $\vdash \YYY' \Froot \rightarrow \phi$, \dots, $\vdash \YYY'^{n+1} \Froot \rightarrow \phi$. 
Then $\vdash (\Froot \lor \YYY' \Froot \lor \dots \lor \YYY'^{n+1} \Froot) \rightarrow \phi$. 
By the rule of \emph{from-$\YYY$-depth-to-derivability}, we can get $\vdash \phi$.

\end{proof}

\paragraph{Remarks}

Note $\not \models \Big( (\Froot \rightarrow \phi^0) \land \dots \land (\YYY'^{n+1} \Froot \rightarrow \phi^{n+1}) \Big) \rightarrow \phi$. Here is a counterexample. Let $\phi = \bot$ and $(\MM,\CC,\pi,2)$ be a contextualized pointed model. Note the $\YYY$-depth of $\bot$ is $0$. It is easy to see $\MM,\CC,\pi,2 \Vdash (\Froot \rightarrow \bot) \land (\YYY' \bot \rightarrow \bot)$ and $\MM,\CC,\pi,2 \not \Vdash \bot$.

\subsection{V-reducibility of $\Phi_\XYH$ to $\Phi^{\mathtt{mo}}_\XYH$}

\begin{definition}[Language $\Phi^{\mathtt{mo}}_\XYH$]

\[
\chi ::= \bot \mid p \mid \neg \chi \mid (\chi \land \chi) \mid \YYY \chi \mid \SHN \phi \mid \WHN \phi
\]

\noindent where $\phi$ is in $\Phi_\XYH$.
\end{definition}

The language $\Phi^{\mathtt{mo}}_\XYH$ consists of moment formulas of $\Phi_\XYH$.

\begin{lemma}[V-reducibility of $\Phi_\XYH$ to $\Phi^{\mathtt{mo}}_\XYH$]
\label{lemma:v-reducibility of XYH to XYH mo}

$\Phi_\XYH$ is v-reducible to $\Phi^{\mathtt{mo}}_\XYH$.

\end{lemma}

\begin{proof}
~

Let $\phi \in \Phi_\XYH$. We claim (1) if $\models \phi$, then $\models \SHN \phi$, and (2) if $\vdash \SHN \phi$, then $\vdash \phi$.
The first statement is easy to check, and the second follows from the axiom \ref{axiom:SHN T}.

\end{proof}

\subsection{V-reducibility of $\Phi^{\mathtt{mo}}_\XYH$ to $\Phi_\YHX$}

\begin{definition}[Language $\Phi_{\YHX}$]
\label{def:The language SWONT}

\[\phi ::= \bot \mid p \mid \neg \phi \mid (\phi \land \phi) \mid \YYY \phi \mid \SHN \XXX \phi \mid \WHN \XXX \phi \]

\end{definition}

\newcommand{\Fdt}{\delta^{\mathtt{mo}}}

\newcommand{\bd}{bd}

\begin{definition}[OK formulas of $\Phi^{\mathtt{mo}}_\XYH$]

A formula $\nabla B$ in $\Phi^{\mathtt{mo}}_\XYH$ is \Fdefs{OK} if every subformula $\nabla' B'$ of $B$ is in $\Phi_\YHX$, where $\nabla, \nabla' \in \{\SHN, \WHN\}$.

\end{definition}

\begin{lemma}
\label{lemma:all OK modality is then is}

For every $\phi \in \Phi^{\mathtt{mo}}_\XYH$, if all of its OK subformulas $\nabla B$ are in $\Phi_\YHX$, then $\phi \in \Phi_\YHX$, where $\nabla \in \{\SHN, \WHN\}$.

\end{lemma}

\begin{proof}
~

Let $\phi \in \Phi^{\mathtt{mo}}_\XYH$.
We put an induction on the structure of $\phi$. We consider only the case $\phi = \nabla \psi$, where $\nabla \in \{\SHN, \WHN\}$. Other cases are easy.

Assume all OK subformulas of $\nabla \psi$ are in $\Phi_\YHX$.
Assume $\nabla \psi$ is OK. Clearly, $\nabla \psi$ is in $\Phi_\YHX$.
Assume $\nabla \psi$ is not OK. Then some subformula $\nabla_1 \psi_1$ of $\psi$ is not in $\Phi_\YHX$. Note that all OK subformulas of $\nabla_1 \psi_1$ are in $\Phi_\YHX$. By the inductive hypothesis, $\nabla_1 \psi_1$ is in $\Phi_\YHX$. We have a contradiction.

\end{proof}

\begin{definition}[Distance of formulas in $\Phi_\XYH$ to be a moment formula]

For every $\phi \in \Phi_\XYH$, define $\Fdt (\phi)$, the \Fdefs{distance} of $\phi$ to be a moment formula, as follows:
\begin{itemize}

\item 

$\Fdt (\bot) = \Fdt (p) = \Fdt (\SHN \psi) = \Fdt (\WHN \psi) = 0$

\item

$\Fdt (\neg \psi) = \Fdt (\YYY \psi) = \Fdt (\psi)$

\item 

$\Fdt (\psi \land \chi) = \max \{\Fdt(\psi), \Fdt(\chi)\}$

\item 

$\Fdt (\XXX \psi) = \Fdt (\psi) + 1$

\end{itemize}

\end{definition}

\begin{lemma}
\label{lemma:distance normal form}

Let $\nabla B$ be an OK formula in $\Phi^{\mathtt{mo}}_\XYH$, where $\nabla \in \{\SHN, \WHN\}$. 

There is $B'$ in $\Phi_\XYH$ such that (1) $\vdash B \leftrightarrow B'$, (2) $\nabla B'$ is OK, (3) $\Fdt (B) = \Fdt (B')$, and (4) $B'$ is in the form $C_1 \land \dots \land C_n$, where every $C_i$ is in the form 
\[
\FBV \{\XXX^{h_a} D_a \mid a \in I_1\}
\lor
\FBV \{\YYY^{i_b} E_b \mid b \in I_2\}
\lor
\FBV \{\YYY'^{j_c} F_c \mid c \in I_3\}
\lor
\FBV \{G_d \mid d \in I_4 \},
\]
where (1) $I_1, I_2, I_3, I_4$ are (possibly empty) finite sets of indices, (2) every $h_a, i_b, j_c$ is not $0$, and (3) every $D_x, E_x, F_x, G_x$ is either a propositional literal, or in the form $\SHN H, \SHP H, \WHN H$ or $\WHP H$.

\end{lemma}

\begin{proof}
~

We describe a procedure for obtaining a formula $B'$ that meets the four conditions mentioned above.
\begin{itemize}

\item 

Make $\neg$ in $B$ go further and further till it is next to $\bot$ or $p$, and let $B_1$ be the result. Note: $\vdash B_1 \leftrightarrow B$ and $\Fdt (B_1) = \Fdt (B)$. Also note every subformula $\nabla' H$ of $B_1$ is in $\Phi_\YHX$, as $\nabla B$ is OK.

\item 

Make $\lor$ be distributed into $\land$, and let $B'$ be the result. Note every subformula $\nabla H$ of $B'$ is in $\Phi_\YHX$. It is clear that the other three conditions mentioned above are met for $B'$.

\end{itemize}

\end{proof}

\begin{lemma}
\label{lemma:Box moment formula}
~

\begin{enumerate}[label=(\arabic*),leftmargin=3.33em]

\item 

$\vdash \SHN \FMF \leftrightarrow \FMF$

\item

$\vdash \WHN \FMF \leftrightarrow (\WHN \bot \lor \FMF)$

\item 

$\vdash \SHN (\phi \lor \FMF) \leftrightarrow (\SHN \phi \lor \FMF)$

\item 

$\vdash \WHN (\phi \lor \FMF) \leftrightarrow (\WHN \phi \lor \FMF)$

\end{enumerate}

\end{lemma}

\begin{proof}
~

\begin{enumerate}[label=(\arabic*),leftmargin=3.33em]

\item By the rule of replacement of equivalence, $\vdash \SHN \FMF \leftrightarrow \SHN (\neg \FMF \rightarrow \bot)$.

By the axiom \ref{axiom:moment formula}, $\vdash \SHN (\neg \FMF \rightarrow \bot) \rightarrow (\neg \FMF \rightarrow \SHN \bot)$.
By the axiom \ref{axiom:SHN T}, $\vdash \neg \SHN \bot$. Then $\vdash \SHN (\neg \FMF \rightarrow \bot) \rightarrow \FMF$. Then $\vdash \SHN \FMF \rightarrow \FMF$.

Note $\vdash \FMF \rightarrow (\neg \FMF \rightarrow \bot)$. By the rule of necessitation of $\SHN$, $\vdash \SHN \big( \FMF \rightarrow (\neg \FMF \rightarrow \bot) \big)$. By the axiom \ref{axiom:moment formula}, $\vdash \FMF \rightarrow \SHN (\neg \FMF \rightarrow \bot) \big)$. Then, $\vdash \FMF \rightarrow \SHN \FMF$.

\item By the rule of replacement of equivalence, $\vdash \WHN \FMF \leftrightarrow \WHN (\neg \FMF \rightarrow \bot)$.

By the axiom \ref{axiom:moment formula WHN}, $\vdash \WHN (\neg \FMF \rightarrow \bot) \rightarrow (\neg \FMF \rightarrow \WHN \bot)$. Then, $\vdash \WHN (\neg \FMF \rightarrow \bot) \rightarrow (\FMF \lor \WHN \bot)$. Then, $\vdash \WHN \FMF \rightarrow (\FMF \lor \WHN \bot)$.

Note $\vdash \FMF \rightarrow (\neg \FMF \rightarrow \bot)$. By the rule of necessitation of $\WHN$, $\vdash \WHN \big( \FMF \rightarrow (\neg \FMF \rightarrow \bot) \big)$. By the axiom \ref{axiom:moment formula WHN}, $\vdash \FMF \rightarrow \WHN (\neg \FMF \rightarrow \bot) \big)$. Then, $\vdash \FMF \rightarrow \WHN \FMF$.
It is easy to get $\vdash \WHN \bot \rightarrow \WHN \FMF$. Then $\vdash (\WHN \bot \lor \FMF) \rightarrow \WHN \FMF$.

\item By the axiom \ref{axiom:moment formula}, $\vdash \SHN (\neg \FMF \rightarrow \phi) \rightarrow (\neg \FMF \rightarrow \SHN \phi)$.
Then $\vdash \SHN (\phi \lor \FMF) \rightarrow (\FMF \lor \SHN \phi)$.

It is easy to show $\vdash \SHN \phi \rightarrow \SHN (\phi \lor \FMF)$. By the first item, it is easy to get $\vdash \FMF \rightarrow \SHN (\phi \lor \FMF)$. Then, $\vdash (\SHN \phi \lor \FMF) \rightarrow \SHN (\phi \lor \FMF)$.

\item Similar to the third item.

\end{enumerate}

\end{proof}

\begin{lemma}
\label{lemma:OK modality crucial}

Let $\Box B \in \Phi^{\mathtt{mo}}_\XYH$ be an OK formula not in $\Phi_\YHX$, where $\Box \in \{\SHN, \WHN\}$. Then, there is $\psi \in \Phi_\XYH^{\mathtt{mo}}$ such that (1) $\vdash \Box B \leftrightarrow \psi$, and (2) for all OK subformula $\Box' \theta$ of $\psi$, if $\Box' \theta \notin \Phi_\YHX$, then $\Fdt (\theta) < \Fdt (B)$.

\end{lemma}

\begin{proof}
~

By Lemma \ref{lemma:distance normal form}, there is $B'$ in $\Phi_\XYH$ such that (1) $\vdash B \leftrightarrow B'$, (2) $\Box B'$ is OK, (3) $\Fdt (B) = \Fdt (B')$, and (4) $B'$ is in the form $C_1 \land \dots \land C_n$, where every $C_i$ is in the form 
$
\FBV \{\XXX^{h_a} D_a \mid a \in I_1\}
\lor
\FBV \{\YYY^{i_b} E_b \mid b \in I_2\}
\lor
\FBV \{\YYY'^{j_c} F_c \mid c \in I_3\}
\lor
\FBV \{G_d \mid d \in I_4 \}$, where (1) $I_1, I_2, I_3, I_4$ are (possibly empty) finite sets of indices, (2) every $h_a, i_b, j_c$ is not $0$, and (3) every $D_x, E_x, F_x, G_x$ is either a propositional literal, or in the form $\SHN H, \SHP H, \WHN H$ or $\WHP H$.

Note $\vdash \Box B \leftrightarrow \Box B'$ and $\vdash \Box B' \leftrightarrow (\Box C1 \land \dots \land \Box C_n)$. Note $\Box B'$ is OK. Then all subformulas $\Box' H$ of $B'$ are in $\Phi_\YHX$. Then all subformulas $\Box' H$ of every $C_i$ are in $\Phi_\YHX$. Then every $\Box C_i$ is OK.

Assume $\Box C_1 \land \dots \land \Box C_n$ is in $\Phi_\YHX$. Then there is no subformula $\Box' \theta$ of $\Box C_1 \land \dots \land \Box C_n$ such that $\Box' \theta \notin \Phi_\YHX$. Then we are done.

\medskip

Assume $\Box C_1 \land \dots \land \Box C_n$ is not in $\Phi_\YHX$.

Pick a $\Box C_i$ not in $\Phi_\YHX$. Let $C_i =
\FBV \{\XXX^{h_a} D_a \mid a \in I_1\}
\lor
\FBV \{\YYY^{i_b} E_b \mid b \in I_2\}
\lor
\FBV \{\YYY'^{j_c} F_c \mid c \in I_3\}
\lor
\FBV \{G_d \mid d \in I_4 \}$. 

\textbf{We want to transform $\Box C_i$ to a formula $\chi \in \Phi_\XYH^{\mathtt{mo}}$ such that $\vdash \Box C_i \leftrightarrow \chi$ and for all OK subformula $\Box' \theta$ of $\chi$, if $\Box' \theta \notin \Phi_\YHX$, then $\Fdt (\theta) < \Fdt (C_i)$.}

Assume $I_1 = \emptyset$. It is easy to see $C_i$ is in $\Phi_\YHX$. Assume $\Box = \SHN$. By the first item of Lemma \ref{lemma:Box moment formula}, we can replace $\Box C_i$ by $C_i$. Assume $\Box = \WHN$. By the second item of Lemma \ref{lemma:Box moment formula} and $\vdash \bot \leftrightarrow \XXX \bot$, which is easy to show, we can replace $\Box C_i$ by $\Box \XXX \bot \lor C_i$. Note $\Box \XXX \bot \lor C_i$ is in $\Phi_\YHX$. Then, we are done.

Assume $I_1 \neq \emptyset$.

By the third and fourth items of Lemma \ref{lemma:Box moment formula}, $\vdash \Box C_i \leftrightarrow
\Big(
\Box \FBV \{\XXX^{h_a} D_a \mid a \in I_1\}
\lor
\FBV \{\YYY^{i_b} E_b \mid b \in I_2\}
\lor
\FBV \{\YYY'^{j_c} F_c \mid c \in I_3\}
\lor
\FBV \{G_d \mid d \in I_4 \}
\Big)$.

Assume $\Box \FBV \{\XXX^{h_a} D_a \mid a \in I_1\}$ is in $\Phi_\YHX$. Then we are done. 

Assume $\Box \FBV \{\XXX^{h_a} D_a \mid a \in I_1\}$ is not in $\Phi_\YHX$. Note $\Fdt (\FBV \{\XXX^{h_a} D_a \mid a \in I_1\}) = \Fdt (C_i)$.

Note $\vdash \FBV \{\XXX^{h_a} D_a \mid a \in I_1\} \leftrightarrow \XXX \FBV \{\XXX^{h_a - 1} D_a \mid a \in I_1\}$, by Lemma \ref{lemma:make T out}.
Then $\vdash \Box C_i \leftrightarrow
\Big(
\Box \XXX \FBV \{\XXX^{h_a-1} D_a \mid a \in I_1\}
\lor
\FBV \{\YYY^{i_b} E_b \mid b \in I_2\}
\lor
\FBV \{\YYY'^{j_c} F_c \mid c \in I_3\}
\lor
\FBV \{G_d \mid d \in I_4 \}
\Big)$.

Assume $\Box \XXX \FBV \{\XXX^{h_a-1} D_a \mid a \in I_1\}$ is in $\Phi_\YHX$. Then we are done.

Assume $\Box \XXX \FBV \{\XXX^{h_a-1} D_a \mid a \in I_1\}$ is not in $\Phi_\YHX$.

Note $\vdash \Box \XXX \FBV \{\XXX^{h_a-1} D_a \mid a \in I_1\} \leftrightarrow \Box \XXX \Box \FBV \{\XXX^{h_a-1} D_a \mid a \in I_1\}$ by the axioms \ref{axiom:inserting SHN} and \ref{axiom:inserting WHN}.
Then $\vdash \Box C_i \leftrightarrow
\Big(
\Box \XXX \Box \FBV \{\XXX^{h_a-1} D_a \mid a \in I_1\}
\lor
\FBV \{\YYY^{i_b} E_b \mid b \in I_2\}
\lor
\FBV \{\YYY'^{j_c} F_c \mid c \in I_3\}
\lor
\FBV \{G_d \mid d \in I_4 \}
\Big)$.

Assume $\Box \FBV \{\XXX^{h_a-1} D_a \mid a \in I_1\}$ is in $\Phi_\YHX$. Then we are done.

Assume $\Box \FBV \{\XXX^{h_a-1} D_a \mid a \in I_1\}$ is not in $\Phi_\YHX$.

Note $\Box C_i$ is OK. It can be seen that (1) $\Box \FBV \{\XXX^{h_a-1} D_a \mid a \in I_1\}$ is OK, and (2) it is the only OK subformula of $\Box \XXX \Box \FBV \{\XXX^{h_a-1} D_a \mid a \in I_1\}
\lor
\FBV \{\YYY^{i_b} E_b \mid b \in I_2\}
\lor
\FBV \{\YYY'^{j_c} F_c \mid c \in I_3\}
\lor
\FBV \{G_d \mid d \in I_4 \}$ such that it is not in $\Phi_\YHX$.
Note $\Fdt (\FBV \{\XXX^{h_a-1} D_a \mid a \in I_1\}) < \Fdt (C_i)$. Then we are done.

\end{proof}

\begin{lemma}
\label{lemma:moment formula from XYH to YHX}

For every formula $\phi$ in $\Phi^{\mathtt{mo}}_{\XYH}$, there is $\phi'$ in $\Phi_{\YHX}$ such that $\vdash \phi \leftrightarrow \phi'$.

\end{lemma}

\begin{proof}
~

Let $A$ be in $\Phi^{\mathtt{mo}}_\XYH$. In what follows, we describe a procedure to transform $A$ to $\Phi_{\YHX}$.

\begin{enumerate}

\item 

Assume $A$ is in $\Phi_\YHX$. Then we stop.

\item

Assume $A$ is not in $\Phi_\YHX$.

Define $\Delta_A = \{\Box B \mid \Box B \text{ occurs in } A, \text{ is OK, but not in } \Phi_\YHX \}$, where $\Box \in \{\SHN, \WHN\}$.
By Lemma \ref{lemma:all OK modality is then is}, $\Delta_A$ is not empty.

Let $n$ be the greatest natural number in the set
$\{\Fdt (B) \mid \Box B \in \Delta_A \}$.
Let $\Delta^n_A = \{\Box B \in \Delta_A \mid \Fdt (B) = n\}$.
Pick a $\Box B$ in $\Delta^n_A$.
By Lemma \ref{lemma:OK modality crucial}, there is $\psi \in \Phi^{\mathtt{mo}}_\XYH$ such that $\vdash \Box B \leftrightarrow \psi$ and for all OK formula $\Box' \theta$ in $\psi$, if $\Box' \theta \notin \Phi_\YHX$, then $\Fdt (\theta) < \Fdt (B)$.
Replace $\Box B$ by $\psi$ in $A$.

\item Repeat.

\end{enumerate}

Assume we will not stop. Note that after a finite number of steps, $n$ will get smaller. Then we will have an infinite descending chain of natural numbers. This is impossible.

\end{proof}

From this lemma and the soundness of $\SWHTN$, the following result follows:

\begin{lemma}[V-reducibility of $\Phi^{\mathtt{mo}}_\XYH$ to $\Phi_\YHX$]
\label{lemma:v-reducibility of XYH mo to YHX}

$\Phi^{\mathtt{mo}}_\XYH$ is v-reducible to $\Phi_\YHX$.

\end{lemma}

\subsection{V-reducibility of $\Phi_\YHX$ to $\Phi_{\YmHX}$}

\begin{definition}[Language $\Phi_{\YmHX}$]

\[\psi ::= \phi \mid \neg \psi \mid (\psi \land \psi) \mid \YYY \psi \]

\noindent where $\phi$ is in $\Phi_\HX$.

\end{definition}

Note: in $\Phi_{\YmHX}$, $\YYY$ does not occur in the scope of $\SHN \XXX$ or $\WHN \XXX$.

\begin{lemma}
\label{lemma:direct container}

For every $\phi$ of $\Phi_{\YHX}$, if $\phi \notin \Phi_\YmHX$, then some subformula $\YYY B$ of $\phi$ has a direct container in the form $\Box \XXX C$, where $\Box \in \{\SHN, \WHN\}$.

\end{lemma}

\begin{proof}
~

Let $\phi \in \Phi_{\YHX}$ such that $\phi \notin \Phi_\YmHX$.

First, we show that there is $\YYY B$ and $\Box \XXX C$ in $\phi$ such that $\YYY B$ is a subformula of $\Box \XXX C$.
Assume for all $\YYY B$ and $\Box \XXX C$ in $\phi$, $\YYY B$ is not a subformula of $\Box \XXX C$. We want to show $\phi$ is in $\Phi_\YmHX$, which causes a contradiction. We put an induction on the structure of $\phi$.
\begin{itemize}

\item 

Suppose $\phi$ equals to $\bot$ or $p$. Clearly, $\phi$ is in $\Phi_\YmHX$.

\item

Suppose $\phi = \neg \psi$. Then for all $\YYY B$ and $\Box \XXX C$ in $\psi$, $\YYY B$ is not a subformula of $\Box \XXX C$. By the inductive hypothesis, $\psi$ is in $\Phi_\YmHX$. Then $\neg \psi$ is in $\Phi_\YmHX$.

\item 

Suppose $\phi = \psi \land \chi$. Then for all $\YYY B$ and $\Box \XXX C$ in $\psi$, $\YYY B$ is not a subformula of $\Box \XXX C$, and for all $\YYY B$ and $\Box \XXX C$ in $\chi$, $\YYY B$ is not a subformula of $\Box \XXX C$. By the inductive hypothesis, $\psi$ is in $\Phi_\YmHX$ and $\chi$ is in $\Phi_\YmHX$. Then $\psi \land \chi$ is in $\Phi_\YmHX$.

\item

Suppose $\phi = \YYY \psi$. Then for all $\YYY B$ and $\Box \XXX C$ in $\psi$, $\YYY B$ is not a subformula of $\Box \XXX C$. By the inductive hypothesis, $\psi$ is in $\Phi_\YmHX$. Then $\YYY \psi$ is in $\Phi_\YmHX$.

\item

Suppose $\phi = \SHN \XXX \psi$. Note for all $\YYY B$ and $\Box \XXX C$ in $\SHN \XXX \psi$, $\YYY B$ is not a subformula of $\Box \XXX C$. This implies that $\psi$ contains no $\YYY$. Then $\SHN \XXX \psi \in \Phi_\YmHX$.

\item 

Suppose $\phi = \WHN \XXX \psi$. Similar to the previous case.

\end{itemize}

Thus, there is a sequence $\nabla_0 \psi_0, \dots, \nabla_n \psi_n$ of subformulas of $\phi$, where $\nabla_i \in \{\YYY, \SHN \XXX, \WHN \XXX\}$, such that (1) $\nabla_0 \psi_0 = \YYY B$, (2) $\nabla_n \psi_n = \Box \XXX C$, and (3) for all $i < n$, $\nabla_{i+1} \psi_{i + 1}$ is a direct container of $\nabla_i \psi_i$.

Let $i < n$ be the greatest natural number such that $\nabla_i = \YYY$ and $\nabla_{i+1} = \Box' \XXX$, where $\Box' \in \{\SHN, \WHN\}$. Then $\Box' \XXX \psi_{i+1}$ is a direct container of $\YYY \psi_i$.

\end{proof}

\begin{lemma}
\label{lemma:X Y}
~

\begin{enumerate}[label=(\arabic*),leftmargin=3.33em]

\item 

$\vdash \XXX \YYY \phi \leftrightarrow \phi$

\item 

$\vdash \XXX \YYY' \phi \leftrightarrow \phi$

\end{enumerate}

\end{lemma}

\begin{proof}
~

(1) By the axiom \ref{axiom: phi XYp phi}, $\vdash \neg \phi \rightarrow \XXX \neg \YYY \neg \neg \phi$. Then $\vdash \neg \phi \rightarrow \XXX \neg \YYY \phi$. Then $\vdash \neg \XXX \neg \YYY \phi \rightarrow \phi$. By the axiom \ref{axiom: X phi not X not phi}, $\vdash \XXX \YYY \phi \rightarrow \phi$.
By the axiom \ref{axiom: phi XYp phi}, $\vdash \phi \rightarrow \XXX \YYY' \phi$. Note by the axiom \ref{axiom:Y partial function}, $\vdash \YYY' \phi \rightarrow \YYY \phi$. It is easy to show $\vdash \XXX \YYY' \phi \rightarrow \XXX \YYY \phi$. Then $\vdash \phi \rightarrow \XXX \YYY \phi$.

(2) By the previous item, $\vdash \XXX \YYY \neg \phi \leftrightarrow \neg \phi$. Then $\vdash \neg \XXX \YYY \neg \phi \leftrightarrow \phi$. Then $\vdash \neg \XXX \neg \YYY' \phi \leftrightarrow \phi$. By the item (2a) in Lemma \ref{lemma:make T out}, $\vdash \XXX \YYY' \phi \leftrightarrow \phi$. 

\end{proof}

\begin{lemma}
\label{lemma:from YHX to YmHX}

For every $\phi$ of $\Phi_{\YHX}$, there is a $\psi$ in $\Phi_{\YmHX}$ such that $\vdash \phi \leftrightarrow \psi$.

\end{lemma}

\begin{proof}~

Let $A$ be a formula in $\Phi_{\YHX}$. If $A$ is in $\Phi_{\YmHX}$, then we are done. Assume $A$ is not in $\Phi_\YmHX$. In what follows, we describe a procedure to transform $A$ to $\Phi_{\YmHX}$.

By Lemma \ref{lemma:direct container}, there are $\YYY B$ and $\Box \XXX C$ in $A$ such that $\Box \XXX C$ is a direct container of $\YYY B$, where $\Box \in \{\SHN, \WHN\}$.

Assume $\Box = \SHN$.

We transform $C$ to $C_1 \land \dots \land C_n$ in conjunctive normal form.
Note $\vdash \SHN \XXX C \leftrightarrow \Big( \SHN \XXX C_1 \land \dots \land \SHN \XXX C_n \Big)$ by Lemma \ref{lemma:make T out}. Replace $\SHN \XXX C$ by $\SHN \XXX C_1 \land \dots \land \SHN \XXX C_n$.

Pick a $C_i$ containing $\YYY B$. Let $C_i = D_1 \lor \dots \lor D_m$. Then some $D_j$ equals to $\YYY B$ or $\neg \YYY B$. Without loss of any generality, we can assume (1) $D_1$ equals $\YYY B$ or $\neg \YYY B$, and (2) $m \geq 2$.

By Lemma \ref{lemma:make T out}, $\SHN \XXX C_i$, that is, $\SHN \XXX ( D_1 \lor \dots \lor D_m)$ is provably equivalent to $\SHN \Big( \XXX ( D_2 \lor \dots \lor D_m ) \lor \XXX D_1 \Big)$.

Assume $D_1 = \YYY B$. Note $\vdash \XXX \YYY B \leftrightarrow B$ by Lemma \ref{lemma:X Y}.
Then, $\SHN \Big( \XXX ( D_2 \lor \dots \lor D_m ) \lor \XXX D_1 \Big)$ is provably equivalent to $\SHN \Big( \XXX ( D_2 \lor \dots \lor D_m ) \lor B \Big)$. Note we remove a $\YYY$. Replace $\SHN \XXX C_i$ by $\SHN \Big( \XXX ( D_2 \lor \dots \lor D_m ) \lor B \Big)$. 

Assume $D_1 = \neg \YYY B$. Note $\vdash \XXX \neg \YYY B \leftrightarrow \neg B$ by Lemma \ref{lemma:X Y}.
Then, $\SHN \Big( \XXX ( D_2 \lor \dots \lor D_m ) \lor \XXX D_1 \Big)$ is provably equivalent to $\SHN \Big( \XXX ( D_2 \lor \dots \lor D_m ) \lor \neg B \Big)$. Note, we remove a $\YYY$. Replace $\SHN \XXX C_i$ by $\SHN \Big( \XXX ( D_2 \lor \dots \lor D_m ) \lor \neg B \Big)$. 

Assume $\Box = \WHN$. Do similar things as above.

Repeat until there are no $\YYY B$ and $\Box \XXX C$ in $A$ such that $\Box \XXX C$ is a direct container of $\YYY B$, where $\Box \in \{\SHN, \WHN\}$.
By Lemma \ref{lemma:direct container}, the resulting formula is in $\Phi_\YmHX$.

\end{proof}

From this lemma and the soundness of $\SWHTN$, the following result follows:

\begin{lemma}[V-reducibility of $\Phi_\YHX$ to $\Phi_\YmHX$]
\label{lemma:v-reducibility of YHX to YmHX}

$\Phi_\YHX$ is v-reducible to $\Phi_\YmHX$.

\end{lemma}

\subsection{V-reducibility of $\Phi_{\YmHX}$ to $\Phi_{\YnbHXb}$}

\begin{definition}[Language $\Phi_{\YnbHXb}$]
\label{def:The language SWONT}

\[\psi ::= \YYY^n \phi \mid \neg \psi \mid (\psi \land \psi) \]

\noindent where $\phi$ is in $\Phi_\HX$.

\end{definition}

\begin{lemma}
\label{lemma:from YmHX to Yn(HX)}

For every $\phi$ in $\Phi_{\YmHX}$, there is $\psi$ in $\Phi_{\YnbHXb}$ such that $\vdash \phi \leftrightarrow \psi$.

\end{lemma}

\begin{proof}
~

Let $A$ be in $\Phi_{\YmHX}$. We put an induction on the structure of $A$. We consider only the case $A = \YYY \psi$; other cases are easy.

Suppose $A = \YYY \psi$. By the inductive hypothesis, there is $\psi'$ in $\Phi_\YnbHXb$ such that $\vdash \psi \leftrightarrow \psi'$. Then $\YYY \psi$ is provably equivalent to $\YYY \psi'$.

We transform $\psi'$ to $C_1 \land \dots \land C_n$ in conjunctive normal form, where every $C_i$ is in the form $D_1 \lor \dots \lor D_m$, where every $D_j$ is in the form $p$, $\neg p$, $\bot$, $\top$, $\YYY E$, $\YYY' E$, $\SHN \XXX E$, $\SHP \XXX E$, $\WHN \XXX E$, or $\WHP \XXX E$.

By Lemma \ref{lemma:make T out}, $\YYY \psi'$ is provably equivalent to $\YYY C_1 \land \dots \land \YYY C_n$. Let $1 \leq i \leq n$. It suffices to show $\YYY C_i$ is provably equivalent to some $\chi$ in $\Phi_{\YnbHXb}$.

Let $C_i = D_1 \lor \dots \lor D_m$. By Lemma \ref{lemma:make T out}, $\YYY C_i$ is provably equivalent to $\YYY D_1 \lor \dots \lor \YYY D_m$. It is easy to show that every $\YYY D_j$ is provably equivalent to some formula in $\Phi_{\YnbHXb}$. Then $\YYY D_1 \lor \dots \lor \YYY D_m$ is provably equivalent to some $\chi$ in $\Phi_{\YnbHXb}$. Then $\YYY C_i$ is provably equivalent to $\chi$.

\end{proof}

From this lemma and the soundness of $\SWHTN$, the following result follows:

\begin{lemma}[V-reducibility of $\Phi_\YmHX$ to $\Phi_{\YnbHXb}$]
\label{lemma:v-reducibility of YmHX to YnbHXb}

$\Phi_\YmHX$ is v-reducible to $\Phi_{\YnbHXb}$.

\end{lemma}

\subsection{V-reducibility of $\Phi_{\YnbHXb}$ to $\Phi_\HX$}

\begin{lemma}
\label{lemma:facts for reduction of Y}
~

\begin{enumerate}[label=(\arabic*),leftmargin=3.33em]

\item 

$\vdash \YYY \XXX \phi \leftrightarrow (\YYY \bot \lor \phi)$

\item For every $\phi \in \Phi_\HX$:
\begin{enumerate}

\item

$\vdash \phi \lor \YYY' \chi$ if and only if $\vdash \phi$

\item

$\models \phi \lor \YYY' \chi$ if and only if $\models \phi$

\end{enumerate}

\item For every $\phi \in \Phi_\HX$:
\begin{enumerate}

\item

$\vdash \phi \lor \YYY \psi$ if and only if $\vdash \XXX \phi \lor \psi$

\item

$\models \phi \lor \YYY \psi$ if and only if $\models \XXX \phi \lor \psi$

\end{enumerate}

\item 

$\vdash \SHP (\FMF \land \XXX \phi) \leftrightarrow (\FMF \land \SHP \XXX \phi)$

\end{enumerate}

\end{lemma}

\begin{proof}
~

\begin{enumerate}[label=(\arabic*),leftmargin=3.33em]

\item

By the axiom \ref{axiom: phi YXp phi}, $\vdash \neg \phi \rightarrow \YYY \neg \XXX \neg \neg \phi$. Then, $\vdash \neg \phi \rightarrow \YYY \neg \XXX \phi$. Then, $\vdash \YYY' \XXX \phi \rightarrow \phi$. As in normal modal logic, we can show $\vdash (\YYY \XXX \phi \land \YYY' \top) \rightarrow \YYY' (\XXX \phi \land \top)$. Then, $\vdash (\YYY \XXX \phi \land \YYY' \top) \rightarrow \YYY' \XXX \phi$.
Then, $\vdash (\YYY \XXX \phi \land \YYY' \top) \rightarrow \phi$. Then, $\vdash \YYY \XXX \phi \rightarrow (\YYY' \top \rightarrow \phi)$. Then, $\vdash \YYY \XXX \phi \rightarrow (\YYY \bot \lor \phi)$.

It is easy to show $\vdash \YYY \bot \rightarrow \YYY \XXX \phi$.
By the axiom \ref{axiom: phi YXp phi}, $\vdash \phi \rightarrow \YYY \neg \XXX \neg \phi$. By the item \ref{lemma item: not X not} in Lemma \ref{lemma:make T out}, $\vdash \phi \rightarrow \YYY \XXX \phi$. Thus, $\vdash (\YYY \bot \lor \phi) \rightarrow \YYY \XXX \phi$.

\item 

Let $\phi \in \Phi_\HX$.
\begin{enumerate}

\item 

The direction from the right to the left is easy. Consider the other direction. Assume $\vdash \phi \lor \YYY' \chi$. Then $\vdash \YYY \neg \chi \rightarrow \phi$. By the rule of from-root-to-derivability, $\vdash \phi$.

\item 

The direction from the right to the left is easy. Consider the other direction. Assume $\models \phi \lor \YYY' \chi$. Then $\models \YYY \neg \chi \rightarrow \phi$. By the rule of from-root-to-derivability preserving validity, $\models \phi$.

\end{enumerate}

\item 

Let $\phi \in \Phi_\HX$.
\begin{enumerate}

\item 

Assume $\vdash \phi \lor \YYY \psi$. By the rule of necessitation of $\XXX$, $\vdash \XXX (\phi \lor \YYY \psi)$. By Lemma \ref{lemma:make T out}, $\vdash \XXX \phi \lor \XXX \YYY \psi$. By Lemma \ref{lemma:X Y}, $\vdash \XXX \phi \lor \psi$.

Assume $\vdash \XXX \phi \lor \psi$. By the rule of necessitation of $\YYY$, $\vdash \YYY (\XXX \phi \lor \psi)$. By Lemma \ref{lemma:make T out}, $\vdash \YYY \XXX \phi \lor \YYY \psi$. By the first item of this lemma, $\vdash \YYY \bot \lor \phi \lor \YYY \psi$. Note $\vdash \YYY \bot \rightarrow \YYY \psi$. Then $\vdash \phi \lor \YYY \psi$.

\item

Assume $\models \phi \lor \YYY \psi$. By the rule of necessitation of $\XXX$ preserving validity, $\models \XXX (\phi \lor \YYY \psi)$. Then $\models \XXX \phi \lor \XXX \YYY \psi$. Then $\models \XXX \phi \lor \psi$.

Assume $\models \XXX \phi \lor \psi$. By the rule of necessitation of $\YYY$ preserving validity, $\models \YYY (\XXX \phi \lor \psi)$. Then $\models \YYY \XXX \phi \lor \YYY \psi$. By the first item of this lemma and the soundness of $\SWHTN$, $\models \YYY \bot \lor \phi \lor \YYY \psi$. Note $\models \YYY \bot \rightarrow \YYY \psi$. Then $\models \phi \lor \YYY \psi$.

\end{enumerate}

\item This item can be easily shown by the third item of Lemma \ref{lemma:Box moment formula}.

\end{enumerate}

\end{proof}

\begin{definition}[Standard disjunctions of $\Phi_\YnbHXb$]

A formula $\phi$ of $\Phi_\YnbHXb$ is called a \Fdefs{standard disjunction} of $\Phi_\YnbHXb$ if $\phi$ is in the form
\[
\FBV_{j \in \FPI_1} \SHP \XXX E_j \lor
\FBV_{j \in \FPI_2} \SHN \XXX F_j \lor
\FBV_{j \in \FPI_3} \WHP \XXX G_j \lor
\FBV_{j \in \FPI_4} \WHN \XXX H_j \lor 
\FBV_{j \in \FPI_5} l_j \lor
\FBV_{j \in \FPI_6} \YYY'^{m_j} I_j \lor 
\FBV_{j \in \FPI_7} \YYY^{n_j} J_j,
\]
\noindent where every $\FPI_i$ is a index set, $\FPI_6 \neq \emptyset$, every $l_j$ is a propositional literal, and all $m^j$ and $n^j$ are not $0$.

\end{definition}

\begin{lemma}[Normal form lemma for $\Phi_\YnbHXb$]
\label{lemma:Normal form of formulas in Yn(HX)}

For every $\psi \in \Phi_\YnbHXb$, there is $\psi' \in \Phi_\YnbHXb$ such that (1) $\vdash \psi \leftrightarrow \psi'$, and (2) $\psi'$ is in the form $\psi_1 \land \dots \land \psi_n$, where every $\psi_i$ is a standard disjunction of $\Phi_\YnbHXb$ with the $\YYY$-depth not greater than the $\YYY$-depth of $\psi$.

\end{lemma}

Note $\vdash \phi \leftrightarrow (\phi \lor \YYY' \bot)$. Then, it is easy to show this result, and we skip its proof.

\begin{lemma}[Reduction of $\YYY$ in standard disjunctions of $\Phi_\YnbHXb$ at the level of derivability]
\label{lemma:Reduction of YYY at the level of derivability}
~

\begin{enumerate}[label=(\arabic*),leftmargin=3.33em]

\item 

$\vdash
\FBV_{j \in \FPI_1} \SHP \XXX E_j \lor
\FBV_{j \in \FPI_2} \SHN \XXX F_j \lor
\FBV_{j \in \FPI_3} \WHP \XXX G_j \lor
\FBV_{j \in \FPI_4} \WHN \XXX H_j \lor 
\FBV_{j \in \FPI_5} l_j \lor
\FBV_{j \in \FPI_6} \YYY'^{m_j} I_j \lor
\FBV_{j \in \FPI_7} \YYY^{n_j - 1} J_j
$

$\Leftrightarrow$

$\vdash 
\FBV_{j \in \FPI_1} \SHP \XXX E_j \lor
\FBV_{j \in \FPI_2} \SHN \XXX F_j \lor
\FBV_{j \in \FPI_3} \WHP \XXX G_j \lor
\FBV_{j \in \FPI_4} \WHN \XXX H_j \lor 
\FBV_{j \in \FPI_5} l_j
$,

where $\FPI_7$ is empty.

\item

$
\vdash
\FBV_{j \in \FPI_1} \SHP \XXX E_j \lor
\FBV_{j \in \FPI_2} \SHN \XXX F_j \lor
\FBV_{j \in \FPI_3} \WHP \XXX G_j \lor
\FBV_{j \in \FPI_4} \WHN \XXX H_j \lor 
\FBV_{j \in \FPI_5} l_j \lor
\FBV_{j \in \FPI_6} \YYY'^{m_j} I_j \lor 
\FBV_{j \in \FPI_7} \YYY^{n_j} J_j
$

$\Leftrightarrow$

$\vdash
\SHN \XXX \bigg(
\FBV_{j \in \FPI_1} \SHP \XXX E_j \lor
\FBV_{j \in \FPI_2} \SHN \XXX F_j \lor
\FBV_{j \in \FPI_3} \WHP \XXX G_j \lor
\FBV_{j \in \FPI_4} \WHN \XXX H_j \lor 
\FBV_{j \in \FPI_5} l_j
\bigg)
\lor
\FBV_{j \in \FPI_6} \YYY'^{m_j - 1} I_j \lor 
\FBV_{j \in \FPI_7} \YYY^{n_j - 1} J_j$,

where $\FPI_7$ is not empty.

\end{enumerate}

\end{lemma}

\begin{proof}
~

\begin{enumerate}[label=(\arabic*),leftmargin=3.33em]

\item 

$\vdash
\FBV_{j \in \FPI_1} \SHP \XXX E_j \lor
\FBV_{j \in \FPI_2} \SHN \XXX F_j \lor
\FBV_{j \in \FPI_3} \WHP \XXX G_j \lor
\FBV_{j \in \FPI_4} \WHN \XXX H_j \lor 
\FBV_{j \in \FPI_5} l_j \lor
\FBV_{j \in \FPI_6} \YYY'^{m_j} I_j \lor 
\FBV_{j \in \FPI_7} \YYY^{n_j} J_j
$

$\Leftrightarrow$ (by $\FPI_7$ is empty)

$
\FBV_{j \in \FPI_1} \SHP \XXX E_j \lor
\FBV_{j \in \FPI_2} \SHN \XXX F_j \lor
\FBV_{j \in \FPI_3} \WHP \XXX G_j \lor
\FBV_{j \in \FPI_4} \WHN \XXX H_j \lor 
\FBV_{j \in \FPI_5} l_j \lor
\FBV_{j \in \FPI_6} \YYY'^{m_j} I_j
$

$\Leftrightarrow$ (by the item \ref{derivative formulas: diamond disjunction} in Lemma \ref{lemma:make T out})

$\vdash 
\FBV_{j \in \FPI_1} \SHP \XXX E_j \lor
\FBV_{j \in \FPI_2} \SHN \XXX F_j \lor
\FBV_{j \in \FPI_3} \WHP \XXX G_j \lor
\FBV_{j \in \FPI_4} \WHN \XXX H_j \lor 
\FBV_{j \in \FPI_5} l_j \lor
\YYY' \FBV_{j \in \FPI_6} \YYY'^{m_j - 1} I_j
$

$\Leftrightarrow$ (by the second item in Lemma \ref{lemma:facts for reduction of Y})

$\vdash 
\FBV_{j \in \FPI_1} \SHP \XXX E_j \lor
\FBV_{j \in \FPI_2} \SHN \XXX F_j \lor
\FBV_{j \in \FPI_3} \WHP \XXX G_j \lor
\FBV_{j \in \FPI_4} \WHN \XXX H_j \lor 
\FBV_{j \in \FPI_5} l_j.
$

\item

$
\vdash
\FBV_{j \in \FPI_1} \SHP \XXX E_j \lor
\FBV_{j \in \FPI_2} \SHN \XXX F_j \lor
\FBV_{j \in \FPI_3} \WHP \XXX G_j \lor
\FBV_{j \in \FPI_4} \WHN \XXX H_j \lor 
\FBV_{j \in \FPI_5} l_j \lor
\FBV_{j \in \FPI_6} \YYY'^{m_j} I_j \lor 
\FBV_{j \in \FPI_7} \YYY^{n_j} J_j
$

$\Leftrightarrow$ (by the item \ref{derivative formulas: diamond disjunction} in Lemma \ref{lemma:make T out})

$
\vdash
\FBV_{j \in \FPI_1} \SHP \XXX E_j \lor
\FBV_{j \in \FPI_2} \SHN \XXX F_j \lor
\FBV_{j \in \FPI_3} \WHP \XXX G_j \lor
\FBV_{j \in \FPI_4} \WHN \XXX H_j \lor 
\FBV_{j \in \FPI_5} l_j \lor
\FBV_{j \in \FPI_6} \YYY'^{m_j} I_j \lor 
\YYY \FBV_{j \in \FPI_7} \YYY^{n_j - 1} J_j
$

$\Leftrightarrow$ (by the third item in Lemma \ref{lemma:facts for reduction of Y})

$
\vdash
\XXX \Big(
\FBV_{j \in \FPI_1} \SHP \XXX E_j \lor
\FBV_{j \in \FPI_2} \SHN \XXX F_j \lor
\FBV_{j \in \FPI_3} \WHP \XXX G_j \lor
\FBV_{j \in \FPI_4} \WHN \XXX H_j \lor 
\FBV_{j \in \FPI_5} l_j \lor
\FBV_{j \in \FPI_6} \YYY'^{m_j} I_j
\Big)
\lor 
\FBV_{j \in \FPI_7} \YYY^{n_j - 1} J_j
$

$\Leftrightarrow$ (by the item \ref{derivative formulas: diamond disjunction} in Lemma \ref{lemma:make T out})

{\small
$
\vdash
\XXX \Big(
\FBV_{j \in \FPI_1} \SHP \XXX E_j \lor
\FBV_{j \in \FPI_2} \SHN \XXX F_j \lor
\FBV_{j \in \FPI_3} \WHP \XXX G_j \lor
\FBV_{j \in \FPI_4} \WHN \XXX H_j \lor 
\FBV_{j \in \FPI_5} l_j
\Big)
\lor
\XXX \FBV_{j \in \FPI_6} \YYY'^{m_j} I_j
\lor 
\FBV_{j \in \FPI_7} \YYY^{n_j - 1} J_j
$
}

$\Leftrightarrow$ (by Lemma \ref{lemma:X Y})

{\small
$
\vdash
\XXX \Big(
\FBV_{j \in \FPI_1} \SHP \XXX E_j \lor
\FBV_{j \in \FPI_2} \SHN \XXX F_j \lor
\FBV_{j \in \FPI_3} \WHP \XXX G_j \lor
\FBV_{j \in \FPI_4} \WHN \XXX H_j \lor 
\FBV_{j \in \FPI_5} l_j
\Big)
\lor
\FBV_{j \in \FPI_6} \YYY'^{m_j - 1} I_j
\lor 
\FBV_{j \in \FPI_7} \YYY^{n_j - 1} J_j
$
}

$\Leftrightarrow$ (by the rule of generalization of $\SHN$ and the axiom \ref{axiom:SHN T})

$
\vdash
\SHN \bigg(
\XXX \Big(
\FBV_{j \in \FPI_1} \SHP \XXX E_j \lor
\FBV_{j \in \FPI_2} \SHN \XXX F_j \lor
\FBV_{j \in \FPI_3} \WHP \XXX G_j \lor
\FBV_{j \in \FPI_4} \WHN \XXX H_j \lor 
\FBV_{j \in \FPI_5} l_j
\Big)
\lor
\FBV_{j \in \FPI_6} \YYY'^{m_j - 1} I_j
\lor 
\FBV_{j \in \FPI_7} \YYY^{n_j - 1} J_j
\bigg)
$

$\Leftrightarrow$ (note $\FBV_{j \in \FPI_6} \YYY'^{m_j - 1} I_j
\lor 
\FBV_{j \in \FPI_7} \YYY^{n_j - 1} J_j$ is a state formula; then use Lemma \ref{lemma:Box moment formula})

$
\vdash
\SHN
\XXX \Big(
\FBV_{j \in \FPI_1} \SHP \XXX E_j \lor
\FBV_{j \in \FPI_2} \SHN \XXX F_j \lor
\FBV_{j \in \FPI_3} \WHP \XXX G_j \lor
\FBV_{j \in \FPI_4} \WHN \XXX H_j \lor 
\FBV_{j \in \FPI_5} l_j
\Big)
\lor
\FBV_{j \in \FPI_6} \YYY'^{m_j - 1} I_j
\lor 
\FBV_{j \in \FPI_7} \YYY^{n_j - 1} J_j
$

\end{enumerate}

\end{proof}

\begin{lemma}[Reduction of $\YYY$ in standard disjunctions of $\Phi_\YnbHXb$ at the level of validity]
\label{lemma:Reduction of YYY at the level of validity}
~

\begin{enumerate}[label=(\arabic*),leftmargin=3.33em]

\item 

$\models
\FBV_{j \in \FPI_1} \SHP \XXX E_j \lor
\FBV_{j \in \FPI_2} \SHN \XXX F_j \lor
\FBV_{j \in \FPI_3} \WHP \XXX G_j \lor
\FBV_{j \in \FPI_4} \WHN \XXX H_j \lor 
\FBV_{j \in \FPI_5} l_j \lor
\FBV_{j \in \FPI_6} \YYY'^{m_j} I_j \lor 
\FBV_{j \in \FPI_7} \YYY^{n_j} J_j
$

$\Leftrightarrow$

$\models
\FBV_{j \in \FPI_1} \SHP \XXX E_j \lor
\FBV_{j \in \FPI_2} \SHN \XXX F_j \lor
\FBV_{j \in \FPI_3} \WHP \XXX G_j \lor
\FBV_{j \in \FPI_4} \WHN \XXX H_j \lor 
\FBV_{j \in \FPI_5} l_j
$,

where $\FPI_7$ is empty.

\item

$
\models
\FBV_{j \in \FPI_1} \SHP \XXX E_j \lor
\FBV_{j \in \FPI_2} \SHN \XXX F_j \lor
\FBV_{j \in \FPI_3} \WHP \XXX G_j \lor
\FBV_{j \in \FPI_4} \WHN \XXX H_j \lor 
\FBV_{j \in \FPI_5} l_j \lor
\FBV_{j \in \FPI_6} \YYY'^{m_j} I_j \lor 
\FBV_{j \in \FPI_7} \YYY^{n_j} J_j
$

$\Leftrightarrow$

$\models
\SHN
\XXX \Big(
\FBV_{j \in \FPI_1} \SHP \XXX E_j \lor
\FBV_{j \in \FPI_2} \SHN \XXX F_j \lor
\FBV_{j \in \FPI_3} \WHP \XXX G_j \lor
\FBV_{j \in \FPI_4} \WHN \XXX H_j \lor 
\FBV_{j \in \FPI_5} l_j
\Big)
\lor
\FBV_{j \in \FPI_6} \YYY'^{m_j - 1} I_j
\lor 
\FBV_{j \in \FPI_7} \YYY^{n_j - 1} J_j
$,

where $\FPI_7$ is not empty.

\end{enumerate}

\end{lemma}

\begin{proof}
~

\begin{enumerate}[label=(\arabic*),leftmargin=3.33em]

\item

$\models
\FBV_{j \in \FPI_1} \SHP \XXX E_j \lor
\FBV_{j \in \FPI_2} \SHN \XXX F_j \lor
\FBV_{j \in \FPI_3} \WHP \XXX G_j \lor
\FBV_{j \in \FPI_4} \WHN \XXX H_j \lor 
\FBV_{j \in \FPI_5} l_j \lor
\FBV_{j \in \FPI_6} \YYY'^{m_j} I_j \lor 
\FBV_{j \in \FPI_7} \YYY^{n_j} J_j
$

$\Leftrightarrow$ (by $\FPI_7$ is empty)

$
\FBV_{j \in \FPI_1} \SHP \XXX E_j \lor
\FBV_{j \in \FPI_2} \SHN \XXX F_j \lor
\FBV_{j \in \FPI_3} \WHP \XXX G_j \lor
\FBV_{j \in \FPI_4} \WHN \XXX H_j \lor 
\FBV_{j \in \FPI_5} l_j \lor
\FBV_{j \in \FPI_6} \YYY'^{m_j} I_j
$

$\Leftrightarrow$ (by the item \ref{derivative formulas: diamond disjunction} in Lemma \ref{lemma:make T out} and the soundness of $\SWHTN$)

$\models 
\FBV_{j \in \FPI_1} \SHP \XXX E_j \lor
\FBV_{j \in \FPI_2} \SHN \XXX F_j \lor
\FBV_{j \in \FPI_3} \WHP \XXX G_j \lor
\FBV_{j \in \FPI_4} \WHN \XXX H_j \lor 
\FBV_{j \in \FPI_5} l_j \lor
\YYY' \FBV_{j \in \FPI_6} \YYY'^{m_j - 1} I_j
$

$\Leftrightarrow$ (by the second item in Lemma \ref{lemma:facts for reduction of Y})

$\models 
\FBV_{j \in \FPI_1} \SHP \XXX E_j \lor
\FBV_{j \in \FPI_2} \SHN \XXX F_j \lor
\FBV_{j \in \FPI_3} \WHP \XXX G_j \lor
\FBV_{j \in \FPI_4} \WHN \XXX H_j \lor 
\FBV_{j \in \FPI_5} l_j
$

\item

$
\models
\FBV_{j \in \FPI_1} \SHP \XXX E_j \lor
\FBV_{j \in \FPI_2} \SHN \XXX F_j \lor
\FBV_{j \in \FPI_3} \WHP \XXX G_j \lor
\FBV_{j \in \FPI_4} \WHN \XXX H_j \lor 
\FBV_{j \in \FPI_5} l_j \lor
\FBV_{j \in \FPI_6} \YYY'^{m_j} I_j \lor 
\FBV_{j \in \FPI_7} \YYY^{n_j} J_j
$

$\Leftrightarrow$ (by the item \ref{derivative formulas: diamond disjunction} in Lemma \ref{lemma:make T out} and the soundness of $\SWHTN$)

$
\models
\FBV_{j \in \FPI_1} \SHP \XXX E_j \lor
\FBV_{j \in \FPI_2} \SHN \XXX F_j \lor
\FBV_{j \in \FPI_3} \WHP \XXX G_j \lor
\FBV_{j \in \FPI_4} \WHN \XXX H_j \lor 
\FBV_{j \in \FPI_5} l_j \lor
\FBV_{j \in \FPI_6} \YYY'^{m_j} I_j \lor 
\YYY \FBV_{j \in \FPI_7} \YYY^{n_j - 1} J_j
$

$\Leftrightarrow$ (by the third item in Lemma \ref{lemma:facts for reduction of Y})

$
\models
\XXX \Big(
\FBV_{j \in \FPI_1} \SHP \XXX E_j \lor
\FBV_{j \in \FPI_2} \SHN \XXX F_j \lor
\FBV_{j \in \FPI_3} \WHP \XXX G_j \lor
\FBV_{j \in \FPI_4} \WHN \XXX H_j \lor 
\FBV_{j \in \FPI_5} l_j \lor
\FBV_{j \in \FPI_6} \YYY'^{m_j} I_j
\Big)
\lor 
\FBV_{j \in \FPI_7} \YYY^{n_j - 1} J_j
$

$\Leftrightarrow$ (by the item \ref{derivative formulas: diamond disjunction} in Lemma \ref{lemma:make T out} and the soundness of $\SWHTN$)

{\small
$
\models
\XXX \Big(
\FBV_{j \in \FPI_1} \SHP \XXX E_j \lor
\FBV_{j \in \FPI_2} \SHN \XXX F_j \lor
\FBV_{j \in \FPI_3} \WHP \XXX G_j \lor
\FBV_{j \in \FPI_4} \WHN \XXX H_j \lor 
\FBV_{j \in \FPI_5} l_j
\Big)
\lor
\XXX \FBV_{j \in \FPI_6} \YYY'^{m_j} I_j
\lor 
\FBV_{j \in \FPI_7} \YYY^{n_j - 1} J_j
$
}

$\Leftrightarrow$ (by Lemma \ref{lemma:X Y} and the soundness of $\SWHTN$)

{\small
$
\models
\XXX \Big(
\FBV_{j \in \FPI_1} \SHP \XXX E_j \lor
\FBV_{j \in \FPI_2} \SHN \XXX F_j \lor
\FBV_{j \in \FPI_3} \WHP \XXX G_j \lor
\FBV_{j \in \FPI_4} \WHN \XXX H_j \lor 
\FBV_{j \in \FPI_5} l_j
\Big)
\lor
\FBV_{j \in \FPI_6} \YYY'^{m_j - 1} I_j
\lor 
\FBV_{j \in \FPI_7} \YYY^{n_j - 1} J_j
$}

$\Leftrightarrow$ (by the rule of generalization of $\SHN$ preserving validity, the axiom \ref{axiom:SHN T}, and the soundness of $\SWHTN$)

$
\models
\SHN \bigg(
\XXX \Big(
\FBV_{j \in \FPI_1} \SHP \XXX E_j \lor
\FBV_{j \in \FPI_2} \SHN \XXX F_j \lor
\FBV_{j \in \FPI_3} \WHP \XXX G_j \lor
\FBV_{j \in \FPI_4} \WHN \XXX H_j \lor 
\FBV_{j \in \FPI_5} l_j
\Big)
\lor
\FBV_{j \in \FPI_6} \YYY'^{m_j - 1} I_j
\lor 
\FBV_{j \in \FPI_7} \YYY^{n_j - 1} J_j
\bigg)
$

$\Leftrightarrow$ (by Lemma \ref{lemma:Box moment formula} and the soundness of $\SWHTN$; note $\FBV_{j \in \FPI_6} \YYY'^{m_j - 1} I_j
\lor 
\FBV_{j \in \FPI_7} \YYY^{n_j - 1} J_j$ is a state formula)

$
\models
\SHN
\XXX \Big(
\FBV_{j \in \FPI_1} \SHP \XXX E_j \lor
\FBV_{j \in \FPI_2} \SHN \XXX F_j \lor
\FBV_{j \in \FPI_3} \WHP \XXX G_j \lor
\FBV_{j \in \FPI_4} \WHN \XXX H_j \lor 
\FBV_{j \in \FPI_5} l_j
\Big)
\lor
\FBV_{j \in \FPI_6} \YYY'^{m_j - 1} I_j
\lor 
\FBV_{j \in \FPI_7} \YYY^{n_j - 1} J_j
$

\end{enumerate}

\end{proof}

\begin{lemma}[V-reducibility of $\Phi_{\YnbHXb}$ to $\Phi_\HX$]
\label{lemma:v-reducibility of Yn(HX) to HX}

$\Phi_{\YnbHXb}$ is v-reducible to $\Phi_\HX$.

\end{lemma}

\begin{proof}
~

Let $\phi$ be in $\Phi_{\YnbHXb}$. We describe a procedure to get a formula $\psi$ in $\Phi_\HX$ such that (1) $\vdash \phi$ if and only if $\vdash \psi$, and (2) $\models \phi$ if and only if $\models \psi$. It follows that $\Phi_{\YnbHXb}$ is v-reducible to $\Phi_\HX$.

By Lemma \ref{lemma:Normal form of formulas in Yn(HX)} and the soundness of $\SWHTN$, there is $\phi'$ in $\Phi_{\YnbHXb}$ such that (1) $\vdash \phi \leftrightarrow \phi'$, (2) $\models \phi \leftrightarrow \phi'$, (3) $\phi'$ is in the form $\chi_1 \land \dots \land \chi_n$, where every $\chi_i$ is a standard disjunction of $\Phi_\YnbHXb$ with the $\YYY$-depth not greater than the $\YYY$-depth of $\phi$.

Pick a $\chi_i$. Let $\chi_i$ be in the form
\[ 
\FBV_{j \in \FPI_1} \SHP \XXX E_j \lor
\FBV_{j \in \FPI_2} \SHN \XXX F_j \lor
\FBV_{j \in \FPI_3} \WHP \XXX G_j \lor
\FBV_{j \in \FPI_4} \WHN \XXX H_j \lor 
\FBV_{j \in \FPI_5} l_j \lor
\FBV_{j \in \FPI_6} \YYY'^{m_j} I_j \lor 
\FBV_{j \in \FPI_7} \YYY^{n_j} J_j
\]

Assume $\FPI_7 = \emptyset$.
Let $\chi'_i = 
\FBV_{j \in \FPI_1} \SHP \XXX E_j \lor
\FBV_{j \in \FPI_2} \SHN \XXX F_j \lor
\FBV_{j \in \FPI_3} \WHP \XXX G_j \lor
\FBV_{j \in \FPI_4} \WHN \XXX H_j \lor 
\FBV_{j \in \FPI_5} l_j
$.
Note the $\YYY$-depth of $\chi_i'$ is lower than the $\YYY$-depth of $\chi_i$. By Lemmas \ref{lemma:Reduction of YYY at the level of derivability} and \ref{lemma:Reduction of YYY at the level of validity}, (1) $\vdash \chi_i$ if and only if $\vdash \chi'_i$, and (2) $\models \chi_i$ if and only if $\models \chi'_i$. 

Assume $\FPI_7 \neq \emptyset$.
Let $\chi'_i = 
\FBV_{j \in \FPI_1} \SHP \XXX E_j \lor
\FBV_{j \in \FPI_2} \SHN \XXX F_j \lor
\FBV_{j \in \FPI_3} \WHP \XXX G_j \lor
\FBV_{j \in \FPI_4} \WHN \XXX H_j \lor 
\FBV_{j \in \FPI_5} l_j \lor
\FBV_{j \in \FPI_6} \YYY'^{m_j - 1} I_j \lor 
\FBV_{j \in \FPI_7} \YYY^{n_j - 1} J_j
$.
Note the $\YYY$-depth of $\chi_i'$ is lower than the $\YYY$-depth of $\chi_i$. By Lemmas \ref{lemma:Reduction of YYY at the level of derivability} and \ref{lemma:Reduction of YYY at the level of validity}, (1) $\vdash \chi_i$ if and only if $\vdash \chi'_i$, and (2) $\models \chi_i$ if and only if $\models \chi'_i$. 

Note $\chi_i$ is arbitrary. Thus, (1) $\vdash \chi_1 \land \dots \land \chi_n$ if and only if $\vdash \chi'_1 \land \dots \land \chi'_n$, and (2) $\models \chi_1 \land \dots \land \chi_n$ if and only if $\models \chi'_1 \land \dots \land \chi'_n$. Note the $\YYY$-depth of $\chi'_1 \land \dots \land \chi'_n$ is lower than the $\YYY$-depth of $\chi_1 \land \dots \land \chi_n$.

Repeat and we will get a formula $\psi$ in $\Phi_\HX$ such that (1) $\vdash \phi$ if and only if $\vdash \psi$, and (2) $\models \phi$ if and only if $\models \psi$.

\end{proof}

\newcommand{\FPL}{\mathtt{PL}}

\subsection{Completeness for $\Phi_{\HX}$}

In this subsection, we treat $\SHN \XXX$ and $\WHN \XXX$ as single modal operators, and define the \Fdefs{modal depth} of formulas of $\Phi_{\HX}$ in the usual way.

\begin{definition}[Standard formulas of $\Phi_\HX$]

A formula $\phi$ of $\Phi_\HX$ is called a \Fdefs{standard formula} of $\Phi_\HX$ if $\phi$ is in the form
\[(\SHN \XXX A \land \WHN \XXX B) \rightarrow 
(
\FBV_{i \in \FPI_1} \SHN \XXX C_i \lor \FBV_{i \in \FPI_2} \WHN \XXX D_i \lor \FBV_{i \in \FPI_3} l_i
),\]

\noindent where $\FPI_1, \FPI_2$ and $\FPI_3$ are pairwise disjoint index sets, $\FPI_1 \neq \emptyset$, and every $l_i$ is a literal.

\end{definition}

\begin{lemma}[Normal form lemma for $\Phi_\HX$]
\label{lemma:Normal form of formulas in HX}

For every $\psi \in \Phi_\HX$ with the modal depth greater than $0$, there is $\psi' \in \Phi_\HX$ such that (1) $\vdash \psi \leftrightarrow \psi'$, and (2) $\psi'$ is a standard formula of $\Phi_\HX$ with the same modal depth as $\psi$.

\end{lemma}

\newcommand{\FNI}{\mathbb{I}}

\begin{proof}~

Let $\psi \in \Phi_\HX$ with the modal depth greater than $0$. We can transform $\psi$ to $\chi = \chi_1 \land \dots \land \chi_n$ in the conjunctive normal form, where every $\chi_i$ is in the form
\[
\FBV_{i \in \FNI_1} \neg \SHN \XXX A_i \lor \FBV_{i \in \FNI_2} \neg \WHN \XXX B_i \lor
\FBV_{i \in \FPI_1} \SHN \XXX C_i \lor \FBV_{i \in \FPI_2} \WHN \XXX D_i \lor \FBV_{i \in \FPI_3} l_i,
\]

\noindent where $\FNI_1, \FNI_2, \FPI_1, \FPI_2$ and $\FPI_3$ are pairwise disjoint index sets, and every $l_i$ is a literal. Note the modal depth of $\chi$ is the same as $\psi$.

It is easy to show $\vdash \SHN \XXX \top$, $\vdash \WHN \XXX \top$, and $\vdash \neg \SHN \XXX \bot$. Then we can assume $\FNI_1, \FNI_2$, and $\FPI_1$ are nonempty. Note the modal depth of $\chi$ is greater than $0$. Then these assumptions do not change the modal depth of $\chi$.

By Lemma \ref{lemma:make T out}, we can show $\vdash (\SHN \XXX \alpha \land \SHN \XXX \beta) \leftrightarrow \SHN \XXX (\alpha \land \beta)$ and $\vdash (\WHN \XXX \alpha \land \WHN \XXX \beta) \leftrightarrow \SHN \XXX (\alpha \land \beta)$. Then the following equivalences are derivable:
\[
\FBV_{i \in \FNI_1} \neg \SHN \XXX A_i \lor \FBV_{i \in \FNI_2} \neg \WHN \XXX B_i \lor
\FBV_{i \in \FPI_1} \SHN \XXX C_i \lor \FBV_{i \in \FPI_2} \WHN \XXX D_i \lor \FBV_{i \in \FPI_3} l_i
\]
\[\updownarrow\]
\[
\neg \SHN (\FBV_{i \in \FNI_1} \XXX A_i) \lor \neg \WHN (\FBV_{i \in \FNI_2} \XXX B_i) \lor
\FBV_{i \in \FPI_1} \SHN \XXX C_i \lor \FBV_{i \in \FPI_2} \WHN \XXX D_i \lor \FBV_{i \in \FPI_3} l_i
\]
\[\updownarrow\]
\[
\Big( \SHN (\FBV_{i \in \FNI_1} \XXX A_i) \land \WHN (\FBV_{i \in \FNI_2} \XXX B_i) \Big) \rightarrow
\Big( \FBV_{i \in \FPI_1} \SHN \XXX C_i \lor \FBV_{i \in \FPI_2} \WHN \XXX D_i \lor \FBV_{i \in \FPI_3} l_i \Big)
\]

Let $\psi' = \Big( \SHN (\FBV_{i \in \FNI_1} \XXX A_i) \land \WHN (\FBV_{i \in \FNI_2} \XXX B_i) \Big) \rightarrow
\Big( \FBV_{i \in \FPI_1} \SHN \XXX C_i \lor \FBV_{i \in \FPI_2} \WHN \XXX D_i \lor \FBV_{i \in \FPI_3} l_i \Big)$. It is easy to see $\vdash \psi \leftrightarrow \psi'$ and $\psi'$ is a standard formula of $\Phi_\HX$ with the same modal depth as $\psi$.

\end{proof}

\begin{lemma}[Special downward validity lemma for $\Phi_\HX$]
\label{lemma: Special downward validity lemma for HX}

Let $(\SHN \XXX A \land \WHN \XXX B) \rightarrow 
(
\FBV_{i \in \FPI_1} \SHN \XXX C_i \lor \FBV_{i \in \FPI_2} \WHN \XXX D_i \lor \FBV_{i \in \FPI_3} l_i
)$ be a standard formula of $\Phi_\HX$, where $A, B$, and all $C_i$ and $D_j$ are propositional formulas.

Assume $\models (\SHN \XXX A \land \WHN \XXX B) \rightarrow 
(
\FBV_{i \in \FPI_1} \SHN \XXX C_i \lor \FBV_{i \in \FPI_2} \WHN \XXX D_i \lor \FBV_{i \in \FPI_3} l_i
)$.

Then, one of the following conditions holds:
\begin{enumerate}[label=(\alph*),leftmargin=3.33em]

\item

there is $i \in \FPI_1$ such that $\models A \rightarrow C_i$;

\item

there is $j \in \FPI_2$ such that $\models ( A \land B ) \rightarrow D_j$;

\item 

$\models \FBV_{i \in \FPI_3} l_i$.

\end{enumerate}

\end{lemma}

\begin{proof}
~

Assume none of the three conditions holds. Then:
\begin{enumerate}[label=(\alph*),leftmargin=3.33em]

\item

for all $i \in \FPI_1$, $A \land \neg C_i$ is satisfiable;

\item

for all $j \in \FPI_2$, $A \land B \land \neg D_j$ is satisfiable;

\item 

$\neg \FBV_{i \in \FPI_3} l_i$ is satisfiable.

\end{enumerate}

\noindent It suffices to show $\not \models (\SHN \XXX A \land \WHN \XXX B) \rightarrow 
(
\FBV_{i \in \FPI_1} \SHN \XXX C_i \lor \FBV_{i \in \FPI_2} \WHN \XXX D_i \lor \FBV_{i \in \FPI_3} l_i
)$.

Note that $A, B$, and all $C_i$ and $D_j$ are propositional formulas. It is easy to see that there is a contextualized pointed model $(\MM, \CC, \pi, 0)$ meeting the following conditions:
\begin{itemize}

\item

$\MM$ has $|\FPI_1| + |\FPI_2|$ timelines.

\emph{For every $i \in \FPI_1$, let $\pi_i$ be a timeline of $\MM$, and for every $j \in \FPI_2$, let $\tau_j$ be a timeline of $\MM$.}
\emph{Note $\FPI_1 \neq \emptyset$ is not empty. Thus, $\MM$ has at least one timeline.}

\item 

$\CC = (\AT,\ET)$, where $\AT = \{ \pi_i \mid i \in \FPI_1 \} \cup \{\tau_j \mid j \in \FPI_2\}$ and $\ET = \{\tau_j \mid j \in \FPI_2\}$.

\item 

$\pi \in \AT$.

\emph{Note $\AT$ is not empty.}

\item 

\begin{itemize}

\item 

for every $i \in \FPI_1$, $\MM,\CC, \pi_i, 1 \Vdash A \land \neg C_i$;

\item 

for every $j \in \FPI_2$, $\MM,\CC, \tau_j, 1 \Vdash A \land B \land \neg D_j$;

\item 

$\MM, \CC, \pi, 0 \Vdash \neg \FBV_{i \in \FPI_3} l_i$.

\end{itemize}

\end{itemize}

It is easy to check $\MM, \CC, \pi, 0 \not \Vdash (\SHN \XXX A \land \WHN \XXX B) \rightarrow 
(
\FBV_{i \in \FPI_1} \SHN \XXX C_i \lor \FBV_{i \in \FPI_2} \WHN \XXX D_i \lor \FBV_{i \in \FPI_3} l_i
)$.

\end{proof}

\begin{lemma}[Special upward derivability lemma for $\Phi_\HX$]
\label{lemma: Special upward derivability lemma for HX}

Let $(\SHN \XXX A \land \WHN \XXX B) \rightarrow 
(
\FBV_{i \in \FPI_1} \SHN \XXX C_i \lor \FBV_{i \in \FPI_2} \WHN \XXX D_i \lor \FBV_{i \in \FPI_3} l_i
)$ be a standard formula of $\Phi_\HX$, where $A, B$, all $C_i$ and $D_j$ are propositional formulas.

Assume one of the following conditions holds:
\begin{enumerate}[label=(\alph*),leftmargin=3.33em]

\item

there is $i \in \FPI_1$ such that $\vdash A \rightarrow C_i$;

\item

there is $j \in \FPI_2$ such that $\vdash ( A \land B ) \rightarrow D_j$;

\item 

$\vdash \FBV_{i \in \FPI_3} l_i$.

\end{enumerate}

Then, $\vdash (\SHN \XXX A \land \WHN \XXX B) \rightarrow 
(
\FBV_{i \in \FPI_1} \SHN \XXX C_i \lor \FBV_{i \in \FPI_2} \WHN \XXX D_i \lor \FBV_{i \in \FPI_3} l_i
)$.

\end{lemma}

\begin{proof}
~

\begin{enumerate}[label=(\alph*),leftmargin=3.33em]

\item

Assume there is $i \in \FPI_1$ such that $\vdash A \rightarrow C_i$.

By the rule of necessitation of $\XXX$ and the axiom \ref{axiom:X K}, we can get $\vdash \XXX A \rightarrow \XXX C_i$. By the rule of necessitation of $\SHN$ and the axiom \ref{axiom:SHN K}, we can get $\vdash \SHN \XXX A \rightarrow \SHN \XXX C_i$.
Then 
$\vdash (\SHN \XXX A \land \WHN \XXX B) \rightarrow 
(
\FBV_{i \in \FPI_1} \SHN \XXX C_i \lor \FBV_{i \in \FPI_2} \WHN \XXX D_i \lor \FBV_{i \in \FPI_3} l_i
)$.

\item

Assume there is $j \in \FPI_2$ such that $\vdash (A \land B) \rightarrow D_j$.

By the rule of necessitation of $\XXX$ and the axiom \ref{axiom:X K}, we can get $\vdash \XXX (A \land B) \rightarrow \XXX D_j$.
By Lemma \ref{lemma:make T out} and the rule of replacement of equivalence, we have $\vdash (\XXX A \land \XXX B) \rightarrow \XXX D_j$.
By the rule of necessitation of $\WHN$ and the axiom \ref{axiom:WHN K}, we can get $\vdash \WHN (\XXX A \land \XXX B) \rightarrow \WHN \XXX D_j$.
By Lemma \ref{lemma:make T out} and the rule of replacement of equivalence, we can get $\vdash (\WHN \XXX A \land \WHN \XXX B) \rightarrow \WHN \XXX D_j$.
By the axiom \ref{axiom:SHN is stronger than WHN}, $\vdash (\SHN \XXX A \land \WHN \XXX B) \rightarrow \WHN \XXX D_j$.
Then 
$\vdash (\SHN \XXX A \land \WHN \XXX B) \rightarrow 
(
\FBV_{i \in \FPI_1} \SHN \XXX C_i \lor \FBV_{i \in \FPI_2} \WHN \XXX D_i \lor \FBV_{i \in \FPI_3} l_i
)$.

\item 

Assume $\vdash \FBV_{i \in \FPI_3} l_i$.

It is easy to see $\vdash (\SHN \XXX A \land \WHN \XXX B) \rightarrow 
(
\FBV_{i \in \FPI_1} \SHN \XXX C_i \lor \FBV_{i \in \FPI_2} \WHN \XXX D_j \lor \FBV_{j \in \FPI_3} l_i
)$.

\end{enumerate}

\end{proof}

\begin{lemma}[Downward validity lemma for $\Phi_\HX$]
\label{lemma:Downward validity for HX}

Let $(\SHN \XXX A \land \WHN \XXX B) \rightarrow 
(
\FBV_{i \in \FPI_1} \SHN \XXX C_i \lor \FBV_{i \in \FPI_2} \WHN \XXX D_i \lor \FBV_{i \in \FPI_3} l_i
)$ be a standard formula of $\Phi_\HX$.

Assume $\models (\SHN \XXX A \land \WHN \XXX B) \rightarrow 
(
\FBV_{i \in \FPI_1} \SHN \XXX C_i \lor \FBV_{i \in \FPI_2} \WHN \XXX D_i \lor \FBV_{i \in \FPI_3} l_i
)$.

Then, one of the following conditions holds:
\begin{enumerate}[label=(\alph*),leftmargin=3.33em]

\item

there is $i \in \FPI_1$ such that $\models \Big( A \land (\WHP \XXX \top \rightarrow B) \Big) \rightarrow C_i$;

\item

there is $j \in \FPI_2$ such that $\models \Big( A \land (\WHP \XXX \top \rightarrow B) \Big) \rightarrow ( \WHP \XXX \top \rightarrow D_j )$;

\item 

$\models \FBV_{i \in \FPI_3} l_i$.

\end{enumerate}

\end{lemma}

\begin{proof}
~

Assume none of the three conditions holds. Then:
\begin{enumerate}[label=(\alph*),leftmargin=3.33em]

\item

for all $i \in \FPI_1$, $A \land (\WHP \XXX \top \rightarrow B) \land \neg C_i$ is satisfiable;

\item

for all $j \in \FPI_2$, $A \land B \land \WHP \XXX \top \land \neg D_j$ is satisfiable;

\item 

$\neg \FBV_{i \in \FPI_3} l_i$ is satisfiable.

\end{enumerate}

\noindent It suffices to show $\not \models (\SHN \XXX A \land \WHN \XXX B) \rightarrow 
(
\FBV_{i \in \FPI_1} \SHN \XXX C_i \lor \FBV_{i \in \FPI_2} \WHN \XXX D_i \lor \FBV_{i \in \FPI_3} l_i
)$.

By Lemma \ref{lemma:generated submodels}, we have the following: \emph{for every $\phi$ of $\Phi_\HX$, if $\phi$ is true at a contextualized pointed model $(\MM,\CC,\pi,i)$, then $\phi$ is true at a contextualized pointed model $(\MM',\CC',\pi',0)$}. Thus, we can do the following:
\begin{itemize}

\item 

For every $i \in \FPI_1$, let $(\MM_i, \CC_i, \pi_i, 0)$ be a contextualized pointed model satisfying $A \land (\WHP \XXX \top \rightarrow B) \land \neg C_i$, where $\MM_i = (W_i, w_i, <_i, V_i)$ and $\CC_i = (\AT_i, \ET_i)$.

\item

For every $j \in \FPI_2$, let $(\MM_j, \CC_j, \pi_j, 0)$ be a contextualized pointed model satisfying $A \land B \land \WHP \XXX \top \land \neg D_j$, where $\MM_j = (W_j, w_j, <_j, V_j)$ and $\CC_j = (\AT_j, \ET_j)$.

\item

Let $(\MM', \CC', \pi', 0)$ be a contextualized pointed model satisfying $\neg \FBV_{i \in \FPI_3} l_i$, where $\MM' = (W', w',$ $<', V')$ and $\CC' = (\AT', \ET')$.

\end{itemize}

\noindent We assume the domains of these models are pairwise disjoint.

Let $w$ be a new state not in any of these models. Define a contextualized pointed model $(\MM, \CC, \tau, 0)$, where $\MM = (W,w,<,V)$ and $\CC = (\AT, \ET)$, as follows:
\begin{itemize}

\item 

$W = \{w\} \cup \bigcup \{W_i \mid i \in \FPI_1 \} \cup \{W_j \mid j \in \FPI_2 \}$.

\item 

$< = \{(w,w_i) \mid i \in \FPI_1\} \cup \{(w,w_j) \mid j \in \FPI_2\} \cup \bigcup \{<_i \mid i \in \FPI_1 \} \cup \bigcup \{<_j \mid j \in \FPI_2 \}$.

\item 

For every $p$, $V(p) = V'(p) \cup \bigcup \{V_i (p) \mid i \in \FPI_1 \} \cup \bigcup \{V_j (p) \mid j \in \FPI_2 \}$.

\item 

For every $i \in \FPI_1$, let $\AT'_i$ be the result of prefixing every $\pi$ in $\AT_i$ with $w$, and let $\ET'_i$ be the result of prefixing every $\pi$ in $\ET_i$ with $w$.

For every $j \in \FPI_2$, let $\AT'_j$ be the result of prefixing every $\pi$ in $\AT_j$ with $w$, and let $\ET'_j$ be the result of prefixing every $\pi$ in $\ET_j$ with $w$.

Let $\AT = \bigcup \{\AT'_i \mid i \in \FPI_1\} \cup \bigcup \{\AT'_j \mid j \in \FPI_2\}$.

Let $\ET = \bigcup \{\ET'_i \mid i \in \FPI_1\} \cup \bigcup \{\ET'_j \mid j \in \FPI_2\}$.

\item 

$\tau \in \AT$.

\emph{Note $\FPI_1$ is not empty. Thus, $\AT$ is not empty.}

\end{itemize}

The following can be verified:
\begin{itemize}

\item 

for every $i \in \FPI_1$, $\MM, \CC, (w,\pi_i), 1 \Vdash A \land (\WHP \XXX \top \rightarrow B) \land \neg C_i$;

\item 

for every $j \in \FPI_2$, $\MM, \CC, (w,\pi_j), 1 \Vdash A \land B \land \WHP \XXX \top \land \neg D_j$;

\item 

$\MM, \CC, \tau, 0 \Vdash \neg \FBV_{i \in \FPI_3} l_i$.

\end{itemize}

Then, the following can be shown:
\begin{itemize}

\item 

$\MM,\CC,\tau,0 \Vdash \SHN \XXX A$. This is easy to see.

\item 

$\MM,\CC,\tau,0 \Vdash \WHN \XXX B$. How? Let $\pi \in \ET$. It suffices to show $\MM,\CC,\pi,1 \Vdash B$. 
Assume $\pi \in \ET'_i$ for some $i \in \FPI_1$. It can be seen $\MM,\CC,\pi,1 \Vdash \WHP \XXX \top$. Note $\MM, \CC, (w,\pi_i), 1 \Vdash A \land (\WHP \XXX \top \rightarrow B) \land \neg C_i$ and $\pi[1] = (w,\pi_i)[1]$. Then $\MM,\CC,\pi,1 \Vdash B$. Assume $\pi \in \ET'_j$ for some $j \in \FPI_2$. Note $\MM, \CC, (w,\pi_j), 1 \Vdash A \land B \land \WHP \XXX \top \land \neg D_j$ and $\pi[1] = (w,\pi_j)[1]$. Then $\MM,\CC,\pi,1 \Vdash B$.

\item

$\MM,\CC,\tau,0 \Vdash \neg \SHN \XXX C_i$ for every $i \in \FPI_1$. How? Let $i \in \FPI_1$. Note $(w,\pi_i) \in \AT$ and $\MM,\CC,(w,\pi_i),1 \Vdash A \land (\WHP \XXX \top \rightarrow B) \land \neg C_i$. Then $\MM,\CC,\tau,0 \Vdash \neg \SHN \XXX C_i$.

\item

$\MM,\CC,\tau,0 \Vdash \neg \WHN \XXX D_j$ for every $j \in \FPI_2$. How? Let $j \in \FPI_2$. Note $(w,\pi_j) \in \ET$ and $\MM,\CC,(w,\pi_j),1 \Vdash A \land B \land \WHP \XXX \top \land \neg D_j$. Then $\MM,\CC,\tau,0 \Vdash \neg \WHN \XXX D_j$.

\end{itemize}

Then $\MM,\CC,\tau,0 \not \Vdash (\SHN \XXX A \land \WHN \XXX B) \rightarrow 
(
\FBV_{i \in \FPI_1} \SHN \XXX C_i \lor \FBV_{i \in \FPI_2} \WHN \XXX D_i \lor \FBV_{i \in \FPI_3} l_i
)$.

\end{proof}

\begin{lemma}
\label{lemma:the magic formula}

$\vdash (\SHN \XXX A \land \WHN \XXX B) \rightarrow \SHN \XXX \Big( A \land (\WHP \XXX \top \rightarrow B) \Big)$, where $A, B \in \Phi_\HX$.

\end{lemma}

\begin{proof}
~

By the axiom \ref{axiom:defining WHN by SHN}, $\vdash \WHN \XXX B \rightarrow \SHN \XXX (\WHP \XXX \top \rightarrow B)$.
Then $\vdash (\SHN \XXX A \land \WHN \XXX B) \rightarrow \Big( \SHN \XXX A \land \SHN \XXX (\WHP \XXX \top \rightarrow B) \Big)$.
By Lemma \ref{lemma:make T out}, $\vdash \Big( \SHN \XXX A \land \SHN \XXX (\WHP \XXX \top \rightarrow B) \Big) \rightarrow \SHN \XXX \Big( A \land (\WHP \XXX \top \rightarrow B) \Big)$.
Then $\vdash (\SHN \XXX A \land \WHN \XXX B) \rightarrow \SHN \XXX \Big( A \land (\WHP \XXX \top \rightarrow B) \Big)$.

\end{proof}

\begin{lemma}[Upward derivability for $\Phi_\HX$]
\label{lemma:Upward derivability for HX}

Let $(\SHN \XXX A \land \WHN \XXX B) \rightarrow 
(
\FBV_{i \in \FPI_1} \SHN \XXX C_i \lor \FBV_{i \in \FPI_2} \WHN \XXX D_i \lor \FBV_{i \in \FPI_3} l_i
)$ be a standard formula of $\Phi_\HX$.

Assume one of the following conditions holds:
\begin{enumerate}[label=(\alph*),leftmargin=3.33em]

\item

there is $i \in \FPI_1$ such that $\vdash \Big( A \land (\WHP \XXX \top \rightarrow B) \Big) \rightarrow C_i$;

\item

there is $j \in \FPI_2$ such that $\vdash \Big( A \land (\WHP \XXX \top \rightarrow B) \Big) \rightarrow ( \WHP \XXX \top \rightarrow D_j )$;

\item 

$\vdash \FBV_{i \in \FPI_3} l_i$.

\end{enumerate}

Then, $\vdash (\SHN \XXX A \land \WHN \XXX B) \rightarrow 
(
\FBV_{i \in \FPI_1} \SHN \XXX C_i \lor \FBV_{i \in \FPI_2} \WHN \XXX D_i \lor \FBV_{i \in \FPI_3} l_i
)$.

\end{lemma}

\begin{proof}
~

\begin{enumerate}[label=(\alph*),leftmargin=3.33em]

\item 

Assume there is some $i \in \FPI_1$ such that $\vdash \Big( A \land (\WHP \XXX \top \rightarrow B) \Big) \rightarrow C_i$.

By the rule of necessitation of $\XXX$ and the axiom \ref{axiom:X K}, we can get $\vdash \XXX \Big( A \land (\WHP \XXX \top \rightarrow B) \Big) \rightarrow \XXX C_i$. By the rule of necessitation of $\SHN$ and the axiom \ref{axiom:SHN K}, we can get $\vdash \SHN \XXX \Big( A \land (\WHP \XXX \top \rightarrow B) \Big) \rightarrow \SHN \XXX C_i$.
By Lemma \ref{lemma:the magic formula}, $\vdash (\SHN \XXX A \land \WHN \XXX B) \rightarrow \SHN \XXX \Big( A \land (\WHP \XXX \top \rightarrow B) \Big)$.
Then 
$\vdash (\SHN \XXX A \land \WHN \XXX B) \rightarrow 
\SHN \XXX C_i
)$.
Then 
$\vdash (\SHN \XXX A \land \WHN \XXX B) \rightarrow 
(
\FBV_{i \in \FPI_1} \SHN \XXX C_i \lor \FBV_{i \in \FPI_2} \WHN \XXX D_i \lor \FBV_{i \in \FPI_3} l_i
)$.

\item

Assume there is $j \in \FPI_2$ such that $\vdash \Big( A \land (\WHP \XXX \top \rightarrow B) \Big) \rightarrow ( \WHP \XXX \top \rightarrow D_j )$.

By the rule of necessitation of $\XXX$ and the axiom \ref{axiom:X K}, we can get $\vdash \XXX \Big( A \land (\WHP \XXX \top \rightarrow B) \Big) \rightarrow \XXX ( \WHP \XXX \top \rightarrow D_j )$.
By the rule of necessitation of $\SHN$ and the axiom \ref{axiom:SHN K}, we can get $\vdash \SHN \XXX \Big( A \land (\WHP \XXX \top \rightarrow B) \Big) \rightarrow \SHN \XXX ( \WHP \XXX \top \rightarrow D_j )$.
By Lemma \ref{lemma:the magic formula}, $\vdash (\SHN \XXX A \land \WHN \XXX B) \rightarrow \SHN \XXX \Big( A \land (\WHP \XXX \top \rightarrow B) \Big)$. By the axiom \ref{axiom:defining WHN by SHN}, $\vdash \SHN \XXX ( \WHP \XXX \top \rightarrow D_j ) \rightarrow \WHN \XXX D_j$.
Then $\vdash (\SHN \XXX A \land \WHN \XXX B) \rightarrow \WHN \XXX D_j$.
Then 
$\vdash (\SHN \XXX A \land \WHN \XXX B) \rightarrow 
(
\FBV_{i \in \FPI_1} \SHN \XXX C_i \lor \FBV_{i \in \FPI_2} \WHN \XXX D_j \lor \FBV_{i \in \FPI_3} l_i
)$.

\item

Assume $\vdash \FBV_{i \in \FPI_3} l_i$.

Clearly, $\vdash (\SHN \XXX A \land \WHN \XXX B) \rightarrow 
(
\FBV_{i \in \FPI_1} \SHN \XXX C_i \lor \FBV_{i \in \FPI_2} \WHN \XXX D_j \lor \FBV_{i \in \FPI_3} l_i
)$.

\end{enumerate}

\end{proof}

\begin{lemma}[Completeness with respect to $\Phi_\HX$]
\label{lemma:completeness for HX}

For every $\psi \in \Phi_\HX$, if $\models \psi$, then $\vdash_\SWHTN \psi$.

\end{lemma}

\begin{proof}
~

Let $\psi \in \Phi_\HX$. Assume $\models \psi$. We show $\vdash \psi$ by induction on the structure of $\psi$.

Assume the modal depth of $\psi$ is $0$. Then $\psi$ is a formula of the classical propositional logic. As the logic $\SWHTN$ is an extension of the classical propositional logic, $\vdash \psi$.

\medskip

Assume the modal depth of $\psi$ is $1$.

By Lemma \ref{lemma:Normal form of formulas in HX}, the normal form lemma for $\Phi_\HX$, there is $\psi' \in \Phi_\HX$ such that (1) $\vdash \psi \leftrightarrow \psi'$, (2) $\psi'$ is in the form $\chi_1 \land \dots \land \chi_m$, where every $\chi_i$ is a standard formula of $\Phi_\HX$ with the same modal depth as $\psi$.
By the soundness of $\SWHTN$, $\models \psi'$.
Let $j$ be such that $1 \leq j \leq m$. Then, $\models \chi_j$. It suffices to show $\vdash \chi_j$.
Let $\chi_j = (\SHN \XXX A \land \WHN \XXX B) \rightarrow 
(
\FBV_{i \in \FPI_1} \SHN \XXX C_i \lor \FBV_{i \in \FPI_2} \WHN \XXX D_i \lor \FBV_{i \in \FPI_3} l_i
)$.

By Lemma \ref{lemma: Special downward validity lemma for HX}, the special downward validity lemma for $\Phi_\HX$, one of the following conditions holds:
\begin{enumerate}[label=(\alph*),leftmargin=3.33em]

\item

there is $i \in \FPI_1$ such that $\models A \rightarrow C_i$;

\item

there is $j \in \FPI_2$ such that $\models (A \land B) \rightarrow D_{j}$.

\item 

$\models \FBV_{i \in \FPI_3} l_i$.

\end{enumerate}

By the inductive hypothesis, one of the following conditions holds:
\begin{enumerate}[label=(\alph*),leftmargin=3.33em]

\item

$\vdash A \rightarrow C_{i}$;

\item

$\vdash (A \land B) \rightarrow D_{j}$;

\item 

$\vdash \FBV_{i \in \FPI_3} l_i$.

\end{enumerate}

By Lemma \ref{lemma: Special upward derivability lemma for HX}, the special upward derivability lemma for $\Phi_\HX$, $\vdash \chi_j$.

\medskip

Assume the modal depth of $\psi$ is greater than $1$.

By Lemma \ref{lemma:Normal form of formulas in HX}, the normal form lemma for $\Phi_\HX$, there is $\psi' \in \Phi_\HX$ such that (1) $\vdash \psi \leftrightarrow \psi'$, and (2) $\psi'$ is in the form $\chi_1 \land \dots \land \chi_m$, where every $\chi_i$ is a standard formula of $\Phi_\HX$ with the same modal depth as $\psi$.
By the soundness of $\SWHTN$, $\models \psi'$.
Let $j$ be such that $1 \leq j \leq m$. Then, $\models \chi_j$. It suffices to show $\vdash \chi_j$.
Let $\chi_j = (\SHN \XXX A \land \WHN \XXX B) \rightarrow 
(
\FBV_{i \in \FPI_1} \SHN \XXX C_i \lor \FBV_{i \in \FPI_2} \WHN \XXX D_i \lor \FBV_{i \in \FPI_3} l_i
)$.

By Lemma \ref{lemma:Downward validity for HX}, the downward validity lemma for $\Phi_\HX$, one of the following conditions holds:
\begin{enumerate}[label=(\alph*),leftmargin=3.33em]

\item

there is $i \in \FPI_1$ such that $\models \Big( A \land (\WHP \XXX \top \rightarrow B) \Big) \rightarrow C_i$;

\item

there is $j \in \FPI_2$ such that $\models \Big( A \land (\WHP \XXX \top \rightarrow B) \Big) \rightarrow ( \WHP \XXX \top \rightarrow D_j )$;

\item 

$\models \FBV_{i \in \FPI_3} l_i$.

\end{enumerate}

Note $\Big( A \land (\WHP \XXX \top \rightarrow B) \Big) \rightarrow C_i$ and $\Big( A \land (\WHP \XXX \top \rightarrow B) \Big) \rightarrow ( \WHP \XXX \top \rightarrow D_i )$ have lower modal depth than $\psi$. By the inductive hypothesis, one of the following conditions holds:
\begin{enumerate}[label=(\alph*),leftmargin=3.33em]

\item

$\vdash \Big( A \land (\WHP \XXX \top \rightarrow B) \Big) \rightarrow C_i$;

\item

$\vdash \Big( A \land (\WHP \XXX \top \rightarrow B) \Big) \rightarrow ( \WHP \XXX \top \rightarrow D_j )$;

\item 

$\vdash \FBV_{i \in \FPI_3} l_i$.

\end{enumerate}

By Lemma \ref{lemma:Upward derivability for HX}, the upward derivability lemma for $\Phi_\HX$, $\vdash \chi_j$.

\end{proof}

\medskip

From Lemma \ref{lemma:v-reducibility of SWHTN to T(X.Y.H)}, Lemma \ref{lemma:v-reducibility of T(X.Y.H.) to X.Y.H.}, Lemma \ref{lemma:v-reducibility of XYH to XYH mo}, Lemma \ref{lemma:v-reducibility of XYH mo to YHX}, Lemma \ref{lemma:v-reducibility of YHX to YmHX}, Lemma \ref{lemma:v-reducibility of YmHX to YnbHXb}, Lemma \ref{lemma:v-reducibility of Yn(HX) to HX}, and Lemma \ref{lemma:completeness for HX}, the following result follows:

\begin{theorem}[Completeness of $\SWHTN$]

The axiomatic system for $\SWHTN$ given in Definition \ref{definition:An axiomatic system for SWHTN} is complete with respect to the set of valid formulas of $\Phi_{\SWHTN}$.

\end{theorem}

\end{document}